\PassOptionsToPackage{table}{xcolor}
\documentclass[lettersize,journal]{IEEEtran}

\usepackage{amsmath,amsfonts}
\usepackage{algorithmic}
\usepackage{algorithm}
\usepackage{multirow}
\usepackage{array}
\usepackage[caption=false,font=normalsize,labelfont=sf,textfont=sf]{subfig}
\usepackage{textcomp}
\usepackage{stfloats}
\usepackage{url}
\usepackage{booktabs}
\usepackage{verbatim}
\usepackage{graphicx}
\usepackage{subfig}
\usepackage[normalem]{ulem}
\usepackage{tikz}
\usepackage{bbding}
\usepackage{pifont}
\usepackage{wasysym}
\usepackage{paralist}
\usepackage{cite}
\usepackage[edges]{forest}
\definecolor{hidden-draw}{RGB}{106,142,189}
\definecolor{hidden-blue}{RGB}{135,206,250}
\definecolor{hidden-orange}{RGB}{217, 232, 252}
\usepackage{makecell}

\usepackage{soul}
\usepackage{hyperref}
\usepackage{amssymb}
\usepackage{tikz}
\usepackage[table]{xcolor} 

\makeatletter
    \newcommand{\linebreakand}{%
      \end{@IEEEauthorhalign}
      \hfill\mbox{}\par
      \mbox{}\hfill\begin{@IEEEauthorhalign}
    }
    \makeatother
      
\begin{document}

\title{Recent Advances in Multimodal Affective Computing: An NLP Perspective}

\author{Guimin Hu$^{\spadesuit}$, Weimin Lyu$^{\heartsuit}$, Chang Sun$^{\diamond}$, Zhihong Zhu$^{\diamondsuit}$ \\Lin Gui$^{\bigtriangleup}$, Ruichu Cai$^{\spadesuit}$, Erik Cambria$^{\bigoplus}$, Hasti Seifi$^{\ddagger}$
\thanks{
$^{\spadesuit}$Guangdong University of Technology, Guangzhou, China. email: rice.hu.x@gmail.com.
}
\thanks{
$^{\heartsuit}$Stony Brook University, United States.
}
\thanks{
$^{\diamond}$University of Bologna, Italy.
}
\thanks{
$^{\diamondsuit}$Peking University, China.
}
\thanks{
$^{\bigtriangleup}$King's College London, United Kingdom.
}
\thanks{
$^{\bigtriangledown}$Guangdong University of Technology, China.
}
\thanks{
$^{\bigoplus}$Nanyang Technological University, Singapore.
}
\thanks{
$^{\ddagger}$Arizona State University, United States.
}
}



\maketitle

\begin{abstract}
Multimodal affective computing has gained increasing attention due to its broad applications in understanding human behavior and intentions, particularly in text-centric multimodal scenarios. Existing research spans diverse tasks, modalities, and modeling paradigms, yet lacks a unified perspective. In this survey, we systematically review recent advances from an NLP perspective, focusing on four representative tasks: multimodal sentiment analysis (MSA), multimodal emotion recognition in conversation (MERC), multimodal aspect-based sentiment analysis (MABSA), and multimodal multi-label emotion recognition (MMER).
We present a unified view by comparing task formulations, benchmark datasets, and evaluation protocols, and by organizing representative methods into key paradigms, including multitask learning, pre-trained models, knowledge enhancement, and contextual modeling. We further extend the discussion to related directions, such as facial, acoustic, and physiological modalities, as well as emotion cause analysis. Finally, we highlight key challenges and outline promising future directions. To facilitate further research, we release a curated repository of relevant works and resources.\footnote{https://anonymous.4open.science/r/Multimodal-Affective-Computing-Survey-9819}.
\end{abstract}

\section{Introduction}
\emph{Affective computing} is an interdisciplinary field spanning computer science, psychology, and cognitive science, aiming to equip machines with the ability to recognize, interpret, and simulate human emotions~\cite{ijcai2024p739,DBLP:conf/coling/ZhuCHLHZ24,ben2001subtlety,shelly2004emotions,davidson2009handbook,DBLP:conf/acii/RamirezBM11}.

Human perception is inherently multimodal, involving diverse sensory channels such as vision, audition, and touch. A modality refers to a specific form of sensory input through which information is perceived and communicated. Recent advances in multimodal learning across various domains have significantly accelerated the development of multimodal affective computing~\cite{DBLP:conf/nips/LiSGJXH21,DBLP:conf/eccv/GarciaMM18}.

\emph{Multimodal affective computing} focuses on building models that can understand and reason about sentiments and emotions by integrating information from multiple modalities. In its early stages, affective computing primarily focused on unimodal settings, with text-, audio-, and vision-based tasks studied independently. For example, D-MILN~\cite{ji2020diversified} addresses textual sentiment classification, while prior work~\cite{DBLP:conf/icfsp/DonnellyP22} employs BiLSTM models on raw audio to predict crowd sentiment.

\begin{figure}[t]
\centerline{\includegraphics[width=0.95\linewidth]{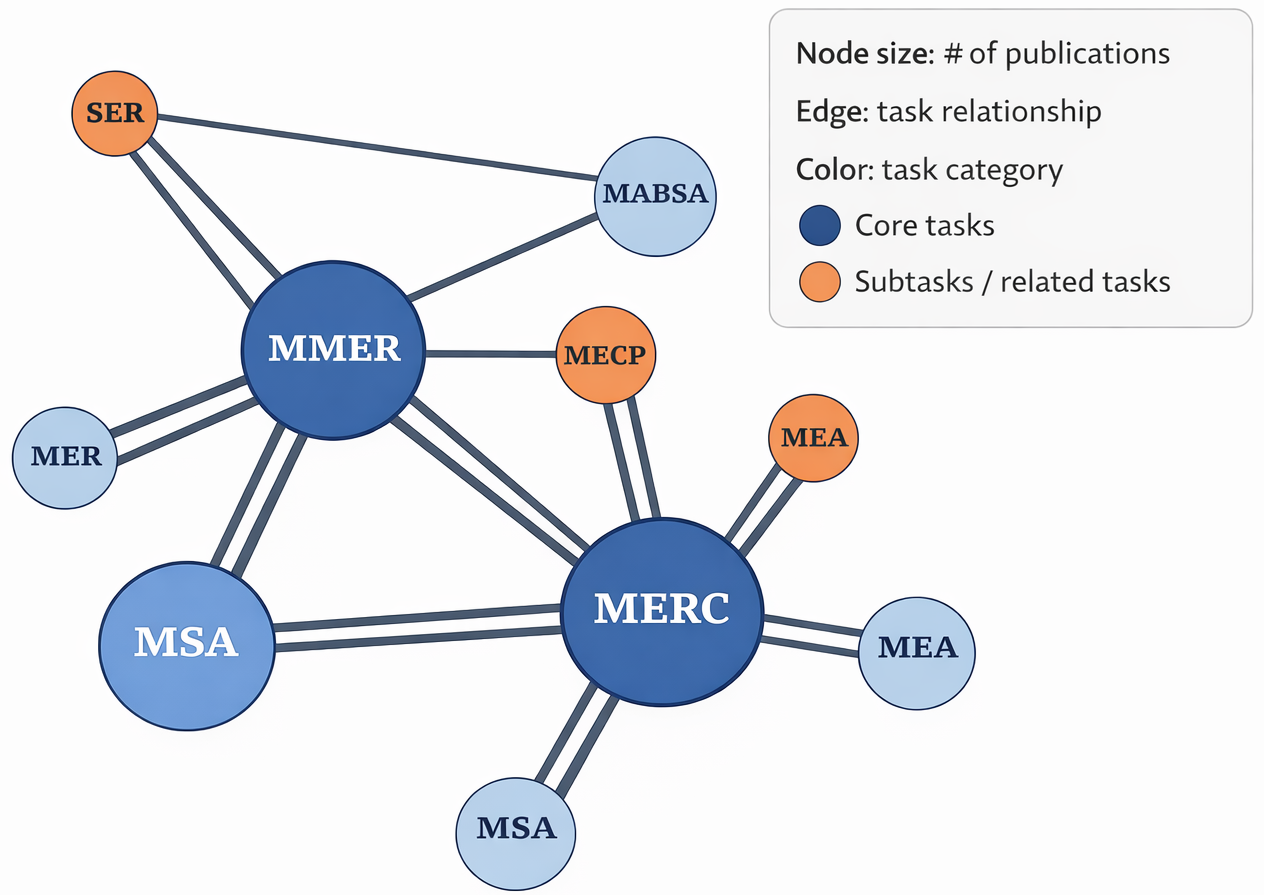}}
\caption{Task relationships in multimodal emotion analysis. Node size indicates publication volume, and edges denote interconnections. MMER and MERC serve as central hubs linking related tasks (e.g., SER, MER, MSA, MABSA, MEA, and MECP).}
\label{fig:affective_world}
\end{figure}

With the rapid development of multimedia technologies~\cite{DBLP:conf/aaai/SunSSL20,DBLP:journals/corr/abs-1911-09826,DBLP:conf/acl/JiZCLN18,DBLP:journals/inffus/GandhiAPCH23}, information is increasingly conveyed through diverse modalities, such as text, audio, and video across social media and online platforms. This shift has driven the integration of textual, acoustic, and visual signals for more comprehensive emotion understanding. Consequently, recent studies have extended traditional unimodal tasks to multimodal settings. For example, Xu et al.~\cite{DBLP:conf/aaai/XuMC19} incorporate visual information into aspect-level sentiment analysis, giving rise to multimodal aspect-based sentiment analysis. Similarly, Wang et al.~\cite{DBLP:journals/corr/abs-2110-08020} extend emotion-cause pair extraction to multimodal conversations, leveraging multimodal signals to enhance emotion understanding.

Despite these advances, most existing approaches remain largely centered on emotion recognition and classification, offering limited ability to model the underlying reasoning process. More recently, research has begun to shift from perception-oriented modeling toward emotion reasoning, with the goal of enabling more interpretable and cognitively plausible analyses. For example, Affective-CoT~\cite{huang2025affective} explicitly disentangles perception from reasoning to support interpretable and faithful emotion analysis. Similarly, VidEmo~\cite{zhangvidemo} introduces an affective cue-guided reasoning framework that progressively integrates attribute perception, expression analysis, and high-level emotion understanding.

From a technical perspective, multimodal affective computing is closely connected to several key learning paradigms, including transfer learning~\cite{DBLP:conf/acl/YangFWSWZHP23,DBLP:journals/kbs/RahmaniHZKKH23,DBLP:conf/acl/LiZLZYH23}, multimodal learning~\cite{DBLP:conf/icml/NgiamKKNLN11,DBLP:conf/acl/RahmanHLZMMH20}, multi-task learning~\cite{DBLP:journals/tkde/ZhangY22,DBLP:conf/emnlp/XieYSLJ21,DBLP:conf/acl/ZhengYXW23}, and semantic understanding~\cite{DBLP:journals/corr/abs-2204-04637,DBLP:journals/corr/abs-2002-06353}. In multimodal affective computing, transfer learning enables cross-domain adaptation of emotion recognition models under limited supervision, improving robustness from controlled to real-world settings. Multimodal learning models interactions among textual, acoustic, and visual cues, where mechanisms such as cross-modal attention support effective alignment and fusion of complementary emotional signals. Multi-task learning jointly optimizes related affective tasks (e.g., emotion recognition, sentiment analysis, and speaker state modeling), promoting shared representations and more generalizable affective patterns.

More recently, the emergence of large language models (LLMs, e.g., LLaMA~\cite{DBLP:journals/corr/abs-2302-13971} and Mistral~\cite{jiang2023mistral}) and vision-language models (VLMs, e.g., LLaVA~\cite{liu2024visual} and Qwen-VL~\cite{bai2023qwen})  has significantly advanced multimodal affective computing. Through large-scale multimodal pre-training, these models learn transferable representations from massive corpora, leading to strong performance on downstream tasks such as multimodal sentiment analysis~\cite{DBLP:conf/icml/RadfordKHRGASAM21,DBLP:conf/nips/BaoW0LMASPW22,DBLP:journals/tmlr/YuWVYSW22,DBLP:conf/nips/AkbariYQCCCG21}. As model scales continue to grow, parameter-efficient adaptation techniques—such as adapters~\cite{DBLP:conf/icml/HoulsbyGJMLGAG19}, prompting~\cite{DBLP:conf/acl/LiL20}, instruction tuning~\cite{wei2021finetuned}, and in-context learning~\cite{DBLP:conf/nips/BrownMRSKDNSSAA20,dong2022survey}—have become increasingly important for effectively leveraging these pre-trained models.

Building on this paradigm, a growing body of work applies parameter-efficient transfer learning to adapt LLMs and VLMs for multimodal affective tasks, achieving strong performance with minimal additional parameters. For example, MPT~\cite{DBLP:journals/corr/abs-2310-04456} enables cross-modal fusion via multimodal prompting, while UniMSE~\cite{DBLP:conf/emnlp/HuLZLWL22} integrates acoustic and visual signals into T5~\cite{DBLP:journals/jmlr/RaffelSRLNMZLL20} through adapter-based fusion. Multimodal affective computing involves tasks such as sentiment analysis, opinion mining, and emotion recognition across diverse modalities, including text, audio, images, video, physiological signals, and haptic feedback. We focus on four representative tasks: \textit{Multimodal Sentiment Analysis} (MSA), \textit{Multimodal Emotion Recognition in Conversation} (MERC), \textit{Multimodal Aspect-Based Sentiment Analysis} (MABSA), and \textit{Multimodal Multi-label Emotion Recognition} (MMER), as shown in Figure \ref{fig:affective_world}. 



In addition, this survey covers recent advances in emotion reasoning, emphasizing interpretable and cognitively plausible analyses. A substantial body of work and several surveys have been devoted to multimodal affective computing~\cite{zhang2024affective,DBLP:journals/ijon/PanHJD23,DBLP:journals/inffus/EzzameliM23,DBLP:journals/inffus/GandhiAPCH23,wangreview,gandhi2023multimodal}. \textit{However, existing reviews largely focus on individual tasks or modalities, leaving a gap in unified, cross-task perspectives and in understanding their shared characteristics and distinctions.} This survey aims to: (1) \textbf{provide a comprehensive introduction to multimodal affective computing for newcomers}, covering tasks, modalities, outputs, and benchmark datasets; and (2) \textbf{offer a structured overview for researchers} by summarizing progress, highlighting trends, and discussing key methodologies, challenges, and future directions.


\begin{figure*}[t]
\centerline{\includegraphics[width=0.95\linewidth]{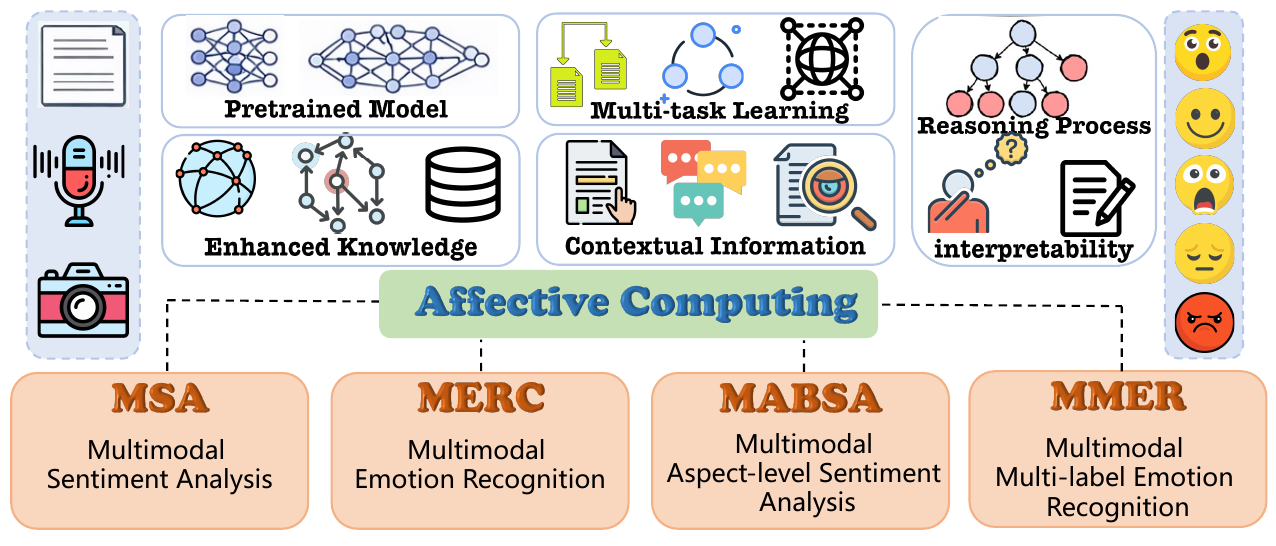}}
\caption{Multimodal affective computing from an NLP perspective.
Multimodal inputs are fused for affective analysis, supporting tasks such as MSA, MERC, MABSA, and MMER, and enabling the transition from emotion recognition to reasoning.}
\label{fig:overview_affective_computing}
\end{figure*}

\section{Organization of this Survey}
Section~\ref{sec:multi_affective_computing_tasks} introduces task formulations and applications. Section~\ref{sec:multimodal_feature_extractor} presents feature extraction and multimodal pre-trained models (e.g., CLIP, BLIP, BLIP2). Section~\ref{sec:ml_mac} analyzes methods for multimodal fusion, alignment, and parameter-efficient adaptation. Section~\ref{sec:models} reviews representative works from the perspectives of multitask learning, pre-trained models, knowledge enhancement, and contextual modeling. Sections~\ref{sec:datasets} and \ref{sec:evaluation} summarize datasets and evaluation metrics. Section~\ref{sec:discuss} discusses additional modalities and emotion causes, highlighting trends from an NLP perspective. Section~\ref{sec:future_work} outlines future directions, and Section~\ref{sec:conclusion} concludes the survey.

\section{Multimodal Affective Computing Tasks}
\label{sec:multi_affective_computing_tasks}
In this section, we show the definition of each task and discuss their application scenarios. 


\subsection{Multimodal Sentiment Analysis}

Multimodal sentiment analysis (MSA) extends traditional sentiment analysis (SA) by incorporating multimodal inputs and aims to predict sentiment polarity and intensity from multimodal signals. Formally, given a multimodal input $I_i = \{I_i^t, I_i^a, I_i^v\}$ from the $i$-th video segment, where $m \in \{t,a,v\}$ denotes text, acoustic, and visual modalities, MSA predicts a real-valued label $y_i^r \in [-3,3]$ representing sentiment intensity, framing the task as both binary classification (for polarity) and regression (for intensity).

\paragraph{Application Scenarios}
Multimodal sentiment analysis (MSA) extends traditional sentiment analysis (SA) by leveraging multimodal inputs to predict sentiment polarity and intensity, supporting applications such as social media monitoring, customer feedback analysis, and market intelligence.

\subsection{Multimodal Emotion Recognition in Conversation} 

Multimodal emotion recognition in conversation (MERC) extends ERC by incorporating multimodal signals (text, audio, and vision) to automatically classify each utterance into predefined emotion categories. Formally, a dialogue with $k$ utterances is denoted as $U = \{u_1, \dots, u_k\}$, where the $i$-th utterance $u_i = \langle I_t^i, I_a^i, I_v^i \rangle$ consists of text, audio, and visual modalities. The corresponding label set is $Y = \{y^i\}_{i=1}^k$, with each $y^i$ drawn from a predefined emotion set.

\paragraph{Application Scenarios}
Multimodal Emotion Recognition in Conversation (MERC) has broad applications in human–computer interaction, virtual assistants, healthcare, and customer service. It enables systems to recognize and adapt to emotional states, improving user experience, conversational naturalness, and customer satisfaction. Moreover, bio-sensing systems (e.g., ECG, PPG, EEG, GSR) further extend MERC to robotics, healthcare, and virtual reality.

\subsection{Multimodal Aspect-based Sentiment Analysis} 
Multimodal aspect-based sentiment analysis (MABSA)~\cite{xu2019multi} extends text-based ABSA~\cite{DBLP:conf/acl/ChenQ19,DBLP:conf/acl/YanDJQ020} to fine-grained image-text inputs. Unlike MSA and ERC, MABSA jointly extracts aspect terms and predicts their sentiment polarities (positive/negative/neutral), framed as classification, tuple extraction, or triple extraction. Formally, given text \(T\) (length \(L\)), image set \(I\) (size \(K\)), and aspect phrase \(A\) (length \(N\)), MABSA predicts sentiment polarities for \(A\).

\paragraph{Application Scenarios}
Multimodal aspect-based sentiment analysis (MABSA) extracts fine-grained, aspect-level opinions from multimodal data, enabling real-world applications such as e-commerce analytics, customer feedback mining, and social media monitoring for product optimization and informed decision-making.

\subsection{Multimodal Multi-label Emotion Recognition}
Multimodal signals often express multiple emotions, motivating multimodal multi-label emotion recognition (MMER), which extends multimodal emotion recognition to predict co-occurring emotions. Formally, given a multimodal input $I_i=\{I_i^t, I_i^a, I_i^v\}$, where each modality $I_i^m \in \mathbb{R}^{d_m \times l_m}$ ($m \in \{t,a,v\}$), MMER aims to assign one or more labels from a predefined set $Y=\{y_1, \dots, y_{|L|}\}$.

\paragraph{Application Scenarios}
Multimodal multi-label emotion recognition aims to jointly interpret and classify co-occurring emotions from multimodal signals, supporting real-world applications such as affect-aware dialogue systems, mental health monitoring, and human–computer interaction. It remains challenging due to the complexity and variability of human emotions, cross-individual and cross-cultural differences, and the difficulty of fusing heterogeneous modalities.

\section{Modal Feature Extractor}
\label{sec:multimodal_feature_extractor}
For multimodal affective computing tasks, inputs typically comprise at least two modalities. This section introduces common feature extractors that convert raw multimodal sequences into feature vectors.

\paragraph{Text Feature Extractor}
For the text modality, early approaches employ static word embeddings such as Word2Vec~\cite{DBLP:conf/nips/MikolovSCCD13} and GloVe~\cite{DBLP:conf/emnlp/PenningtonSM14} to initialize token representations. More advanced methods leverage pre-trained language models (PLMs), including BERT~\cite{DBLP:conf/nips/VaswaniSPUJGKP17}, BART~\cite{DBLP:conf/acl/LewisLGGMLSZ20}, and T5~\cite{DBLP:journals/jmlr/RaffelSRLNMZLL20}, to encode contextualized text features. Recently, foundation models such as LLaMA~\cite{DBLP:journals/corr/abs-2302-13971,DBLP:journals/corr/abs-2307-09288}, Mistral~\cite{jiang2023mistral}, and  Mamba~\cite{DBLP:journals/corr/abs-2312-00752} have further advanced text representation learning.

\paragraph{Audio Feature Extractor}
For the audio modality, raw acoustic signals are typically transformed into numerical sequences. A common approach is to use \texttt{librosa}\footnote{https://github.com/librosa/librosa.} to extract Mel-spectrograms, which capture the short-term power spectrum of sound and are widely adopted in audio processing. Inspired by the success of Transformers, Gong et al.~\cite{DBLP:conf/interspeech/GongCG21} propose AST, which encodes waveforms into log-Mel filterbank features. Beyond it, self-supervised models (Wav2Vec 2.0~\cite{baevski2020wav2vec}, HuBERT~\cite{hsu2021hubert}) learn rich representations from unlabeled speech, while multi-task Whisper~\cite{radford2023robust} provides robust encoder features from labeled audio for downstream tasks like emotion recognition.

\paragraph{Vision Feature Extractor}
For the visual modality, a common strategy is to sample a fixed number of frames from each video segment and extract features using pretrained CNNs such as EfficientNet~\cite{DBLP:conf/icml/TanL19}. Transformer-based approaches, such as Vision Transformer (ViT)~\cite{DBLP:conf/iclr/DosovitskiyB0WZ21}, model images by splitting them into patches and encoding them as sequences. Additionally, multimodal models like CLIP~\cite{radford2021learning} learn aligned visual-text representations via contrastive learning, enabling the extraction of semantically rich visual features.

\paragraph{Multimodal Feature Extractor}
Multimodal pre-trained models (MPMs) advance multimodal integration (e.g., GPT-4, Gemini)~\cite{2023GPT4VisionSC,team2023gemini}. Open-source efforts include Flamingo (cross-attention)~\cite{alayrac2022flamingo}, BLIP-2 (Q-Former adapter)~\cite{li2023blip}, MiniGPT-4 (linear projection)~\cite{zhu2023minigpt}, InstructBLIP (instruction tuning)~\cite{instructblip}, and LLaVA (CLIP+LLaMA)~\cite{liu2024visual}. VATT~\cite{DBLP:conf/nips/AkbariYQCCCG21} is trained end-to-end with contrastive losses for video, audio, image, and text tasks. These MPMs enable direct extraction of modality-specific features from raw signals.

\section{Multimodal Learning on Multimodal Affective Computing}
\label{sec:ml_mac}

With the scaling of pre-trained models, parameter-efficient transfer methods (e.g., adapters~\cite{houlsby2019parameter}, prompting~\cite{hetowards}, instruction tuning~\cite{li2021prefix}, and in-context learning~\cite{brown2020language}) have emerged, reformulating downstream tasks to align with pre-training objectives. In VLMs (e.g., GPT-4V~\cite{achiam2023gpt} and Flamingo~\cite{alayrac2022flamingo}), prompting enables joint visual-textual reasoning, while instruction tuning improves generalization. These techniques facilitate efficient knowledge transfer to multimodal affective tasks with minimal fine-tuning. Multimodal learning learns representations from multiple modalities by first aligning them semantically, then fusing them into a single vector.

\begin{figure}[t]
\centerline{\includegraphics[width=\linewidth]{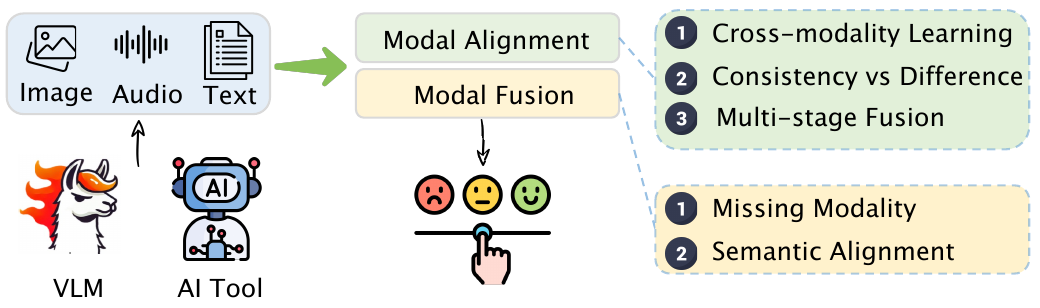}}
\caption{Illustration of modal fusion and modal alignment.}
\label{fig:overview_modal_fusion}
\end{figure}

Given the inherently multimodal nature of affective computing, existing works can be broadly analyzed from two perspectives: \textit{multimodal alignment} and \textit{multimodal fusion}, as illustrated in Fig.~\ref{fig:overview_modal_fusion} and Fig.~\ref{fig:summary_models}.

\tikzstyle{my-box}=[
 rectangle,
 draw=hidden-draw,
 rounded corners,
 text opacity=1,
 minimum height=1.5em,
 minimum width=5em,
 inner sep=2pt,
 align=center,
 fill opacity=.5,
 ]
 \tikzstyle{leaf}=[my-box, minimum height=1.5em,
 fill=hidden-orange!60, text=black, align=left,font=\scriptsize,
 inner xsep=2pt,
 inner ysep=4pt,
 ]
\begin{figure*}[h]
	\centering
	\resizebox{\textwidth}{!}{
		\begin{forest}
			forked edges,
			for tree={
				grow=east,
				reversed=true,
				anchor=base west,
				parent anchor=east,
				child anchor=west,
				base=left,
				font=\small,
				rectangle,
				draw=hidden-draw,
				rounded corners,
				align=left,
				minimum width=4em,
				edge+={darkgray, line width=1pt},
				s sep=3pt,
				inner xsep=2pt,
				inner ysep=3pt,
				ver/.style={rotate=90, child anchor=north, parent anchor=south, anchor=center},
			},
			where level=1{text width=4.8em,font=\scriptsize}{},
			where level=2{text width=6.3em,font=\scriptsize}{},
			where level=3{text width=5.5em,font=\scriptsize}{},
			[
			Multimodal Learning on \\ Affective Computing, ver
  			[
			 Multimodal \\ Fusion (\S\ref{sec:multimodal_fusion})
			[
          Cross-modal Learning
            [
            TFN~{\cite{DBLP:conf/emnlp/ZadehCPCM17},}
            MuLT~{\cite{DBLP:conf/acl/TsaiBLKMS19},}
            TCDN~{\cite{DBLP:journals/taslp/ChenHGS23},}
            CM-BERT~{\cite{DBLP:conf/mm/YangXG20},}
            HGraph-CL~{\cite{DBLP:conf/coling/LinLLDYZX22},}
            BAFN~{\cite{DBLP:journals/tcsv/TangLJPZDK23},}
            TeFNA~{\cite{DBLP:journals/kbs/HuangZWW0H23},}\\
            CMCF-SRNet~{\cite{zhang-li-2023-cross},}
            MultiEMO~{\cite{DBLP:conf/acl/ShiH23},}
            MM-RBN~{\cite{chen2024mmrbn},}
            MAGDRA~{\cite{li2024magdra},}
            AMuSE~{\cite{DBLP:journals/corr/abs-2401-15164}.},leaf, text width=34.1em
            ]
			]
  			[
          Modal Consistency \\ and Difference
            [
              MMIM~{\cite{DBLP:conf/emnlp/HanCP21},}
              MPT~{\cite{DBLP:journals/corr/abs-2109-01797},}
              MMMIE~{\cite{zheng2022multimodal},}
              MISA~{\cite{DBLP:conf/mm/HazarikaZP20},}
              CoolNet~{\cite{DBLP:journals/ipm/XiaoWYXZH23},}
              ModalNet~{\cite{DBLP:journals/www/ZhangWLLGY21},}
              MAN~{\cite{DBLP:journals/corr/abs-2401-15164},}
              TAILOR~{\cite{zhang2022tailor},}\\
              AMP~{\cite{DBLP:conf/www/GeJCWYG23},}
              STCN~{\cite{DBLP:journals/bspc/SharafiYRN22},}
              EMOE~{\cite{fang2025emoe},}
              MCG~{\cite{DBLP:conf/ijcai/ZhuSHZCH25}.}
              ,leaf, text width=34.1em
            ]
			]
          [
         Multi-stage\\ Modal Fusion
            [
              TSCL-FHFN~{\cite{li2024tscl},} 
              HFFN~{\cite{DBLP:conf/acl/MaiHX19},} CLMLF~{\cite{DBLP:conf/naacl/LiXZZ22},}
              RMFN~{\cite{DBLP:conf/emnlp/LiangLZM18},}
              CTFN~{\cite{DBLP:conf/acl/TangLJCZK20},}
              MCM~{\cite{DBLP:journals/inffus/LiGPDYZLCWX23},}
              FmlMSN~{\cite{DBLP:journals/eswa/PengWZCTYH23},}\\
              ScaleVLAD~{\cite{DBLP:journals/corr/abs-2112-01368},}
              MUG~{\cite{mai2024meta},}
              HFCE~{\cite{minglong2024multimodal},}
              MTAG~{\cite{DBLP:conf/naacl/YangWYZRZPM21},} CHFusion~{\cite{majumder2018multimodal},} 
              BIG-FUSION{~\cite{DBLP:conf/aaai/WangFYLLXXZX25},}
              ,leaf, text width=34.1em
            ]
			]
			]
  			[
			 Multimodal \\Alignment (\S\ref{sec:multimodal_alignment})
			[
          Miss Modality
            [
            MMIN~{\cite{DBLP:conf/acl/ZhaoLJ20},} 
            CMAL~{\cite{DBLP:conf/icmi/ParthasarathyS20},}
            M2R2~{\cite{DBLP:journals/tai/WangCZCYZ23},}
            EMMR~{\cite{zeng2022mitigating},}
            TFR-Net~{\cite{yuan2021transformer},}
            MRAN~{\cite{DBLP:conf/mmm/LuoXL23},}
            VIGAN~{\cite{DBLP:conf/bigdataconf/ShangPSCLB17},}\\
            TATE~{\cite{DBLP:conf/sigir/ZengL022},}
            IF-MMIN~{\cite{DBLP:conf/icassp/ZuoLZGL23},}
            CTFN~{\cite{DBLP:conf/acl/TangLJCZK20},}
            MTMSA~{\cite{DBLP:journals/inffus/LiuZCSM24},}
            FGR~{\cite{DBLP:journals/ijon/ZhouCVR21},}
            MMTE+AMMTD ~{\cite{DBLP:conf/acii/VazquezRodriguezLCC23}.}
            , leaf, text width=34.1em
            ]
			]
  			[
          Semantic Alignment
            [
            MuLT~{\cite{DBLP:conf/acl/TsaiBLKMS19},}
            ScaleVLAD~{\cite{DBLP:journals/corr/abs-2112-01368},}
            Robust-MSA~{\cite{DBLP:conf/aaai/MaoZXYL23},}
            HGraph-CL~{\cite{DBLP:conf/coling/LinLLDYZX22},}
            SPIM~{\cite{lai2023shared},}
            MA-CMU-SGRNet~{\cite{zhang2024multi},}\\
            DaNet{~\cite{zhu2025danet}.}
              ,leaf, text width=34.1em
            ]
			]
			]
			]
		\end{forest}
}
\caption{Taxonomy of multimodal affective computing, highlighting multimodal fusion (cross-modal learning, modality consistency/difference and modal fusion) and multimodal alignment (missing modality, semantic alignment).} 
\vspace{-0.1in}
\label{fig:summary_models}
\end{figure*}

\subsection{Multimodal Alignment}
\label{sec:multimodal_alignment}

\paragraph{Alignment for Missing Modality}

In real-world scenarios, modalities are often partially missing, violating the assumption of complete inputs and challenging multimodal fusion and alignment.
Existing methods for handling missing modalities can be grouped into four categories.(1) \textit{Data augmentation} simulates missing cases by randomly ablating inputs during training~\cite{DBLP:conf/icmi/ParthasarathyS20,DBLP:journals/tai/WangCZCYZ23}.
(2) \textit{Generative methods} reconstruct missing modalities from available ones via prediction or feature recovery~\cite{du2018semi,DBLP:conf/acl/ZhaoLJ20,zeng2022mitigating,yuan2021transformer,DBLP:conf/mmm/LuoXL23}.
(3) \textit{Joint representation} learning learns modality-invariant and complementary features in a shared latent space to mitigate missing data~\cite{DBLP:conf/www/WangW020,DBLP:conf/icmcs/MaHZ21,DBLP:conf/sigir/ZengL022,DBLP:conf/icassp/ZuoLZGL23,DBLP:journals/ijon/ZhouCVR21}.
(4) \textit{Translation-based methods} infer missing modalities via cross-modal translation~\cite{DBLP:conf/acl/TangLJCZK20,DBLP:journals/inffus/LiuZCSM24}.
Overall, these approaches focus on reconstructing missing modalities or learning robust representations from incomplete inputs.

\paragraph{Alignment for Cross-modal Semantics}
Semantic alignment aims to establish cross-modal correspondences within a sample, enabling one modality to retrieve or ground information from another. In multimodal sentiment analysis (MSA), existing methods address alignment from multiple perspectives. Cross-modal and multi-scale alignment mechanisms are widely used to capture semantic consistency~\cite{DBLP:conf/acl/TsaiBLKMS19}, while approaches such as ScaleVLAD~\cite{luo2021scalevlad} aggregate multi-scale representations to handle unaligned data. Graph-based methods further model temporal and cross-modal interactions by converting multimodal sequences into heterogeneous graphs~\cite{DBLP:conf/naacl/YangWYZRZPM21}. To explicitly align sequences, some works enforce temporal consistency across modalities~\cite{DBLP:journals/corr/abs-2207-12895}, whereas translation-based methods learn bidirectional mappings to enhance cross-modal correspondence~\cite{DBLP:conf/acl/2021-1,DBLP:journals/cogcom/WangTYLWLW23}. Recent studies extend alignment to multiple levels, capturing both instance- and prototype-level relationships~\cite{zhang2024multi}, or minimizing distribution gaps via distance-based objectives such as Wasserstein distance~\cite{yu2022dual}. Additionally, shared–private representation learning~\cite{lai2023shared} disentangles modality-invariant and modality-specific information, while robustness-oriented frameworks analyze the impact of modality noise on alignment performance~\cite{DBLP:conf/aaai/MaoZXYL23}. Overall, semantic alignment focuses on modeling cross-modal correspondence through temporal alignment, structural modeling, translation, and shared–private representation learning.

\subsection{Multimodal Fusion}
\label{sec:multimodal_fusion}
Multimodal signals are heterogeneous, requiring effective fusion into unified representations. Fusion strategies are typically categorized into early, late, and intermediate fusion~\cite{DBLP:conf/acl/TsaiBLKMS19}. Early and late fusion often fail to capture intra- and inter-modal dynamics, while intermediate fusion enables richer cross-modal interactions~\cite{poria2017context,DBLP:journals/kbs/MiddyaNR22,DBLP:conf/emnlp/HuLZLWL22}. Building on this taxonomy, we review multimodal fusion from three perspectives: cross-modality learning, modal consistency and difference, and multi-stage fusion.

\paragraph{Cross-modality Learning}
Cross-modality learning aims to model inter- and intra-modal dependencies to improve multimodal representation learning. Early approaches~\cite{DBLP:conf/emnlp/ZadehCPCM17} primarily relied on geometric operations in feature space for modality fusion. More recent methods commonly adopt attention-based mechanisms to capture complex cross-modal interactions. For instance, MuLT~\cite{DBLP:conf/acl/TsaiBLKMS19} employs a multimodal Transformer to learn inter-modal dependencies, while CM-BERT~\cite{DBLP:conf/mm/YangXG20} extends pre-trained BERT to model text–audio interactions. Subsequent works further enhance cross-modal modeling by jointly capturing inter- and intra-modal dynamics, such as trimodal collaborative interaction~\cite{DBLP:journals/taslp/ChenHGS23}, cross-modal attention for unaligned sequences~\cite{DBLP:journals/kbs/HuangZWW0H23}, and dynamic mechanisms to reduce intra-modal redundancy~\cite{DBLP:journals/tcsv/TangLJPZDK23}. In emotion recognition, models like CMCF-SRNet~\cite{zhang-li-2023-cross} and MultiEMO~\cite{DBLP:conf/acl/ShiH23} further exploit cross-modal dependencies through transformers and attention-based correlation modeling. Overall, cross-modality learning focuses on effectively capturing relationships and interactions across modalities.

\paragraph{Modal Consistency and Difference}
Modal consistency refers to the shared representation across modalities for the same sample, while modal difference captures modality-specific information. Most multimodal fusion approaches explicitly disentangle these two components into modality-invariant (consistency) and modality-specific (difference) representations. Modal consistency improves robustness to missing modalities, whereas modal difference exploits complementary information to enhance overall understanding.

Recent studies often leverage contrastive learning to model this decomposition. For example, prior works~\cite{DBLP:journals/corr/abs-2109-01797,zheng2022multimodal} jointly perform intra- and inter-modal contrastive learning to reduce modality gaps and preserve semantic relationships. Han et al.~\cite{DBLP:conf/emnlp/HanCP21} maximize mutual information across modalities to capture shared representations, while Zheng et al.~\cite{zheng2022multimodal} combine mutual information maximization and minimization to extract modality-invariant and task-relevant features. From a representation perspective, modal consistency corresponds to projecting modalities into a shared latent space, whereas modal difference involves learning modality-specific spaces. For instance, Hazarika et al.~\cite{DBLP:conf/mm/HazarikaZP20} jointly learn invariant and specific representations with reconstruction constraints. Subsequent works further enhance this paradigm through attention-based interaction modeling~\cite{DBLP:journals/corr/abs-2401-15164}, fine-grained vision-language integration~\cite{DBLP:journals/ipm/XiaoWYXZH23}, and consistency-aware attention mechanisms for sentiment analysis~\cite{DBLP:journals/www/ZhangWLLGY21}.


\paragraph{Multi-stage Modal Fusion}
Multi-stage multimodal fusion integrates information across multiple stages or scales to enhance representation learning~\cite{baltruvsaitis2018multimodal,DBLP:journals/pami/ZengPRH09}. Existing approaches decompose fusion into hierarchical or staged processes to capture both local and global interactions. For example, HFFN~\cite{DBLP:conf/acl/MaiHX19} models local and global dependencies via a divide–conquer–combine strategy, while other works perform token-level alignment and contrastive learning to capture shared sentiment features~\cite{DBLP:conf/naacl/LiXZZ22}.

Additionally, hierarchical and multi-level fusion methods progressively integrate unimodal, bimodal, and trimodal information~\cite{DBLP:conf/emnlp/LiangLZM18,majumder2018multimodal}, enabling fine-grained and scale-aware interaction modeling. Recent studies further incorporate multi-level correlation mining and multi-task learning to improve multimodal understanding~\cite{DBLP:journals/inffus/LiGPDYZLCWX23,DBLP:journals/eswa/PengWZCTYH23}. Overall, multi-stage fusion emphasizes structured, hierarchical integration of modalities prior to final decision-making.

\tikzstyle{my-box}=[
 rectangle,
 draw=hidden-draw,
 rounded corners,
 text opacity=1,
 minimum height=1.5em,
 minimum width=5em,
 inner sep=2pt,
 align=center,
 fill opacity=.5,
 ]
 \tikzstyle{leaf}=[my-box, minimum height=1.5em,
 fill=hidden-orange!60, text=black, align=left,font=\scriptsize,
 inner xsep=2pt,
 inner ysep=4pt,
 ]
\begin{figure*}[h]
	\begin{center}
	\resizebox{\textwidth}{!}{
		\begin{forest}
			forked edges,
			for tree={
				grow=east,
				reversed=true,
				anchor=base west,
				parent anchor=east,
				child anchor=west,
				base=left,
				font=\small,
				rectangle,
				draw=hidden-draw,
				rounded corners,
				align=left,
				minimum width=4em,
				edge+={darkgray, line width=1pt},
				s sep=3pt,
				inner xsep=2pt,
				inner ysep=3pt,
				ver/.style={rotate=90, child anchor=north, parent anchor=south, anchor=center},
			},
			where level=1{text width=4.8em,font=\scriptsize}{},
			where level=2{text width=6.3em,font=\scriptsize}{},
			where level=3{text width=5.5em,font=\scriptsize}{},
			[
			Multimodal Affective Computing, ver
  			[
			 Multi-task \\ Learning (\S\ref{sec:multitask_learning})
			[
          MSA (\S\ref{sec:mt_msa})
            [
            Self-MM~{\cite{DBLP:conf/aaai/YuXYW21},} 
            ARGF~{\cite{DBLP:conf/aaai/Mai0X20},} 
            MultiSE~{\cite{DBLP:conf/naacl/AkhtarCGPEB19},} 
            VCAN~{\cite{DBLP:journals/tcsv/ChenZLZ22},}
            DTN~{\cite{DBLP:journals/inffus/ZengYMH24},}
            MMMIE~{\cite{zheng2022multimodal},}
            MMIM~{\cite{DBLP:conf/emnlp/HanCP21},}
            MISA~{\cite{DBLP:conf/mm/HazarikaZP20}.}
            , leaf, text width=34.1em
            ]
			]
  			[
          MERC (\S\ref{sec:mt:merc})
            [
              FacialMMT~{\cite{DBLP:conf/acl/ZhengYXW23},} MMMIE~{\cite{zheng2022multimodal},} AuxEmo~{\cite{DBLP:journals/corr/abs-2302-13661},} TDFNet~{\cite{DBLP:journals/taslp/ZhaoWSXZ23},} 
              MALN~{\cite{DBLP:journals/tcsv/RenHLLLL23},} 
              LGCCT~{\cite{DBLP:journals/entropy/LiuSFWZQ22},}
              MultiEMO~{\cite{DBLP:conf/acl/ShiH23},}\\
              RLEMO~{\cite{10446459},}
              Multi-to-Single~{\cite{DBLP:conf/aaai/LiuCLZ25}.}
              ,leaf, text width=34.1em
            ]
			]
   			[
          MABSA (\S\ref{sec:mt:absa})
            [
              CMMT~{\cite{DBLP:journals/ipm/YangNY22},} AbCoRD~{\cite{DBLP:conf/mm/JainSG023},} JML~{\cite{DBLP:conf/emnlp/JuZXLLZZ21},}
              MPT~{\cite{DBLP:journals/corr/abs-2310-04456},}
              MMRBN~{\cite{chen2024mmrbn}.}
              ,leaf, text width=34.1em
            ]
			]
          [
          MMER (\S\ref{sec:mt:mmer})
            [
              AMP~{\cite{DBLP:conf/www/GeJCWYG23},} MEGLN-LDA~{\cite{DBLP:conf/mrac/LianLLZTZZ23},} MultiSE~{\cite{DBLP:conf/naacl/AkhtarCGPEB19}.}
              ,leaf, text width=34.1em
            ]
			]
			]
  			[
			 Pre-trained \\Model (\S\ref{sec:pretrained_model})
			[
          MSA (\S\ref{sec:pm:msa})
            [
            MAG-XLNet~{\cite{DBLP:conf/acl/RahmanHLZMMH20},} 
            UniMSE~{\cite{DBLP:conf/emnlp/HuLZLWL22},} 
            AOBERT~{\cite{KIM202337},}
            SKESL~{\cite{qian2023sentiment},}
            TEASAL~{\cite{DBLP:journals/corr/abs-2109-05522},}
            TO-BERT~{\cite{DBLP:conf/ijcai/Yu019},}
            SPT~{\cite{DBLP:conf/emnlp/ChengFB021}.}\\
            ALMT~{\cite{DBLP:journals/corr/abs-2310-05804}}
            , leaf, text width=34.1em
            ]
			]
  			[
          MERC (\S\ref{sec:pm:merc})
            [
              FacialMMT~{\cite{DBLP:conf/acl/ZhengYXW23},}
              QAP~{\cite{DBLP:conf/acl/LiZLZYH23},}
              UniMSE~{\cite{DBLP:conf/emnlp/HuLZLWL22},} 
              GraphSmile~{\cite{li2024tracing},} 
              MSE-Adapter~{\cite{DBLP:conf/aaai/YangDQ25},}
              UniMEEC~{\cite{hu2024unimeec}.}
              ,leaf, text width=34.1em
            ]
			]
   			[
          MABSA (\S\ref{sec:pm:mabsa})
            [
              MIMN~{\cite{DBLP:conf/acl/ZhengYXW23},}
              GMP~{\cite{DBLP:conf/acl/YangFWSWZHP23},}
              ERUP~{\cite{DBLP:conf/nlpcc/LiuWZ23},}
              VLP-MABSA~{\cite{DBLP:conf/acl/LingYX22},}
              DR-BERT~{\cite{DBLP:conf/acl/ZhangZZZL0C22},}
              DTCA~{\cite{DBLP:conf/ijcnlp/YuWYZ22},}
              MSRA~{\cite{DBLP:conf/cikm/JinTLQYCZ23},}\\
              AOF-ABSA~{\cite{wang2023image},}
              AD-GCFN~{\cite{wang2024adaptive},}
              MOCOLNet~{\cite{mu2023mocolnet}.}
              ,leaf, text width=34.1em
            ]
			]
        [
          MMER (\S\ref{sec:pm:mmer})
            [
              TAILOR~{\cite{zhang2022tailor},}
              ,leaf, text width=34.1em
            ]
			]
			]
  			[
			 Enhanced \\ Knowledge \\ (\S\ref{sec:enhanced_knowlege})
			[
          MSA (\S\ref{sec:ek:msa})
            [
            TETFN~{\cite{DBLP:journals/pr/WangGTLHL23},} 
            ITP~{\cite{DBLP:journals/kbs/RahmaniHZKKH23},} 
            SKEAFN~{\cite{DBLP:journals/inffus/ZhuCZSLLC23},}
            SAWFN~{\cite{chen2020swafn},}
            MTAG~{\cite{DBLP:conf/naacl/YangWYZRZPM21},} 
            KuDA~{\cite{feng2024knowledge}.}
            ,leaf, text width=34.1em
            ]
			]
  			[
          MERC (\S\ref{sec:ek:merc})
            [
              ConSK-GCN~{\cite{DBLP:journals/ieeemm/FuOWGSLD22},} 
              DMD~{\cite{DBLP:conf/cvpr/LiW023},}
              MRST~{\cite{DBLP:conf/ccis/SunHTZS23},}
              SF~{\cite{DBLP:journals/corr/abs-2309-02106},}
              TGMFN~{\cite{yuan2024topics},}
              RLEMO~{\cite{10446459},}
              DEAN~{\cite{DBLP:journals/inffus/ZhangLLHDZ22}.}
              ,leaf, text width=34.1em
            ]
			]
   			[
          MABSA (\S\ref{sec:ek:mabsa})
            [
              KNIT~{\cite{DBLP:conf/icmcs/XuSX23},} 
              FITE~{\cite{DBLP:conf/emnlp/YangZ022},} 
              CoolNet~{\cite{xiao2023cross},} 
              HIMT~{\cite{yu2022hierarchical},}
              DEQA~{\cite{DBLP:conf/aaai/HanHBWL25},}
              AETS~{\cite{DBLP:conf/aaai/ZhuSGL025}.}
              ,leaf, text width=34.1em
            ]
			]
        [
          MMER (\S\ref{sec:ek:mmer})
            [
              UniVA-RoBERTa~{\cite{zheng2024unimodal},} 
              CARAT~{\cite{peng2024carat},} 
              M3TR~{\cite{zhao2021m3tr},} 
              MAGDRA~{\cite{li2024magdra},} 
              HHMPN~{\cite{zhang2021multi}.} 
              ,leaf, text width=34.1em
            ]
			]
			]
			[
			 Contextual \\ Information \\ (\S\ref{sec:context_information})
			[
          MSA (\S\ref{sec:ci:msa})
            [
            MuLT~{\cite{DBLP:conf/acl/TsaiBLKMS19},}
            CIA~{\cite{DBLP:conf/emnlp/ChauhanAEB19},} 
            CAT-LSTM~{\cite{DBLP:conf/icdm/PoriaCHMZM17},} 
            CAMFNet~{\cite{DBLP:journals/taslp/HuangQTX23},} 
            MTAG~{\cite{DBLP:conf/naacl/YangWYZRZPM21},}
            CTNet~{\cite{DBLP:journals/pami/LiSZT22},}
            ScaleVLAD~{\cite{DBLP:journals/corr/abs-2112-01368},}\\
            MMML~{\cite{sun2024novel},}
            GFML~{\cite{sun2024novel},}
            CHFusion~{\cite{majumder2018multimodal},}
            KAN-MCP~{\cite{luo2025towards},}
            EMOE~{\cite{fang2025emoe}.}
            , leaf, text width=34.1em
            ]
			]
  			[
          MERC (\S\ref{sec:ci:merc})
            [
              CMCF-SRNet~{\cite{zhang-li-2023-cross},}
              MMGCN~{\cite{DBLP:conf/acl/HuLZJ20},}
              MM-DFN~{\cite{DBLP:conf/icassp/HuHWJM22},}
              SAMGN~{\cite{DBLP:journals/tmm/ZhangCCCT24},} 
              M3Net~{\cite{DBLP:conf/cvpr/Chen0ZS23},} 
              M3GAT~{\cite{DBLP:conf/acl/HuLZJ20},} 
              RL-EMO~{\cite{10446459},}\\
              SCMFN~{\cite{10447720},} 
              EmoCaps~{\cite{li2022emocaps},}
              GA2MIF~{\cite{DBLP:journals/taffco/LiWLZ24},}
              MALN~{\cite{DBLP:journals/tcsv/RenHLLLL23},}
              COGMEN~{\cite{DBLP:journals/corr/abs-2205-02455}.}
              ,leaf, text width=34.1em
            ]
			]
   			[
          MABSA (\S\ref{sec:ci:absa})
            [
              DTCA~{\cite{DBLP:conf/ijcnlp/YuWYZ22},} 
              MCPR~{\cite{DBLP:conf/iccc2/XuLW22},} 
              Elbphilharmonie~{\cite{DBLP:conf/semco/AnschutzEG23},} 
              M2DF~{\cite{DBLP:journals/corr/abs-2310-14605},} 
              AoM~{\cite{DBLP:conf/acl/ZhouGLYZY23},}
              FGSN~{\cite{zhao2023fusion},} 
              MIMN~{\cite{DBLP:conf/aaai/XuMC19}.}
              ,leaf, text width=34.1em
            ]
			]
        [
          MMER (\S\ref{sec:ci:mmer})
            [  MMS2S~{\cite{DBLP:conf/emnlp/ZhangJLLZZ20},}
              MESGN~{\cite{DBLP:conf/mm/JuZLZ20},}
              MDI~{\cite{DBLP:conf/acl/ZhaoZ0LJW022}.}
              ,leaf, text width=34.1em
            ]
			]
			]
        [
			 Reasoning \\Model (\S\ref{sec:reasoning})
			[
          Benchmark (\S\ref{sec:pm:msa})
            [
            CA-MER~{\cite{han2025benchmarking},} 
            MTMEUR~{\cite{hu2025beyond},} 
            AffectGPT~{\cite{lian2025affectgpt},}
            HitEmotion~{\cite{luo2026unveiling},}
            EmoReAlM~{\cite{chaubeyavere}.}
            , leaf, text width=34.1em
            ]
			]
  			[
           Framework (\S\ref{sec:pm:merc})
            [
              EmotionThinker~{\cite{dingdongemotionthinker},}
              VidEmo~{\cite{zhangvidemo},}
              Emosym~{\cite{zhu2025emosym},} 
              Deemo~{\cite{li2025deemo},} 
              AVERE~{\cite{chaubeyavere},} 
              Affective-CoT~{\cite{huang2025affective},}
              TMPO~{\cite{luo2026unveiling}.}
              ,leaf, text width=34.1em
            ]
			]
			]
			]
		\end{forest}
}
\end{center}
\vspace{-0.1in}
\caption{Taxonomy of multimodal affective computing works across multitask learning, pre-trained models, knowledge enhancement, contextual information, and reasoning models.} 
\label{fig:multimodal_works}
\end{figure*}

\section{Models Across Multimodal Affective Computing}
\label{sec:models}
We categorize multimodal affective computing into four paradigms—multitask learning, pre-trained models, knowledge enhancement, and contextual modeling—and review advances in MSA, MERC, MABSA, MMER, and emotion reasoning (Fig.~\ref{fig:multimodal_works}, Table~\ref{tab:summary_models}).

\subsection{Multi-task Learning}
\label{sec:multitask_learning}
Multitask learning jointly optimizes multiple related tasks with shared representations, improving performance through shared supervision. In multimodal affective computing, it facilitates the disentanglement of modality-invariant and modality-specific features while unifying emotion-related sub-tasks within a single framework. 

\paragraph{Multimodal Sentiment Analysis}
\label{sec:mt_msa}
In multimodal sentiment analysis, multitask learning is widely adopted to enhance representation learning. For example, Self-MM~\cite{DBLP:conf/aaai/YuXYW21} generates pseudo-labels~\cite{ge2020mutual,DBLP:conf/cvpr/PhamDXL21,DBLP:conf/acl/ZhangZWZ22} for unimodal data and jointly trains unimodal and multimodal representations. Translation-based auxiliary tasks, such as ARGF~\cite{DBLP:conf/aaai/Mai0X20}, further regularize cross-modal learning. Other works exploit task interdependence and auxiliary information to improve performance, including joint sentiment–emotion learning~\cite{DBLP:conf/naacl/AkhtarCGPEB19} and cross-modal auxiliary networks like VCAN~\cite{DBLP:journals/tcsv/ChenZLZ22}. More recent approaches focus on disentangled and information-theoretic learning, such as DTN~\cite{DBLP:journals/inffus/ZengYMH24} and mutual-information-based methods~\cite{zheng2022multimodal}, to capture modality-invariant, task-relevant features while reducing redundancy.

\paragraph{Multimodal Emotion Recognition in Conversation}
\label{sec:mt:merc}
In multimodal emotion recognition, multitask learning is widely used to enhance feature learning and cross-modal interaction. For example, FacialMMT~\cite{DBLP:conf/acl/ZhengYXW23} employs a two-stage framework that jointly optimizes face recognition, clustering, and matching to incorporate frame-level facial cues into utterance-level emotion recognition. Zhang et al.~\cite{DBLP:journals/inffus/ZhangWLRZSTQ23} explore both single-level and multi-level MTL decoders, while Sun et al.~\cite{DBLP:journals/corr/abs-2302-13661} introduce auxiliary tasks to improve cross-modal fusion and alignment.
Other works further enhance representation learning through advanced architectures, including transformer-based fusion (TDFNet) with pre-trained knowledge integration~\cite{DBLP:journals/taslp/ZhaoWSXZ23}, adversarial learning for speaker-aware modeling (MALN)~\cite{DBLP:journals/tcsv/RenHLLLL23}, and lightweight cross-modal transformers (LGCCT)~\cite{DBLP:journals/entropy/LiuSFWZQ22}.

\paragraph{Multimodal Aspect-based Sentiment Analysis} 
\label{sec:mt:absa}
Extending multitask learning to finer-grained, structured settings, recent works explore its potential in multimodal aspect-level analysis and generation. Yang et al.~\cite{DBLP:journals/ipm/YangNY22} propose CMMT, leveraging auxiliary tasks for aspect- and sentiment-aware intra-modal learning and a text-guided module for dynamic cross-modal interaction. Jain et al.~\cite{DBLP:conf/mm/JainSG023} reformulate multitask learning as a multimodal text-to-text generation problem via the hierarchical AbCoRD framework. Ju et al.~\cite{DBLP:conf/emnlp/JuZXLLZZ21} introduce JML to jointly perform multimodal ATE and ASC with auxiliary cross-modal relation detection. More recent methods integrate prompt learning and contrastive objectives, such as MPT~\cite{DBLP:journals/corr/abs-2310-04456}, which combines multimodal prompting with hybrid contrastive learning for low-resource robustness, and MMRBN~\cite{chen2024mmrbn}, which applies rule-based constraints to guide dynamic fusion of modality-specific emotional representations.

\paragraph{Multimodal Multi-label Emotion Recognition}
\label{sec:mt:mmer}
For multimodal multi-label emotion recognition, recent works enhance model robustness and task synergy from different perspectives. Ge et al.~\cite{DBLP:conf/www/GeJCWYG23} introduce adversarial temporal masking and parameter perturbation to improve cross-modal encoding and generalization. MER-MULTI~\cite{DBLP:conf/mrac/LianLLZTZZ23} addresses distribution mismatch by adapting label distributions between training and testing data. Additionally, Akhtar et al.~\cite{akhtar2019multi} propose a multitask framework that jointly models sentiment and emotion, leveraging their interdependence for mutual performance gains. LDDU~\cite{DBLP:conf/acl/HuangZLGYWPW25} proposes a latent emotional distribution decomposition framework with uncertainty modeling from a probabilistic latent emotion space perspective.

\subsection{Pre-trained Model}
\label{sec:pretrained_model}
In recent years, large language models (LLMs)~\cite{DBLP:conf/acl/LewisLGGMLSZ20,DBLP:journals/corr/abs-2201-05966} and multimodal pre-trained models~\cite{DBLP:conf/nips/LuBPL19,DBLP:journals/corr/abs-2002-06353,DBLP:conf/acl/RahmanHLZMMH20,DBLP:journals/corr/abs-2111-02358} have achieved remarkable progress~\cite{DBLP:journals/corr/abs-2204-04637,DBLP:journals/corr/abs-2201-05966,DBLP:conf/aaai/ZhangMWJLY22}. Compared with non-pretrained models, pre-trained models encode rich transferable knowledge~\cite{DBLP:conf/icml/HoulsbyGJMLGAG19,DBLP:conf/icml/RadfordKHRGASAM21}, which can be leveraged to enhance multimodal representation learning. 


\paragraph{Multimodal Sentiment Analysis}
\label{sec:pm:msa}
In multimodal sentiment analysis, pre-trained models are widely extended to incorporate nonverbal modalities. Rahman et al.~\cite{DBLP:conf/acl/RahmanHLZMMH20} propose MAG, which adapts BERT and XLNet by injecting visual and acoustic information into their internal representations. UniMSE~\cite{DBLP:conf/emnlp/HuLZLWL22}, built on T5~\cite{DBLP:journals/jmlr/RaffelSRLNMZLL20}, integrates nonverbal signals into a unified Transformer framework to better exploit knowledge in LLMs, while AOBERT~\cite{KIM202337} adopts a single-stream architecture to jointly model all modalities.

Other works further enhance multimodal representation learning through task-specific designs. Qian et al.~\cite{qian2023sentiment} incorporate word-level sentiment information into pre-trained representations, and TEASAL~\cite{DBLP:journals/corr/abs-2109-05522} leverages a speech-prefixed language model for cross-modal encoding. Yu et al.~\cite{DBLP:conf/ijcai/Yu019} extend BERT for target-oriented multimodal sentiment classification, while Cheng et al.~\cite{DBLP:conf/emnlp/ChengFB021} enable fine-grained cross-modal interaction via layer-wise parameter sharing and co-attention. More recently, ALMT~\cite{DBLP:journals/corr/abs-2310-05804} introduces adaptive modality learning to suppress irrelevant or conflicting information across modalities.
\paragraph{Multimodal Emotion Recognition in Conversation}
\label{sec:pm:merc}
In multimodal emotion recognition in conversation, pre-trained models serve as key backbones for representation learning. FacialMMT~\cite{DBLP:conf/acl/ZhengYXW23} adopts RoBERTa~\cite{DBLP:journals/corr/abs-1907-11692} and Swin Transformer in a two-stage framework. Qiu et al.~\cite{DBLP:conf/acl/QiuSS23} leverage VATT~\cite{DBLP:conf/nips/AkbariYQCCCG21} for multimodal encoding and alignment, while QAP~\cite{DBLP:conf/acl/LiZLZYH23} incorporates ALBERT with a quantum-inspired mechanism to adaptively learn modal priorities. Other works focus on enhancing fusion with pre-trained knowledge, such as UniMSE~\cite{DBLP:conf/emnlp/HuLZLWL22}, and modeling complex emotional dependencies, as in GraphSmile~\cite{li2024tracing}, which utilizes RoBERTa to capture inter- and intra-modal interactions in dialogue.
\begin{table*}[]
\caption{A taxonomy of multimodal affective computing covering MSA, MERC, MABSA, and MMER.}
\centering
\footnotesize
\resizebox{0.95\textwidth}{!}{\begin{tabular}{l|cccccc}
\hline
\rowcolor{green!20}
\textbf{Task}     & \textbf{Model} &\textbf{Publish Year}                  & \textbf{Venue}    & Arch.       & \textbf{Miss Modality} & \textbf{Datasets}              \\ \hline
\multirow{21}{*}{\textbf{MSA}}  & TFN~\cite{DBLP:conf/emnlp/ZadehCPCM17} &     2017  & EMNLP   &   LSTM                 & \XSolidBrush      & MOSI,MOSEI                 \\
            & MuLT~\cite{DBLP:conf/acl/TsaiBLKMS19} &  2019     & ACL      & Cross-Modal Transformer          & \XSolidBrush      & MOSI,MOSEI                 \\
            & MISA~\cite{DBLP:conf/mm/HazarikaZP20} &   2020    & ACM Multimedia  &   Transformer              & \XSolidBrush      & MOSI,MOSEI,UR\_FUNNY            \\
            & Self-MM~\cite{DBLP:conf/aaai/YuXYW21}  &   2021   & AAAI  &         BiLSTM           & \XSolidBrush      & MOSI,MOSEI,SIMS               \\
            & MTAG~\cite{DBLP:conf/naacl/YangWYZRZPM21} &  2021   & NAACL &      GNN     & \XSolidBrush      & MOSI, IEMOCAP                \\
            & MMIM~\cite{DBLP:conf/emnlp/HanCP21} &   2021     & EMNLP &           CPC         & \XSolidBrush      & MOSI,MOSEI                 \\
            & UniMSE~\cite{DBLP:conf/emnlp/HuLZLWL22} & 2022     & EMNLP &  T5, Adapter     & \XSolidBrush      & MOSI,MOSEI,IEMOCAP,MELD           \\
            & RMFN~\cite{DBLP:conf/emnlp/LiangLZM18}&     2018   & EMNLP  &      LSTHM              &    \XSolidBrush        & MOSI                    \\
            & GFML~\cite{sun2024novel}  &    2024         & ICASSP     &   CNN      & \XSolidBrush      & MOSEI,MOSI                 \\
            & TATE~\cite{DBLP:conf/sigir/ZengL022} &   2022     & SIGIR &  Transformer                & \Checkmark       & IEMOCAP,MOSI                \\ 
            &ALMT~\cite{zhang-etal-2023-learning-language}&  2023  & EMNLP& Transformer                  & \Checkmark       & CH-SIMS v2                \\ 
            &KAN-MCP~\cite{luo2025towards}& 2025 &     ACM Multimedia      & DNN,KAN      & \XSolidBrush       & CH-SIMS v2,MOSEI,MOSI               \\ 
            &KuDA~\cite{feng2024knowledge}& 2024  & EMNLP  &  Cross-modal Attention         & \XSolidBrush       & MOSI,MOSEI,CH-SIMS,CH-SIMS v2\\ 
            &DMD~\cite{li2023decoupled}& 2023 &     CVPR &   Decoupled Module          & \XSolidBrush       & MOSI,MOSEI\\ 
            &EMOE~\cite{fang2025emoe}& 2025 &     CVPR &   Expert Network          & \XSolidBrush       & MOSI,MOSEI,MIntRec\\
            &DiffuFuse~\cite{lv2025diffufuse} &    2025      & ACM Multimedia&       Dual-Stream Fusion            & \XSolidBrush      & MOSI,MOSEI  \\
            &DLF~\cite{wang2025dlf}&    2025      & AAAI &  Cross Attention                & \XSolidBrush      & MOSI,MOSEI,CH-SIMS,CH-SIMS v2  \\
            &Msamba~\cite{he2025msamba}&    2025      & AAAI& Cross-modal Hybrid Mamba                 & \XSolidBrush      & MOSI,MOSEI,CH-SIMS \\
            &DPDF-LQ~\cite{zhou2025dual}&    2025      & EMNLP&  Cross-modal Attention                 & \XSolidBrush      & MOSI,MOSEI \\
            &DEVA~\cite{wu2025enriching} &    2025      & AAAI&  Cross-modal Attention                 & \XSolidBrush      & MOSI,MOSEI,CH-SIMS \\
            \hline
\multirow{30}{*}{\textbf{MERC}} & FacialMMT-RoBERTa~\cite{zheng2023facial}      & 2023 &ACL &RoBERTa                     & \XSolidBrush      & MELD, Aff-Wild2               \\
            & MultiEMO~\cite{DBLP:conf/acl/ShiH23} &  2023      & ACL&             Multi-head Attention   & \XSolidBrush      & IEMOCAP, MELD                \\
            & MMGCN~\cite{DBLP:conf/acl/HuLZJ20}  &   2020     & ACL& GCN          & \XSolidBrush      & IEMOCAP, MELD                \\
            & MM-DFN~\cite{DBLP:conf/icassp/HuHWJM22} &  2022    & ICASSP&  GCN        & \XSolidBrush      & IEMOCAP,MELD                \\
            & EmoCaps~\cite{li2022emocaps}  &    2022       & ACL      &  Multi-head Attention       & \XSolidBrush      & IEMOCAP,MELD                \\
            & GA2MIF~\cite{DBLP:journals/taffco/LiWLZ24}  & 2024   & TAFFC& GAT  & \XSolidBrush      & IEMOCAP,MELD                \\
            & MALN~\cite{DBLP:journals/tcsv/RenHLLLL23}  & 2023   & TCSV& Transformer & \XSolidBrush      & IEMOCAP, MELD                \\
            & TDF-Net~\cite{DBLP:journals/taslp/ZhaoWSXZ23}&  2023  & TASLP& Transformer              & \Checkmark       & IEMOCAP                   \\
            & SDT~\cite{DBLP:journals/corr/abs-2310-20494} &  2024  & TMM &   Transformer              & \XSolidBrush      & IEMOCAP,MELD                \\
            & COGMEN~\cite{DBLP:journals/corr/abs-2205-02455}& 2022  & NAACL &    GNN   & \XSolidBrush      & CMU-MOSI, IEMOCAP              \\
            & Qiu et al.~\cite{DBLP:conf/acl/QiuSS23}  &  2023    & ACL&  Video-Audio-Text Transformer     & \XSolidBrush      & CMU-MOSEI                  \\
            & QAP~\cite{DBLP:conf/acl/LiZLZYH23}  &  2023      & ACL&   ALBERT                 & \XSolidBrush      & IEMOCAP,CMU-MOSEI              \\
            & MPT-HCL~\cite{DBLP:journals/corr/abs-2310-04456}&   2023& ACM Multimedia &    Transformer            & \XSolidBrush      & IEMOCAP,MELD                \\
            & CMCF-SRNet~\cite{zhang-li-2023-cross} &   2023    & ACL &   Transformer             & \XSolidBrush      & IEMOCAP,MELD                \\
            & SAMGN~\cite{DBLP:journals/tmm/ZhangCCCT24} &  2024   & TMM &  GNN            & \XSolidBrush      & IEMOCAP,MELD                \\
            & M3Net~\cite{DBLP:conf/cvpr/Chen0ZS23}   &   2023  & CVPR &    GNN          & \XSolidBrush      & IEMOCAP,MELD                \\
            & IF-MMIN~\cite{DBLP:conf/icassp/ZuoLZGL23}  & 2023   &   ICASSP & Imagination Module                    & \Checkmark       & IEMOCAP                   \\
            & MMIN~\cite{DBLP:conf/acl/ZhaoLJ20}    &  2020    & ACL &             Transformer     & \Checkmark       & IEMOCAP,MSP-IMPROV             \\
            & RLEMO~\cite{10446459}         &  2024&  ICASSP   & GCN        & \XSolidBrush      & IEMOCAP,MELD                \\
            & Yao et al.~\cite{10447720}     &    2024     & ICASSP   &  GCN       &\XSolidBrush      & IEMOCAP,MELD                \\
            & BCFN~\cite{ruan2024fusing}    &     2024 &  ICASSP  & CNN                       & \XSolidBrush      & CHERMA,CH-SIMS               \\
            & MM-RBN~\cite{chen2024mmrbn}   &    2024 &  ICASSP  &   Transformer                 & \XSolidBrush      & IEMOCAP                   \\ 
            & UniMEEC~\cite{hu2024unimeec}   &    2024      & EMNLP   &  Prompt Engineer              & \XSolidBrush      & IEMOCAP,MELD                  \\ 
            &VBH-GNN~\cite{liu2024vbh} &    2024      & ICLR            &  GNN    & \XSolidBrush      & DEAP,DREAMER   \\
            &GS-MCC~\cite{ai2025revisiting} &    2025      & AAAI   &  Graph Spectrum              & \XSolidBrush      & MELD,IEMOCAP   \\
            &Multi-to-Single~\cite{DBLP:conf/aaai/LiuCLZ25}&    2025      & AAAI &     Contrastive Learning             & \XSolidBrush      & DEAP,DREAMER   \\
            &BIG-FUSION~\cite{DBLP:conf/aaai/WangFYLLXXZX25}&    2025      & AAAI &   GNN               & \XSolidBrush      & IEMOCAP, MELD   \\
            &MSE-Adapter~\cite{DBLP:conf/aaai/YangDQ25}&    2025      & AAAI  &     Prompt Engineer          & \XSolidBrush      &MOSEI, MELD,SIMS-V2, CHERMA   \\
            \hline
\multirow{12}{*}{\textbf{MABSA}} & AoM~\cite{DBLP:conf/acl/ZhouGLYZY23} & 2023      & ACL   &   BART               & \XSolidBrush      & Twitter2015, Twitter2017          \\
            & JML~\cite{DBLP:conf/emnlp/JuZXLLZZ21}  &  2021    & EMNLP  &  ResNet               & \XSolidBrush      & TRC, Twitter2015, Twitter2017        \\
            & VLP-MABSA~\cite{DBLP:conf/acl/LingYX22}  &  2022   & ACL&   Vision-language Model     & \XSolidBrush      & Twitter2015, Twitter2017          \\
            & M2DF~\cite{DBLP:conf/emnlp/ZhaoLWOZD23}  &  2023   & EMNLP  &       CLIP    & \XSolidBrush      & Twitter2015, Twitter2017          \\
            & MIMN~\cite{DBLP:conf/aaai/XuMC19}   &  2019     & AAAI &  Memory Network              & \XSolidBrush      & ZOL                     \\
            & KNIT~\cite{DBLP:conf/icmcs/XuSX23}  &  2023      & ICME    &  Transformer           & \XSolidBrush      & Twitter2015, Twitter2017          \\ 
            &DaNet~\cite{zhu2025danet}&  2025      & ACL    &     ViT,Attention         & \XSolidBrush      & Twitter2015, Twitter2017          \\ 
            &DEQA~\cite{DBLP:conf/aaai/HanHBWL25}&  2025      & AAAI & Multiple Expert                 & \XSolidBrush      & Twitter2015, Twitter2017          \\
            &AETS~\cite{DBLP:conf/aaai/ZhuSGL025}&  2025      & AAAI  & RoBERTa, ViT                & \XSolidBrush      & Twitter2015, Twitter2017          \\
            &SDG-MLLM~\cite{DBLP:conf/mm/LiuLT0JL25}&  2025      & ACM Multimedia &     Graph-Augmented MLLM             & \XSolidBrush      & PanoSent          \\
            &MCG~\cite{DBLP:conf/ijcai/ZhuSHZCH25}&  2025      & IJCAI   &       GAT         & \XSolidBrush      & Twitter2015, Twitter2017 \\
            \hline
\multirow{9}{*}{\textbf{MMER}} & MMS2S~\cite{DBLP:conf/emnlp/ZhangJLLZZ20}  & 2020   & EMNLP &  Cross-attention   & \XSolidBrush      & MOSEI                    \\
            & MESGN~\cite{DBLP:conf/mm/JuZLZ20} &   2020      & ACM Multimedia &   Cross-modal Transformer         & \XSolidBrush      & MOSEI                    \\
            & TAILOR~\cite{DBLP:conf/aaai/ZhangCSW22}  &  2022   & AAAI &       Transformer    & \XSolidBrush      & MOSEI                    \\
            & HHMPN~\cite {DBLP:conf/aaai/ZhangJZLLZZ21}  &  2021  & AAAI &      Multi-head Attention     & \XSolidBrush      & MOSEI                    \\
            & AMP~\cite{DBLP:conf/www/GeJCWYG23}  &  2023   & WWW &   Transformer   &\XSolidBrush      & CMU-MOSEI,NEMu               \\
            & M3TR~\cite{DBLP:conf/mm/ZhaoZL21} &  2021       & AAAI &   Transformer    & \XSolidBrush      & MS-COCO,VOC 2007              \\
            & UniVA~\cite{zheng2024unimodal}    & 2024      &       ACM Multimedia &     Contrastive Learning            & \XSolidBrush      &  MOSEI, $M^{3}ED$                     \\
            &CARAT~\cite{DBLP:conf/aaai/PengCS024} & 2024      &       AAAI &     Contrastive Learning, Transformer            & \XSolidBrush      &  MOSEI, $M^{3}ED$                     \\
            &LDDU~\cite{DBLP:conf/acl/HuangZLGYWPW25} & 2025      &   ACL &     Contrastive Learning            & \XSolidBrush      &  MOSEI, $M^{3}ED$                     \\
            \hline
\end{tabular}}
\label{tab:summary_models}
\end{table*}
\paragraph{Multimodal Aspect-based Sentiment Analysis} 
\label{sec:pm:mabsa}
In multimodal aspect-based sentiment analysis (MABSA), early work by Xu et al.~\cite{xu2019multi} introduces the task and proposes MIMN to model cross- and intra-modal interactions via memory networks. Subsequent studies extend MABSA through diverse paradigms, including prompt-based generation (GMP)~\cite{DBLP:conf/acl/YangFWSWZHP23}, vision-language pre-training (VLP-MABSA)~\cite{DBLP:conf/acl/LingYX22}, and unsupervised pre-training with visual prompts~\cite{DBLP:conf/nlpcc/LiuWZ23}. Other works focus on improving aspect-aware representation learning, such as dynamic semantic modeling (DR-BERT)~\cite{DBLP:conf/acl/ZhangZZZL0C22}, multimodal interaction and alignment~\cite{zhang2021modalnet}, and query–item semantic matching~\cite{DBLP:conf/cikm/JinTLQYCZ23}. Meanwhile, modality gap reduction and fusion are addressed via image-to-text translation~\cite{wang2023image}, graph-based fusion (AD-GCFN)~\cite{wang2024adaptive}, and multi-grained fusion with self-distillation~\cite{yang2024multi}. More recent approaches further incorporate contrastive learning and unified encoder–decoder frameworks (e.g., MOCOLNet~\cite{mu2023mocolnet}) to alleviate data scarcity and enhance multimodal representation learning.

\paragraph{Multimodal Multi-label Emotion Recognition}
\label{sec:pm:mmer}
A few works on multimodal multi-label emotion recognition leverage pre-trained model to improve model performance. To our best known, TAILOR~\cite{zhang2022tailor} is a novel framework of versatile multimodal learning for multi-labeL emotion recognition, which adversarially depicts commonality and diversity among multiple modalities. TAILOR adversarially extracts private and common modality representations. Then a BERT-like transformer encoder is devised to gradually fuse these representations in a granularity descent way. 

\subsection{Enhanced Knowledge}
\label{sec:enhanced_knowlege}
External knowledge in machine learning and AI refers to information from outside the training dataset, including knowledge bases, text corpora, knowledge graphs, pre-trained models, and expert insights. Integrating this knowledge can improve performance, generalization, interpretability, and robustness to noisy or limited data \cite{li2023skier,wen2023dynamic}. 


\paragraph{Multimodal Sentiment Analysis}
\label{sec:ek:msa}
In the field of multimodal sentiment analysis, Rahmani et al.~\cite{DBLP:journals/kbs/RahmaniHZKKH23} construct an adaptive tree by hierarchically partitioning users and employ attention-based fusion to propagate cognitive-oriented knowledge within the tree. TETFN~\cite{DBLP:journals/pr/WangGTLHL23}, a novel text-enhanced Transformer fusion network, learns text-driven pairwise cross-modal mappings to obtain effective unified multimodal representations. Zhu et al.~\cite{DBLP:journals/inffus/ZhuCZSLLC23} introduce the Sentiment Knowledge Enhanced Attention Fusion Network (SKEAFN), an end-to-end fusion framework that incorporates external sentiment knowledge to improve multimodal representation learning. Similarly, Chen et al.~\cite{chen2020swafn} integrate sentiment word knowledge into the fusion network to guide the learning of joint multimodal feature representations.

\paragraph{Multimodal Emotion Recognition in Conversation}
\label{sec:ek:merc}
In the field of multimodal emotion recognition in conversation, Fu et al.~\cite{DBLP:journals/ieeemm/FuOWGSLD22} integrate context modeling, knowledge enrichment, and multimodal (text and audio) learning within a GCN-based architecture. Li et al.~\cite{DBLP:conf/cvpr/LiW023} propose a decoupled multimodal distillation (DMD) framework that enables flexible cross-modal knowledge transfer, enhancing the discriminative capacity of each modality. Sun et al.~\cite{DBLP:conf/ccis/SunHTZS23} introduce a multimodal fusion transformer network grounded in rough set theory, which promotes feature interaction and guidance via rough set cross-attention. Wang et al.~\cite{DBLP:journals/corr/abs-2309-02106} design a label-guided attentive fusion module to integrate label-aware text and speech representations, learning label-enhanced embeddings for each utterance through label-token and label-frame interactions.

\paragraph{Multimodal Aspect-based Sentiment Analysis} 
\label{sec:ek:mabsa}
In multimodal aspect-based sentiment analysis, Xu et al.~\cite{DBLP:conf/icmcs/XuSX23} incorporate external knowledge, including textual syntax and cross-modal relevance, into the Transformer layer, using a knowledge-induced matrix to suppress irrelevant connections. Yang et al.~\cite{DBLP:conf/emnlp/YangZ022} distill visual emotional cues and align them with textual content for selective fusion with the target aspect. CoolNet~\cite{xiao2023cross}, a cross-modal fine-grained alignment and fusion network, enhances visual-language integration by converting an image into a caption and graph structure, dynamically aligning semantic and syntactic information from both the input text and generated caption, while modeling object-level visual features. To further strengthen image semantics, Yu et al.~\cite{yu2022hierarchical} detect salient objects via a pre-trained Faster R-CNN, represent each object with hidden features and associated semantic concepts, and employ an Aspect-Guided Attention layer to capture concept relevance guided by target aspects.

\paragraph{{Multimodal Multi-label Emotion Recognition}}
\label{sec:ek:mmer}
In multimodal multi-label emotion recognition (MMER), Zheng et al.~\cite{zheng2024unimodal} represent each emotion in the valence-arousal (VA) space to capture inter-emotion correlations and introduce a unimodal VA-driven contrastive learning algorithm. CARAT~\cite{peng2024carat} employs contrastive feature reconstruction and aggregation to model fine-grained modality-to-label dependencies via modal-separated, label-specific features. Zhao et al.~\cite{zhao2021m3tr} propose M3TR, a multimodal Transformer that unifies high-level semantics, visual structures, and label co-occurrences across modalities. Li et al.~\cite{li2024magdra} present MAGDRA, a multimodal attention graph network with dynamic routing-by-agreement, fusing heterogeneous graph data and capturing cross-modal and temporal interactions without pre-alignment. Zhang et al.~\cite{zhang2021multi} develop HHMPN, a heterogeneous hierarchical message passing network that simultaneously models feature-to-label, label-to-label, and modality-to-label dependencies.

\subsection{Contextual Information}
\label{sec:context_information}

Context encompasses surrounding textual, visual, and acoustic information (e.g., words, dialogue history, or document-level content) that shapes meaning and improves prediction accuracy in tasks such as dialogue and sentiment analysis.


\paragraph{Multimodal Sentiment Analysis}
\label{sec:ci:msa}
In the community of multimodal sentiment analysis, Chauhan et al.~\cite{DBLP:conf/emnlp/ChauhanAEB19} employ a context-aware attention module to learn intra-modality interaction among particating modalities through encoder-decoder structure. Multimodal context integrate the unimodal context, Poria et al.~\cite{DBLP:conf/icdm/PoriaCHMZM17} propose a recurrent model with multi-level multiple attentions to capture contextual information among utterances, and design a recurrent model to capture contextual information among utterances and introduced attention-based networks for improving both context learning and dynamic feature fusion. Huang et al.~\cite{DBLP:journals/taslp/HuangQTX23} propose a novel context-based adaptive multimodal fusion network (CAMFNet) for consecutive frame-level sentiment prediction. Li et al.~\cite{DBLP:journals/pami/LiSZT22} propose a spatial context extraction block to explore the spatial context by calculating the relationships between feature maps and the higher-level semantic representation in images.

\paragraph{Multimodal Emotion Recognition in Conversation}
\label{sec:ci:merc}
In the realm of research concerning multimodal emotion recognition in conversation, Hu et al.~\cite{DBLP:conf/acl/HuLZJ20} make use of multimodal dependencies effectively, and leverages speaker information to model inter-speaker and intra-speaker dependency. 
Zhang et al.~\cite{zhang-li-2023-cross} propose a cross-modality context fusion and semantic refinement network (CMCF-SRNet) to solve the limitation of insufficient semantic relationship information between utterances. Zhang et al.~\cite{DBLP:journals/tmm/ZhangCCCT24} construct multiple modality-specific graphs to model the heterogeneity of the multimodal context. Chen et al.~\cite{DBLP:conf/cvpr/Chen0ZS23} propose a GNN-based model that explores multivariate relationships and captures the varying importance of emotion discrepancy and commonality by valuing multi-frequency signals. Zhang et al.~\cite{DBLP:journals/tois/ZhangJWZZLHJSQ24} propose a multimodal, multi-task interactive graph attention network, termed M3GAT, to simultaneously solve conversational context dependency, multimodal interaction, and multi-task correlation in a unified framework. RL-EMO~\cite{10446459} is a novel reinforcement Learning framework for the multimodal emotion recognition task, which combines reinforcement learning (RL) module to model context at both the semantic and emotional levels respectively. Yao et al.~\cite{10447720} propose a speaker-centric multimodal fusion network for emotion recognition in a conversation, to model intra-modal feature fusion and speaker-centric cross-modal feature fusion. 

\paragraph{Multimodal Aspect-based Sentiment Analysis} 
\label{sec:ci:absa}
In the stduy of multimodal aspect-based sentiment analysis, Yu et al.~\cite{DBLP:conf/ijcnlp/YuWYZ22} propose an unsupervised approach which minimizes the Wasserstein distance between both modalities, forcing both encoders to produce more appropriate representations for the final extraction. Xu et al.~\cite{DBLP:conf/iccc2/XuLW22} design and construct a multimodal Chinese product review dataset (MCPR) to support the research of MABSA. Anschutz et al.~\cite{DBLP:conf/semco/AnschutzEG23} report the results of an empirical study on how semantic computing can provide insights into user-generated content for domain experts. In addition, this work discussed different image-based aspect retrieval and aspect-based sentiment analysis approaches to handle and structure large datasets. Zhao et al.~\cite{DBLP:journals/corr/abs-2310-14605} borrow the idea of Curriculum Learning and propose a multi-grained multi-curriculum denoising Framework (M2DF) to adjust the order of training data, so as to obtain more contextual information. Zhou et al.~\cite{DBLP:conf/acl/ZhouGLYZY23} propose an aspect-oriented method (AoM) to detect aspect-relevant semantic and sentiment information. Specifically, an aspect-aware attention module is designed to simultaneously select textual tokens and image blocks that are semantically related to the aspects. Zhao et al.~\cite{zhao2023fusion} propose a fusion with GCN and SE ResNeXt Network (FGSN), which constructs a graph convolution network on the dependency tree of sentences to obtain the context representation and aspects words representation by using syntactic information and word dependency.

\paragraph{{Multimodal Multi-label Emotion Recognition}}
\label{sec:ci:mmer}
MMS2S~\cite{DBLP:conf/emnlp/ZhangJLLZZ20} is a multimodal sequence-to-set approach to effectively model label dependence and modality dependence. MESGN~\cite{DBLP:conf/mm/JuZLZ20} firstly proposes this task, which simultaneously models the modality-to-label and label-to-label dependencies. Many works consider the dependencies of multi-label based on the characteristics of co-occurrence labels. Zhao et al.~\cite{DBLP:conf/acl/ZhaoZ0LJW022} propose a general multimodal dialogue-aware interaction framework, named by MDI, to model the impacts of dialogue context on emotion recognition.

\begin{table*}[]
\centering
\caption{List of multimodal affective computing datasets. T, A, and V denote text, audio, and visual modalities, respectively. “Emotion” indicates samples labeled with emotion categories, and “Sentiment” indicates samples labeled with sentiment polarity.}
\resizebox{1.0\textwidth}{!}{\begin{tabular}{l|ccccccc}
\toprule
\rowcolor{green!20}
{\bf Task}         & {\bf Dataset}  &{\bf Modalities} &{\bf Source} &{\bf Emotion} & {\bf Sentiment}& {\bf Language} &{\bf Datasize} \\
\midrule
\multirow{7}{*}{\textbf{MSA}} & CMU-MOSI~\cite{DBLP:journals/expert/ZadehZPM16}   &  T,A,V     & Video Blog, YouTube & \XSolidBrush & \Checkmark &  English   & 2,199 \\
           & CMU-MOSEI~\cite{DBLP:conf/acl/MorencyCPLZ18}   &  T,A,V     & YouTube & \Checkmark & \Checkmark &   English  &  22,856   \\
           & CH-SIMS~\cite{yu2020ch}  &  T,A,V    &  Movies,TVs  & \XSolidBrush & \Checkmark &  Chinese   & 2,281 \\
           & CMU-MOSEAS~\cite{zadeh2020cmu}& T,A,V & YouTube &  \Checkmark & \Checkmark  & Spanish,Portuguese,German,French & 4,000    \\
           &Mintrec~\cite{zhang2022mintrec}& T,A,V & TV series Superstore &  \Checkmark & \XSolidBrush  & Spanish,Portuguese,German,French & 2,224 \\
           & ICT-MMMO~\cite{DBLP:journals/expert/WollmerWKSSSM13}  &  T,A,V     &  reviews   &   -   & -    & - &-\\
           & YouTube~\cite{DBLP:conf/icmi/MorencyMD11}  & T,A,V &  YouTube & - & - & English & 300  \\
          &CH-SIMS v2.0\cite{liu2022make}&  T,A,V    &  TV series,Shows,Movies & \XSolidBrush & \Checkmark  &  Chinese    &14,402 
          \\
\midrule
\multirow{17}{*}{\textbf{MERC}} & MELD~\cite{DBLP:conf/acl/PoriaHMNCM19}    &   T,A,V    & Friends TV    & \Checkmark &  \XSolidBrush& English  & 13,707     \\
           & IEMOCAP~\cite{DBLP:journals/lre/BussoBLKMKCLN08}   &  T,A,V    &  Act  & \Checkmark & \XSolidBrush &  English    &7,532   \\
        &DFEW~\cite{jiang2020dfew}   &  T,V    &  Movies  & \Checkmark & \XSolidBrush &  English    & 12,059   \\
        &MER2023 ~\cite{lian2024merbench}   &  A,V,T    &  Movies  & \Checkmark & \XSolidBrush &  Chinese    & 5,030   \\
           & HED~\cite{DBLP:journals/corr/abs-2211-15425}  &  T,V     & Movies, TVs  & \Checkmark     & \XSolidBrush  & English &17,441\\
           & RML~\cite{DBLP:journals/tmm/WangG08}&   A,V   &  Video   &  \Checkmark   & \XSolidBrush
           & English, Mandarin, Urdu, Punjabi, Persian, and Italian & -\\
           & BAUM-1~\cite{DBLP:journals/taffco/ZhalehpourOAE17}   &  A,V    &  Data collection  &   \Checkmark  & \XSolidBrush  & Turkish & 1,184 \\
           & MAHNOB-HCI\cite{DBLP:journals/taffco/SoleymaniLPP12}  & V, EEG & Data collection & \Checkmark & \XSolidBrush & - & -\\
           & Deap~\cite{koelstra2011deap} & EEG      &  Act, Data collection   &   \Checkmark  & \Checkmark &Physiological signal & -\\
           & MuSe-CaR~\cite{DBLP:journals/taffco/StappenBSS23}  &  T,A,V     & Car Reviews,YouTube   & \Checkmark & \Checkmark & English &\\
           & CHEAVD~\cite{li2018mec}  &  A,V     &  Movies,TVs  & - & - &  Mandarin   & 7,030  \\
           &MSP-IMPROV~\cite{7374697}&  T,A,V     &  Act & \Checkmark & \XSolidBrush  &  English  &8,438  \\
          &MEISD~\cite{firdaus2020meisd}&  T,A,V     &  TVs & \Checkmark & \Checkmark  &  English   & - \\
          &MESD~\cite{DBLP:conf/naacl/JiaHZU0L22}&  T,A,V     &  TVs & \Checkmark & \Checkmark  &  English    &9,190  \\
          &Ulm-TSST~\cite{DBLP:conf/mm/StappenBCSSMCZS21}&  A,V,EEG    &  Job Interviews & \Checkmark & \Checkmark  &  English    &-  
          \\
          &CHERMA~\cite{sun2023layer}&  T,A,V    &  TV Series,Shows,Movies & \Checkmark & \Checkmark  &  Chinese   &28,717 \\
          &AMIGOS~\cite{DBLP:journals/taffco/CorreaASP21}&  EEG, ECG, GSR  & Data collection & \XSolidBrush& \Checkmark  &  - &- 
          \\
\midrule
\multirow{7}{*}{\textbf{MABSA}} & Twitter2015~\cite{DBLP:conf/aaai/0001FLH18}   &  T,V     & Twitter  & \XSolidBrush & \Checkmark  &  English    & - \\
            & Twitter2017~\cite{DBLP:conf/aaai/0001FLH18}   &  T,V     &  Twitter  & \XSolidBrush & \Checkmark  &  English   & -   \\
           & MCPR~\cite{xu2022mcpr}   &  T,V    &  Product Reviews & \XSolidBrush & \Checkmark  &  Chinese   & 15,000    \\
           & Multi-ZOL~\cite{xu2019multi}  &  T,V    & Product Reviews  & \XSolidBrush & \Checkmark  & Chinese    & 5,288    \\
           & MACSA~\cite{DBLP:journals/corr/abs-2206-13969} &  T,V     & Hotel Service Reviews  & \XSolidBrush & \Checkmark  &   Chinese  &  21,108   \\
           &{MASAD}~\cite{zhou2021masad} &T,V &Visual Sentiment Ontology datasets & \XSolidBrush &\Checkmark &English &38,532
           \\
           &{PanoSent}~\cite{luo2024panosent} &T,A,V &Social Media & \XSolidBrush &\Checkmark &English,Chinese, Spanish &10,000
           \\
\midrule
\multirow{2}{*}{\textbf{MMER}} & CMU-MOSEI~\cite{DBLP:conf/acl/MorencyCPLZ18}    &  T,A,V     & YouTube  & \Checkmark & \Checkmark &  English   &22,856    \\
           & $M^{3}ED$~\cite{DBLP:conf/acl/ZhaoZ0LJW022}   &   T,A,V    &  56 TVs  & \Checkmark & \XSolidBrush  &  Mandarin  & 24,449     \\
\midrule
\multirow{6}{*}{\textbf{Emotion Reasoning}} & CA-MER~\cite{han2025benchmarking}   &  T,A,V     & MER  & \Checkmark & \XSolidBrush &  Chinese   &1,500    \\
           &MTMEUR~\cite{hu2025beyond}   &   T,A,V    & Websites: Pexels,Mixkit  & \Checkmark & \XSolidBrush &  English  & 5,101     \\
           &AffectGPT~\cite{lian2025affectgpt}   &   T,A,V    &  MER 2023  & \Checkmark & \XSolidBrush &  Mandarin  & 115,595     \\
           &HitEmotion~\cite{luo2026unveiling}   &   V, Image    &  24 Diverse Datasets & \Checkmark & \XSolidBrush & English  & 20,114     \\
           &EmoReAIM~\cite{chaubeyavere}   &  T,V     &  DFEW~\cite{jiang2020dfew} & \Checkmark & \XSolidBrush &  Mandarin  & 4,000     \\
           &EMER~\cite{lian2023explainable} &  T,V,A     &  MER & \Checkmark & \XSolidBrush &  Mandarin  & 332     \\
\bottomrule
\end{tabular}}
\label{tab:dataset_details}
\end{table*}

\subsection{Multimodal Emotional Reasoning and Understanding}
\label{sec:reasoning}

Multimodal emotional understanding integrates heterogeneous signals (e.g., text, audio, and vision) to infer affect. Early work focuses on MER and MSA via feature fusion, while recent advances shift toward emotional reasoning, modeling cross-modal interactions, temporal dynamics, and causal structures. Emerging benchmarks, such as MTMEUR \cite{hu2025beyond} and CA-MER \cite{han2025benchmarking}, along with multimodal foundation models, further frame this task as a reasoning problem requiring explanation and generalization. For example, MTMEUR \cite{hu2025beyond} and CA-MER \cite{han2025benchmarking} introduce benchmarks for multimodal emotion reasoning, focusing on multi-turn interactions and cross-modal conflicts. With the rise of multimodal foundation models, emotional understanding is increasingly framed as a reasoning task emphasizing interpretability and generalization. Representative efforts include EmoSym \cite{zhu2025emosym}, which leverages latent CoT reasoning, DEEMO \cite{li2025deemo} for identity-free understanding, and EmoReAlM with AVEm-DPO \cite{chaubeyavere} for improving cue–emotion reasoning and mitigating hallucinations.

\section{Datasets of Multimodal Affective Computing}
\label{sec:datasets}
In this section, we summarize benchmark datasets for MSA, MERC, MABSA, and MMER. Detailed statistics are provided in Table~\ref{tab:dataset_details}.

\paragraph{Multimodal Sentiment Analysis}
\begin{itemize}
\item {\bf MOSI}~\cite{DBLP:journals/expert/ZadehZPM16}: 2,199 video utterances annotated with sentiment scores in [-3, +3].
\item {\bf MOSEI}~\cite{DBLP:conf/acl/MorencyCPLZ18}: a large-scale extension of MOSI with 22,856 clips, annotated with both sentiment and multi-label emotions.
\item {\bf CH-SIMS}~\cite{yu2020ch}: a Chinese dataset with 2,281 in-the-wild video segments, providing both multimodal and unimodal annotations.
\item {\bf CH-SIMS v2.0}~\cite{liu2022make}: an extended version with more samples and modality-specific as well as overall sentiment annotations.
\item {\bf CMU-MOSEAS}~\cite{zadeh2020cmu}: a multilingual dataset (Spanish, Portuguese, German, French) with ~4,000 samples collected from YouTube.
\item {\bf ICT-MMMO}~\cite{DBLP:journals/expert/WollmerWKSSSM13}: real-world review videos with diverse expressions and recording conditions.
\item {\bf YouTube}~\cite{DBLP:conf/icmi/MorencyMD11}: 47 videos with 3–11 utterances each, extracted from online content.
\end{itemize}

\paragraph{Multimodal Emotion Recognition in Conversation}
\begin{itemize}
  \item {\bf MELD}~\cite{DBLP:conf/acl/PoriaHMNCM19}: 13,707 multi-party dialogue clips annotated with Ekman’s six basic emotions.  
  \item {\bf IEMOCAP}~\cite{DBLP:journals/lre/BussoBLKMKCLN08}: 7,532 conversational clips with standard emotion labels. 
  \item {\bf DFEW}~\cite{jiang2020dfew}: an in-the-wild video-based facial expression dataset with diverse scenes and emotion annotations.
  \item {\bf MER2023}~\cite{lian2024merbench}: a unified evaluation benchmark for multimodal emotion recognition.
  \item {\bf HED}~\cite{DBLP:journals/corr/abs-2211-15425}: large-scale multimodal dataset with emotion-aligned face, body, and text under a psychological paradigm.  
  \item {\bf RML}~\cite{DBLP:journals/tmm/WangG08}: 500 multilingual video samples (6 languages) annotated with six emotions.  
  \item {\bf BAUM-1}~\cite{DBLP:journals/taffco/ZhalehpourOAE17}: two subsets with basic and extended emotion categories, including cognitive states.  
  \item {\bf MAHNOB-HCI}~\cite{DBLP:journals/taffco/SoleymaniLPP12}: 527 recordings with facial videos and physiological signals (e.g., EEG, ECG).  
  \item {\bf DEAP}~\cite{koelstra2011deap}: 32 participants rating 40 music videos on valence, arousal, dominance, and related factors.  
  \item {\bf MuSe-CaR}~\cite{DBLP:journals/taffco/StappenBSS23}: multimodal dataset for emotion, engagement, and trustworthiness recognition.  
  \item {\bf CHEAVD 2.0}~\cite{li2018mec}: Chinese audiovisual dataset with real-world noise conditions.  
  \item {\bf MSP-IMPROV}~\cite{7374697}: dyadic interaction dataset for audiovisual emotion perception.  
  \item {\bf MEISD}~\cite{firdaus2020meisd}: large-scale multimodal dialogue dataset with emotion, intensity, and sentiment labels.  
  \item {\bf MESD}~\cite{DBLP:conf/naacl/JiaHZU0L22}: multimodal multi-task dataset (sentiment, emotion, desire) with 9,190 text-image pairs.  
  \item {\bf Ulm-TSST}~\cite{DBLP:conf/mm/StappenBCSSMCZS21}: stress-induced multimodal dataset following the TSST protocol.  
  \item {\bf CHERMA}~\cite{sun2023layer}: dataset with both unimodal and multimodal annotations from diverse media sources.  
  \item {\bf AMIGOS}~\cite{DBLP:journals/taffco/CorreaASP21}: multimodal dataset with individual and group settings, including physiological signals (EEG, ECG, GSR).  
\end{itemize}

\paragraph{Multimodal Aspect-based Sentiment Analysis}

\begin{itemize}
  \item {\bf Twitter2015} and {\bf Twitter2017}~\cite{DBLP:conf/aaai/0001FLH18,DBLP:conf/acl/JiZCLN18}: multimodal Twitter datasets with aspect-level sentiment annotations.  
  \item {\bf MCPR}~\cite{xu2022mcpr}: 2,719 text-image pairs with 610 aspects, collected from Chinese e-commerce reviews; the first Chinese MABSA dataset.  
  \item {\bf Multi-ZOL}~\cite{xu2019multi}: 5,288 multimodal samples of mobile phone reviews, annotated with aspect-level sentiment intensity (1–10).  
  \item {\bf MACSA}~\cite{DBLP:journals/corr/abs-2206-13969}: 21K+ text-image pairs with fine-grained cross-modal annotations aligned via aspect categories.  
  \item {\bf MASAD}~\cite{zhou2021masad}: 38,532 samples across 7 domains with 57 predefined aspects, derived from a visual sentiment dataset~\cite{borth2013large}.  
  \item {\bf PanoSent}~\cite{luo2024panosent}: large-scale (10K dialogues), multilingual, multi-scenario dataset with multimodal and fine-grained sentiment annotations.  
\end{itemize}

\paragraph{Multimodal Multi-label Emotion Recognition}
\begin{itemize}
  \item {\bf CMU-MOSEI}~\cite{DBLP:conf/acl/MorencyCPLZ18}: 22,856 multimodal video clips (text, audio, visual) annotated with six emotion categories.  
  \item $\textbf{M}^{3}\textbf{ED}$~\cite{DBLP:conf/acl/ZhaoZ0LJW022}: Chinese multimodal dialogue dataset with 9,082 turns and 24,449 utterances, labeled with seven emotion categories.  
\end{itemize}

\section{Evaluation Metrics}
\label{sec:evaluation}
We summarize common evaluation metrics for multimodal affective computing tasks.

\paragraph{\textbf{Multimodal Sentiment Analysis}} Evaluation typically includes mean absolute error (MAE), Pearson correlation (Corr), seven-class accuracy (ACC-7), binary accuracy (ACC-2), and F1 scores for both positive/negative and non-negative/negative classifications.

\paragraph{\textbf{Multimodal Emotion Recognition in Conversations}} Models are evaluated using accuracy (ACC) and weighted F1 (WF1). Due to label imbalance, per-category ACC and F1 are also reported to assess performance across different emotions.

\paragraph{\textbf{Multimodal Aspect-based Sentiment Analysis}} For multimodal aspect term extraction (MATE) and joint multimodal aspect sentiment analysis (JMASA), precision (P), recall (R), and micro-F1 are used. For multimodal aspect sentiment classification (MASC), accuracy (ACC) and macro-F1 are adopted.

\paragraph{\textbf{Multimodal Multi-label Emotion Recognition}} Evaluation typically uses accuracy (ACC), micro-F1, precision (P), and recall (R).

\section{Discuss}
\label{sec:discuss}
In this section, We review multimodal affective computing based on facial, acoustic, and physiological signals, as well as emotion causes, and analyze approaches across tasks to highlight their commonalities and differences.

\subsection{Other Multimodal Affective Computing}
\paragraph{Multimodal Affective Computing Based on Facial Expression Recognition} 
Facial expression recognition has progressed from static methods—using single-frame images with traditional features like LBP and Gabor filters—to deep learning-based CNNs that improve accuracy~\cite{zhang2021relative,farzaneh2021facial}. Static approaches, however, struggle to capture temporal dynamics, motivating dynamic methods that extract local–global features and key segments for robustness and uncertainty-aware inference~\cite{wang2023rethinking}. Recent advances in dynamic facial expression recognition (DFER) leverage multimodal data, parameter-efficient fine-tuning of large pre-trained models, and generative techniques for expression normalization to handle in-the-wild noise. Emerging evidential deep learning (EDL) further enables uncertainty quantification in latent spaces, enhancing interpretability and demonstrating efficacy in zero-shot learning, multi-view classification, video understanding, and multimodal tasks ~\cite{Gao2023TAL,ma2024beyond,Gao2023vectorized}.

\paragraph{Multimodal Affective Computing Based on Acoustic Signal}
Single-sentence single-task models are the most common in speech emotion recognition. Representative approaches include: CNN on temporal features (MFSC) with max pooling~\cite{DBLP:conf/icassp/AldenehP17}; dual-kernel CNN on spectrograms with attention pooling~\cite{DBLP:conf/interspeech/LiSMGD18}; end-to-end CNN on raw waveforms~\cite{DBLP:conf/icassp/TrigeorgisRBMNS16}; Bi-LSTM with attention~\cite{DBLP:conf/icassp/MirsamadiBZ17}; CNN and attention for temporal/spatial spectrum features~\cite{DBLP:conf/interspeech/ZhaoZZWZL18}; dual-channel CNN+RNN on spectrograms and HSFS features~\cite{DBLP:conf/interspeech/LuoZH18}.

\paragraph{Multimodal Affective Computing Based on Physiological Signals}

EEG-based emotion recognition (EER) is a promising direction in affective computing, with nearly 1,000 publications since 2010~\cite{quan2021physiological}. Multimodal EER (EMER) methods~\cite{DBLP:conf/embc/ChaparroGSQLV18,DBLP:journals/cin/HuangYLP17,DBLP:conf/icmlsc/ZhuLY20,DBLP:conf/iconip/TangLZL17,DBLP:journals/bspc/TanSDSC21} exploit complementarity between EEG and other physiological signals. For instance, Vazquez et al.~\cite{DBLP:conf/acii/Vazquez-Rodriguez22} show Transformers are effective for physiological-based multimodal emotion recognition.

\paragraph{Multimodal Affective Computing Based on Emotion Cause}
Understanding emotion causes is essential, leading to emotion-cause pair extraction (ECPE). Text-based ECPE has advanced significantly~\cite{DBLP:conf/emnlp/HuLZ21,DBLP:journals/kbs/LiZGY23}. Extending this, multimodal ECPE (MECPE)~\cite{9926166} extracts emotion-cause pairs from multimodal data using joint training with emotion and cause detection subtasks. Recent work, such as HiLo~\cite{10.1145/3689646}, leverages attention-based cross-modality and cross-utterance interactions for accurate cause extraction.

\subsection{Consistency among Multimodal Affective Computing}
We categorize the multimodal affective computing tasks into several key areas: multimodal alignment and fusion, multi-task learning, pre-trained models, enhanced knowledge, and contextual information. To ensure clarity, we discuss the consistencies across these aspects.

\paragraph{Multimodal alignment and fusion} MSA, MERC, MABSA, and MMER all require extracting features from at least two modalities and fusing them into a unified representation. Key challenges are modal alignment and fusion. Unlike vision-dominated tasks (e.g., image captioning), multimodal affective computing emphasizes language over vision~\cite{DBLP:conf/emnlp/HuLZLWL22,DBLP:conf/emnlp/ZengM021}.

\paragraph{Pre-trained model} Multimodal affective computing tasks typically use pre-trained models as backbones to encode raw modalities, followed by fine-tuning. For instance, UniMSE~\cite{DBLP:conf/emnlp/HuLZLWL22} adopts T5, while GMP~\cite{DBLP:conf/acl/YangFWSWZHP23} uses BART, transferring general knowledge from language models to affective computing.

\paragraph{Enhanced knowledge} Commonsense knowledge—facts and judgments about the world—is essential for machines to understand human emotions and causes. Researchers enhance affective computing by integrating external sources such as sentiment lexicons~\cite{DBLP:conf/emnlp/FanYDGBYXM19}, English knowledge bases~\cite{DBLP:conf/aaai/SpeerCH17,senticnet,DBLP:conf/ijcai/ZhangKSR20,DBLP:conf/acl/BosselutRSMCC19,DBLP:conf/aaai/SapBABLRRSC19}, and Chinese knowledge bases~\cite{DBLP:journals/corr/abs-2204-02549}.

\paragraph{Contextual information} Affective computing requires understanding context. In MERC, context includes the full dialogue around an utterance; in MABSA, it is the complete sentence containing opinions. Researchers model context via hierarchical approaches~\cite{DBLP:conf/naacl/YangYDHSH16,DBLP:conf/acl/ZhouMLXDZXL20}, self-attention~\cite{DBLP:conf/nips/VaswaniSPUJGKP17}, and graph-based dependency~\cite{DBLP:journals/kbs/HuLZ21,DBLP:conf/emnlp/GhosalMPCG19}. Non-verbal cues (facial expressions, vocal tone) further enrich understanding.

\subsection{Difference among Multimodal Affective Computing}
We differentiate multimodal affective computing tasks by downstream objective, sentiment granularity, and application context. \textit{MSA} is regression (sentiment intensity). \textit{MERC} is multi-class classification (emotion categories). \textit{MMER} is multilabel emotion recognition. \textit{MABSA} is information extraction (aspect–opinion polarity). \textit{\textbf{Granularity}}: MERC/MECPE focus on utterance/speaker level within dialogues; MSA/MMER on sentence level within documents; MABSA on aspect level within comments. Some studies bridge granularities via unified frameworks~\cite{DBLP:conf/acl/ZhangZWZ22,DBLP:conf/emnlp/ZengM021}. \textit{\textbf{Context}} varies accordingly: MABSA uses comment–image–aspect descriptions; MERC uses full dialogue and speaker info. \textit{\textbf{Applications}}: MSA/MMER/MABSA serve opinion mining and user experience analysis; MERC/MECPE enable empathetic dialogue agents. Despite task-specific designs, unified frameworks for multi-granularity emotion analysis are emerging~\cite{DBLP:conf/emnlp/HuLZLWL22,DBLP:conf/naacl/AkhtarCGPEB19}.

\section{Future work}
\label{sec:future_work}
We outline future directions, including multimodal transfer learning, task unification, knowledge distillation, and underexplored modalities.

\paragraph{Unification of Multimodal Affective Computing tasks}
Recent work unifies related tasks into single frameworks (e.g., T5 for NLP)~\cite{DBLP:journals/jmlr/RaffelSRLNMZLL20}, boosting performance and generalization~\cite{DBLP:journals/corr/abs-2111-02358,DBLP:conf/emnlp/ChengYZS21}. This suggests potential for unifying multimodal affective computing across: (1) granularity levels~\cite{DBLP:conf/acl/ZhangZWZ22}, (2) language–vision–audio modalities~\cite{DBLP:journals/corr/abs-2208-10442}, and (3) emotion-cause integration.

\paragraph{Transfer Learning with External Knowledge Distill}
Incorporating external knowledge (e.g., sentiment lexicons, commonsense) is crucial for understanding emotional expressions within social and cultural contexts~\cite{hershcovich-etal-2022-challenges}, as emotion expression and perception vary across cultures in both text and face-to-face communication~\cite{hareli2015cross}—a key consideration for cross-cultural sentiment analysis.

\paragraph{Affective Computing with Less-studied Modalities}
Language, visual, and auditory signals have long been central to multimodal affective computing. Recently, less-studied modalities such as haptic and ECG are gaining attention~\cite{zhu2024neurosymbolic}. Haptic signals, which convey sensory and emotional attributes through touch, enhance user engagement in gaming, VR, and mobile apps~\cite{obrist2013talking}—and are poised to complement established modalities in advancing the field.

\section{Conclusion}
\label{sec:conclusion}
Multimodal affective computing has become a pivotal research area in artificial intelligence, achieving significant advances in emotion understanding and interpretation. This survey presents a comprehensive overview of its associated tasks, including research background, definitions, related work, technical approaches, benchmark datasets, and evaluation metrics. We categorize multimodal affective computing across MSA, MERC, MABSA, and MMER tasks into four main directions: multi-task learning, pre-trained modalities, knowledge enhancement, and context modeling. Furthermore, we analyze the consistencies and differences among these tasks, highlight the inherent challenges in multimodal sentiment analysis, and discuss promising directions for future research.
 
\bibliography{manuscript} 

@conference{senticnet,
	Author = {Cambria, Erik and Zhang, Xulang and Mao, Rui and Chen, Melvin and Kwok, Kenneth},
	Booktitle = {{International Conference on Human-Computer Interaction (HCII)}},
	Title = {{SenticNet} 8: Fusing Emotion {AI} and Commonsense {AI} for Interpretable, Trustworthy, and Explainable Affective Computing},
	Year = {2024}}

@inproceedings{DBLP:conf/emnlp/HanCP21,
	Author = {Wei Han and Hui Chen and Soujanya Poria},
	Booktitle = {Proceedings of the 2021 Conference on Empirical Methods in Natural Language Processing, {EMNLP} 2021, Virtual Event / Punta Cana, Dominican Republic, 7-11 November, 2021},
	Editor = {Marie{-}Francine Moens and Xuanjing Huang and Lucia Specia and Scott Wen{-}tau Yih},
	Pages = {9180--9192},
	Title = {Improving Multimodal Fusion with Hierarchical Mutual Information Maximization for Multimodal Sentiment Analysis}}

@inproceedings{DBLP:conf/aaai/Mai0X20,
	Author = {Sijie Mai and Haifeng Hu and Songlong Xing},
	Booktitle = {The Thirty-Fourth {AAAI} Conference on Artificial Intelligence, {AAAI} 2020, The Thirty-Second Innovative Applications of Artificial Intelligence Conference, {IAAI} 2020, The Tenth {AAAI} Symposium on Educational Advances in Artificial Intelligence, {EAAI} 2020, New York, NY, USA, February 7-12, 2020},
	Pages = {164--172},
	Publisher = {{AAAI} Press},
	Title = {Modality to Modality Translation: An Adversarial Representation Learning and Graph Fusion Network for Multimodal Fusion},
	Year = {2020}}

@inproceedings{DBLP:conf/mm/HazarikaZP20,
	Author = {Devamanyu Hazarika and Roger Zimmermann and Soujanya Poria},
	Booktitle = {{MM} '20: The 28th {ACM} International Conference on Multimedia, Virtual Event / Seattle, WA, USA, October 12-16, 2020},
	Editor = {Chang Wen Chen and Rita Cucchiara and Xian{-}Sheng Hua and Guo{-}Jun Qi and Elisa Ricci and Zhengyou Zhang and Roger Zimmermann},
	Pages = {1122--1131},
	Publisher = {{ACM}},
	Title = {{MISA:} Modality-Invariant and -Specific Representations for Multimodal Sentiment Analysis},
	Year = {2020}}

@inproceedings{DBLP:conf/aaai/YuXYW21,
	Author = {Wenmeng Yu and Hua Xu and Ziqi Yuan and Jiele Wu},
	Booktitle = {Thirty-Fifth {AAAI} Conference on Artificial Intelligence, {AAAI} 2021, Thirty-Third Conference on Innovative Applications of Artificial Intelligence, {IAAI} 2021, The Eleventh Symposium on Educational Advances in Artificial Intelligence, {EAAI} 2021, Virtual Event, February 2-9, 2021},
	Pages = {10790--10797},
	Title = {Learning Modality-Specific Representations with Self-Supervised Multi-Task Learning for Multimodal Sentiment Analysis}}

@inproceedings{DBLP:conf/acl/PoriaHMNCM19,
	Author = {Soujanya Poria and Devamanyu Hazarika and Navonil Majumder and Gautam Naik and Erik Cambria and Rada Mihalcea},
	Booktitle = {Proceedings of the 57th Conference of the Association for Computational Linguistics, {ACL} 2019, Florence, Italy, July 28- August 2, 2019, Volume 1: Long Papers},
	Editor = {Anna Korhonen and David R. Traum and Llu{\'{\i}}s M{\`{a}}rquez},
	Pages = {527--536},
	Title = {{MELD:} {A} Multimodal Multi-Party Dataset for Emotion Recognition in Conversations}}

@inproceedings{DBLP:conf/acl/MorencyCPLZ18,
	Author = {Amir Zadeh and Paul Pu Liang and Soujanya Poria and Erik Cambria and Louis{-}Philippe Morency},
	Booktitle = {Proceedings of the 56th Annual Meeting of the Association for Computational Linguistics, {ACL} 2018, Melbourne, Australia, July 15-20, 2018, Volume 1: Long Papers},
	Editor = {Iryna Gurevych and Yusuke Miyao},
	Pages = {2236--2246},
	Publisher = {Association for Computational Linguistics},
	Title = {Multimodal Language Analysis in the Wild: {CMU-MOSEI} Dataset and Interpretable Dynamic Fusion Graph},
	Year = {2018}}

@inproceedings{DBLP:conf/icml/NgiamKKNLN11,
	Author = {Jiquan Ngiam and Aditya Khosla and Mingyu Kim and Juhan Nam and Honglak Lee and Andrew Y. Ng},
	Booktitle = {Proceedings of the 28th International Conference on Machine Learning, {ICML} 2011, Bellevue, Washington, USA, June 28 - July 2, 2011},
	Editor = {Lise Getoor and Tobias Scheffer},
	Pages = {689--696},
	Publisher = {Omnipress},
	Title = {Multimodal Deep Learning},
	Year = {2011}}

@inproceedings{DBLP:conf/acl/ZhangZWZ22,
	Author = {Yiming Zhang and Min Zhang and Sai Wu and Junbo Zhao},
	Booktitle = {Findings of the Association for Computational Linguistics: {ACL} 2022, Dublin, Ireland, May 22-27, 2022},
	Editor = {Smaranda Muresan and Preslav Nakov and Aline Villavicencio},
	Pages = {20--30},
	Publisher = {Association for Computational Linguistics},
	Title = {Towards Unifying the Label Space for Aspect- and Sentence-based Sentiment Analysis},
	Year = {2022}}

@inproceedings{DBLP:conf/cvpr/PhamDXL21,
	Author = {Hieu Pham and Zihang Dai and Qizhe Xie and Quoc V. Le},
	Booktitle = {{IEEE} Conference on Computer Vision and Pattern Recognition, {CVPR} 2021, virtual, June 19-25, 2021},
	Pages = {11557--11568},
	Publisher = {Computer Vision Foundation / {IEEE}},
	Title = {Meta Pseudo Labels},
	Year = {2021}}

@article{DBLP:journals/jmlr/RaffelSRLNMZLL20,
	Author = {Colin Raffel and Noam Shazeer and Adam Roberts and Katherine Lee and Sharan Narang and Michael Matena and Yanqi Zhou and Wei Li and Peter J. Liu},
	Journal = {J. Mach. Learn. Res.},
	Pages = {140:1--140:67},
	Title = {Exploring the Limits of Transfer Learning with a Unified Text-to-Text Transformer},
	Volume = {21},
	Year = {2020}}

@inproceedings{DBLP:conf/nips/LuBPL19,
	Author = {Jiasen Lu and Dhruv Batra and Devi Parikh and Stefan Lee},
	Booktitle = {Advances in Neural Information Processing Systems 32: Annual Conference on Neural Information Processing Systems 2019, NeurIPS 2019, December 8-14, 2019, Vancouver, BC, Canada},
	Editor = {Hanna M. Wallach and Hugo Larochelle and Alina Beygelzimer and Florence d'Alch{\'{e}}{-}Buc and Emily B. Fox and Roman Garnett},
	Pages = {13--23},
	Title = {ViLBERT: Pretraining Task-Agnostic Visiolinguistic Representations for Vision-and-Language Tasks},
	Year = {2019}}

@inproceedings{DBLP:conf/nips/VaswaniSPUJGKP17,
	Author = {Ashish Vaswani and Noam Shazeer and Niki Parmar and Jakob Uszkoreit and Llion Jones and Aidan N. Gomez and Lukasz Kaiser and Illia Polosukhin},
	Booktitle = {Advances in Neural Information Processing Systems 30: Annual Conference on Neural Information Processing Systems 2017, December 4-9, 2017, Long Beach, CA, {USA}},
	Editor = {Isabelle Guyon and Ulrike von Luxburg and Samy Bengio and Hanna M. Wallach and Rob Fergus and S. V. N. Vishwanathan and Roman Garnett},
	Pages = {5998--6008},
	Title = {Attention is All you Need},
	Year = {2017}}

@inproceedings{DBLP:conf/emnlp/ZadehCPCM17,
	Author = {Amir Zadeh and Minghai Chen and Soujanya Poria and Erik Cambria and Louis{-}Philippe Morency},
	Booktitle = {Proceedings of the 2017 Conference on Empirical Methods in Natural Language Processing, {EMNLP} 2017, Copenhagen, Denmark, September 9-11, 2017},
	Editor = {Martha Palmer and Rebecca Hwa and Sebastian Riedel},
	Pages = {1103--1114},
	Publisher = {Association for Computational Linguistics},
	Title = {Tensor Fusion Network for Multimodal Sentiment Analysis},
	Year = {2017}}

@inproceedings{DBLP:conf/acl/TsaiBLKMS19,
	Author = {Yao{-}Hung Hubert Tsai and Shaojie Bai and Paul Pu Liang and J. Zico Kolter and Louis{-}Philippe Morency and Ruslan Salakhutdinov},
	Booktitle = {Proceedings of the 57th Conference of the Association for Computational Linguistics, {ACL} 2019, Florence, Italy, July 28- August 2, 2019, Volume 1: Long Papers},
	Editor = {Anna Korhonen and David R. Traum and Llu{\'{\i}}s M{\`{a}}rquez},
	Pages = {6558--6569},
	Publisher = {Association for Computational Linguistics},
	Title = {Multimodal Transformer for Unaligned Multimodal Language Sequences},
	Year = {2019}}

@inproceedings{DBLP:conf/icmi/MorencyMD11,
	Author = {Louis{-}Philippe Morency and Rada Mihalcea and Payal Doshi},
	Booktitle = {Proceedings of the 13th International Conference on Multimodal Interfaces, {ICMI} 2011, Alicante, Spain, November 14-18, 2011},
	Pages = {169--176},
	Title = {Towards multimodal sentiment analysis: harvesting opinions from the web},
	Year = {2011}}

@article{DBLP:journals/expert/ZadehZPM16,
	Author = {Amir Zadeh and Rowan Zellers and Eli Pincus and Louis{-}Philippe Morency},
	Journal = {{IEEE} Intell. Syst.},
	Number = {6},
	Pages = {82--88},
	Title = {Multimodal Sentiment Intensity Analysis in Videos: Facial Gestures and Verbal Messages},
	Volume = {31},
	Year = {2016}}

@article{DBLP:journals/lre/BussoBLKMKCLN08,
	Author = {Carlos Busso and Murtaza Bulut and Chi{-}Chun Lee and Abe Kazemzadeh and Emily Mower and Samuel Kim and Jeannette N. Chang and Sungbok Lee and Shrikanth S. Narayanan},
	Journal = {Lang. Resour. Evaluation},
	Number = {4},
	Pages = {335--359},
	Title = {{IEMOCAP:} interactive emotional dyadic motion capture database},
	Volume = {42},
	Year = {2008}}

@inproceedings{DBLP:conf/emnlp/LiangLZM18,
	Author = {Paul Pu Liang and Ziyin Liu and Amir Zadeh and Louis{-}Philippe Morency},
	Booktitle = {Proceedings of the 2018 Conference on Empirical Methods in Natural Language Processing, Brussels, Belgium, October 31 - November 4, 2018},
	Pages = {150--161},
	Title = {Multimodal Language Analysis with Recurrent Multistage Fusion},
	Year = {2018}}

@inproceedings{DBLP:conf/aaai/SunSSL20,
	Author = {Zhongkai Sun and Prathusha Kameswara Sarma and William A. Sethares and Yingyu Liang},
	Booktitle = {The Thirty-Fourth {AAAI} Conference on Artificial Intelligence, {AAAI} 2020, The Thirty-Second Innovative Applications of Artificial Intelligence Conference, {IAAI} 2020, The Tenth {AAAI} Symposium on Educational Advances in Artificial Intelligence, {EAAI} 2020, New York, NY, USA, February 7-12, 2020},
	Pages = {8992--8999},
	Title = {Learning Relationships between Text, Audio, and Video via Deep Canonical Correlation for Multimodal Language Analysis},
	Year = {2020}}

@article{DBLP:journals/corr/abs-1911-09826,
	Author = {Amir Zadeh and Chengfeng Mao and Kelly Shi and Yiwei Zhang and Paul Pu Liang and Soujanya Poria and Louis{-}Philippe Morency},
	Journal = {CoRR},
	Title = {Factorized Multimodal Transformer for Multimodal Sequential Learning},
	Volume = {abs/1911.09826},
	Year = {2019}}

@article{DBLP:journals/corr/abs-2002-06353,
	Author = {Huaishao Luo and Lei Ji and Botian Shi and Haoyang Huang and Nan Duan and Tianrui Li and Xilin Chen and Ming Zhou},
	Journal = {CoRR},
	Title = {UniViLM: {A} Unified Video and Language Pre-Training Model for Multimodal Understanding and Generation},
	Volume = {abs/2002.06353},
	Year = {2020}}

@inproceedings{DBLP:conf/acl/RahmanHLZMMH20,
	Author = {Wasifur Rahman and Md. Kamrul Hasan and Sangwu Lee and AmirAli Bagher Zadeh and Chengfeng Mao and Louis{-}Philippe Morency and Mohammed E. Hoque},
	Booktitle = {Proceedings of the 58th Annual Meeting of the Association for Computational Linguistics, {ACL} 2020, Online, July 5-10, 2020},
	Pages = {2359--2369},
	Title = {Integrating Multimodal Information in Large Pretrained Transformers},
	Year = {2020}}

@inproceedings{DBLP:conf/naacl/YangWYZRZPM21,
	Author = {Jianing Yang and Yongxin Wang and Ruitao Yi and Yuying Zhu and Azaan Rehman and Amir Zadeh and Soujanya Poria and Louis{-}Philippe Morency},
	Booktitle = {Proceedings of the 2021 Conference of the North American Chapter of the Association for Computational Linguistics: Human Language Technologies, {NAACL-HLT} 2021, Online, June 6-11, 2021},
	Pages = {1009--1021},
	Title = {{MTAG:} Modal-Temporal Attention Graph for Unaligned Human Multimodal Language Sequences},
	Year = {2021}}

@inproceedings{DBLP:conf/icml/HoulsbyGJMLGAG19,
	Author = {Neil Houlsby and Andrei Giurgiu and Stanislaw Jastrzebski and Bruna Morrone and Quentin de Laroussilhe and Andrea Gesmundo and Mona Attariyan and Sylvain Gelly},
	Booktitle = {Proceedings of the 36th International Conference on Machine Learning, {ICML} 2019, 9-15 June 2019, Long Beach, California, {USA}},
	Pages = {2790--2799},
	Title = {Parameter-Efficient Transfer Learning for {NLP}},
	Year = {2019}}

@inproceedings{DBLP:conf/icassp/HuHWJM22,
	Author = {Dou Hu and Xiaolong Hou and Lingwei Wei and Lian{-}Xin Jiang and Yang Mo},
	Booktitle = {{IEEE} International Conference on Acoustics, Speech and Signal Processing, {ICASSP} 2022, Virtual and Singapore, 23-27 May 2022},
	Pages = {7037--7041},
	Title = {{MM-DFN:} Multimodal Dynamic Fusion Network for Emotion Recognition in Conversations},
	Year = {2022}}

@article{DBLP:journals/corr/abs-2201-05966,
	Author = {Tianbao Xie and Chen Henry Wu and Peng Shi and Ruiqi Zhong and Torsten Scholak and Michihiro Yasunaga and Chien{-}Sheng Wu and Ming Zhong and Pengcheng Yin and Sida I. Wang and Victor Zhong and Bailin Wang and Chengzu Li and Connor Boyle and Ansong Ni and Ziyu Yao and Dragomir R. Radev and Caiming Xiong and Lingpeng Kong and Rui Zhang and Noah A. Smith and Luke Zettlemoyer and Tao Yu},
	Journal = {CoRR},
	Title = {UnifiedSKG: Unifying and Multi-Tasking Structured Knowledge Grounding with Text-to-Text Language Models},
	Volume = {abs/2201.05966},
	Year = {2022}}

@inproceedings{DBLP:conf/aaai/ZhangMWJLY22,
	Author = {Zhengkun Zhang and Xiaojun Meng and Yasheng Wang and Xin Jiang and Qun Liu and Zhenglu Yang},
	Booktitle = {Thirty-Sixth {AAAI} Conference on Artificial Intelligence, {AAAI} 2022, Thirty-Fourth Conference on Innovative Applications of Artificial Intelligence, {IAAI} 2022, The Twelveth Symposium on Educational Advances in Artificial Intelligence, {EAAI} 2022 Virtual Event, February 22 - March 1, 2022},
	Pages = {11757--11764},
	Title = {UniMS: {A} Unified Framework for Multimodal Summarization with Knowledge Distillation},
	Year = {2022}}

@inproceedings{DBLP:conf/acl/LewisLGGMLSZ20,
	Author = {Mike Lewis and Yinhan Liu and Naman Goyal and Marjan Ghazvininejad and Abdelrahman Mohamed and Omer Levy and Veselin Stoyanov and Luke Zettlemoyer},
	Booktitle = {Proceedings of the 58th Annual Meeting of the Association for Computational Linguistics, {ACL} 2020, Online, July 5-10, 2020},
	Pages = {7871--7880},
	Title = {{BART:} Denoising Sequence-to-Sequence Pre-training for Natural Language Generation, Translation, and Comprehension},
	Year = {2020}}

@article{DBLP:journals/corr/abs-2112-01368,
	Author = {Huaishao Luo and Lei Ji and Yanyong Huang and Bin Wang and Shenggong Ji and Tianrui Li},
	Journal = {CoRR},
	Title = {ScaleVLAD: Improving Multimodal Sentiment Analysis via Multi-Scale Fusion of Locally Descriptors},
	Volume = {abs/2112.01368},
	Year = {2021}}

@article{DBLP:journals/corr/abs-2109-01797,
	Author = {Sijie Mai and Ying Zeng and Shuangjia Zheng and Haifeng Hu},
	Journal = {CoRR},
	Title = {Hybrid Contrastive Learning of Tri-Modal Representation for Multimodal Sentiment Analysis},
	Volume = {abs/2109.01797},
	Year = {2021}}

@inproceedings{DBLP:conf/emnlp/ZengM021,
	Author = {Ying Zeng and Sijie Mai and Haifeng Hu},
	Booktitle = {Findings of the Association for Computational Linguistics: {EMNLP} 2021, Virtual Event / Punta Cana, Dominican Republic, 16-20 November, 2021},
	Pages = {1262--1274},
	Title = {Which is Making the Contribution: Modulating Unimodal and Cross-modal Dynamics for Multimodal Sentiment Analysis},
	Year = {2021}}

@inproceedings{DBLP:conf/emnlp/XieYSLJ21,
	Author = {Yunhe Xie and Kailai Yang and Chengjie Sun and Bingquan Liu and Zhenzhou Ji},
	Booktitle = {Findings of the Association for Computational Linguistics: {EMNLP} 2021, Virtual Event / Punta Cana, Dominican Republic, 16-20 November, 2021},
	Editor = {Marie{-}Francine Moens and Xuanjing Huang and Lucia Specia and Scott Wen{-}tau Yih},
	Pages = {2879--2889},
	Publisher = {Association for Computational Linguistics},
	Title = {Knowledge-Interactive Network with Sentiment Polarity Intensity-Aware Multi-Task Learning for Emotion Recognition in Conversations},
	Year = {2021}}

@article{DBLP:journals/corr/abs-2205-02455,
	Author = {Abhinav Joshi and Ashwani Bhat and Ayush Jain and Atin Vikram Singh and Ashutosh Modi},
	Journal = {CoRR},
	Title = {{COGMEN:} COntextualized {GNN} based Multimodal Emotion recognitioN},
	Volume = {abs/2205.02455},
	Year = {2022}}

@inproceedings{DBLP:conf/acl/HuLZJ20,
	Author = {Jingwen Hu and Yuchen Liu and Jinming Zhao and Qin Jin},
	Booktitle = {Proceedings of the 59th Annual Meeting of the Association for Computational Linguistics and the 11th International Joint Conference on Natural Language Processing, {ACL/IJCNLP} 2021, (Volume 1: Long Papers), Virtual Event, August 1-6, 2021},
	Editor = {Chengqing Zong and Fei Xia and Wenjie Li and Roberto Navigli},
	Pages = {5666--5675},
	Publisher = {Association for Computational Linguistics},
	Title = {{MMGCN:} Multimodal Fusion via Deep Graph Convolution Network for Emotion Recognition in Conversation},
	Year = {2021}}

@inproceedings{DBLP:conf/emnlp/GhosalMPCG19,
	Author = {Deepanway Ghosal and Navonil Majumder and Soujanya Poria and Niyati Chhaya and Alexander F. Gelbukh},
	Booktitle = {Proceedings of the 2019 Conference on Empirical Methods in Natural Language Processing and the 9th International Joint Conference on Natural Language Processing, {EMNLP-IJCNLP} 2019, Hong Kong, China, November 3-7, 2019},
	Editor = {Kentaro Inui and Jing Jiang and Vincent Ng and Xiaojun Wan},
	Pages = {154--164},
	Publisher = {Association for Computational Linguistics},
	Title = {DialogueGCN: {A} Graph Convolutional Neural Network for Emotion Recognition in Conversation},
	Year = {2019}}

@inproceedings{DBLP:conf/emnlp/ChengFB021,
	Author = {Junyan Cheng and Iordanis Fostiropoulos and Barry W. Boehm and Mohammad Soleymani},
	Booktitle = {Proceedings of the 2021 Conference on Empirical Methods in Natural Language Processing, {EMNLP} 2021, Virtual Event / Punta Cana, Dominican Republic, 7-11 November, 2021},
	Editor = {Marie{-}Francine Moens and Xuanjing Huang and Lucia Specia and Scott Wen{-}tau Yih},
	Pages = {2447--2458},
	Publisher = {Association for Computational Linguistics},
	Title = {Multimodal Phased Transformer for Sentiment Analysis},
	Year = {2021}}

@inproceedings{DBLP:conf/emnlp/ChauhanAEB19,
	Author = {Dushyant Singh Chauhan and Md. Shad Akhtar and Asif Ekbal and Pushpak Bhattacharyya},
	Booktitle = {Proceedings of the 2019 Conference on Empirical Methods in Natural Language Processing and the 9th International Joint Conference on Natural Language Processing, {EMNLP-IJCNLP} 2019, Hong Kong, China, November 3-7, 2019},
	Editor = {Kentaro Inui and Jing Jiang and Vincent Ng and Xiaojun Wan},
	Pages = {5646--5656},
	Publisher = {Association for Computational Linguistics},
	Title = {Context-aware Interactive Attention for Multi-modal Sentiment and Emotion Analysis},
	Year = {2019}}

@inproceedings{DBLP:conf/emnlp/ChengYZS21,
	Author = {Kewei Cheng and Ziqing Yang and Ming Zhang and Yizhou Sun},
	Booktitle = {Proceedings of the 2021 Conference on Empirical Methods in Natural Language Processing, {EMNLP} 2021, Virtual Event / Punta Cana, Dominican Republic, 7-11 November, 2021},
	Editor = {Marie{-}Francine Moens and Xuanjing Huang and Lucia Specia and Scott Wen{-}tau Yih},
	Pages = {9753--9771},
	Title = {UniKER: {A} Unified Framework for Combining Embedding and Definite Horn Rule Reasoning for Knowledge Graph Inference}}

@article{DBLP:journals/corr/abs-2111-02358,
	Author = {Wenhui Wang and Hangbo Bao and Li Dong and Furu Wei},
	Journal = {CoRR},
	Title = {VLMo: Unified Vision-Language Pre-Training with Mixture-of-Modality-Experts},
	Volume = {abs/2111.02358},
	Year = {2021}}

@inproceedings{DBLP:conf/icml/TanL19,
	Author = {Mingxing Tan and Quoc V. Le},
	Booktitle = {Proceedings of the 36th International Conference on Machine Learning, {ICML} 2019, 9-15 June 2019, Long Beach, California, {USA}},
	Editor = {Kamalika Chaudhuri and Ruslan Salakhutdinov},
	Pages = {6105--6114},
	Publisher = {{PMLR}},
	Series = {Proceedings of Machine Learning Research},
	Title = {EfficientNet: Rethinking Model Scaling for Convolutional Neural Networks},
	Volume = {97},
	Year = {2019}}

@inproceedings{DBLP:conf/naacl/AkhtarCGPEB19,
	Author = {Md. Shad Akhtar and Dushyant Singh Chauhan and Deepanway Ghosal and Soujanya Poria and Asif Ekbal and Pushpak Bhattacharyya},
	Booktitle = {Proceedings of the 2019 Conference of the North American Chapter of the Association for Computational Linguistics: Human Language Technologies, {NAACL-HLT} 2019, Minneapolis, MN, USA, June 2-7, 2019, Volume 1 (Long and Short Papers)},
	Editor = {Jill Burstein and Christy Doran and Thamar Solorio},
	Pages = {370--379},
	Publisher = {Association for Computational Linguistics},
	Title = {Multi-task Learning for Multi-modal Emotion Recognition and Sentiment Analysis},
	Year = {2019}}

@article{DBLP:journals/kbs/HuLZ21,
	Author = {Guimin Hu and Guangming Lu and Yi Zhao},
	Journal = {Knowl. Based Syst.},
	Pages = {106584},
	Title = {{FSS-GCN:} {A} graph convolutional networks with fusion of semantic and structure for emotion cause analysis},
	Volume = {212},
	Year = {2021}}

@inproceedings{DBLP:conf/emnlp/HuLZ21,
	Author = {Guimin Hu and Guangming Lu and Yi Zhao},
	Booktitle = {Findings of the Association for Computational Linguistics: {EMNLP} 2021, Virtual Event / Punta Cana, Dominican Republic, 16-20 November, 2021},
	Pages = {558--568},
	Title = {Bidirectional Hierarchical Attention Networks based on Document-level Context for Emotion Cause Extraction},
	Year = {2021}}

@article{DBLP:journals/corr/abs-2204-04637,
	Author = {Zhi Chen and Lu Chen and Bei Chen and Libo Qin and Yuncong Liu and Su Zhu and Jian{-}Guang Lou and Kai Yu},
	Journal = {CoRR},
	Title = {UniDU: Towards {A} Unified Generative Dialogue Understanding Framework},
	Volume = {abs/2204.04637},
	Year = {2022}}

@inproceedings{DBLP:conf/icdm/PoriaCHMZM17,
	Author = {Soujanya Poria and Erik Cambria and Devamanyu Hazarika and Navonil Majumder and Amir Zadeh and Louis{-}Philippe Morency},
	Booktitle = {2017 {IEEE} International Conference on Data Mining, {ICDM} 2017, New Orleans, LA, USA, November 18-21, 2017},
	Pages = {1033--1038},
	Title = {Multi-level Multiple Attentions for Contextual Multimodal Sentiment Analysis},
	Year = {2017}}

@article{baltruvsaitis2018multimodal,
	Author = {Baltru{\v{s}}aitis, Tadas and Ahuja, Chaitanya and Morency, Louis-Philippe},
	Journal = {IEEE transactions on pattern analysis and machine intelligence},
	Number = {2},
	Pages = {423--443},
	Publisher = {IEEE},
	Title = {Multimodal machine learning: A survey and taxonomy},
	Volume = {41},
	Year = {2018}}

@book{davidson2009handbook,
	Author = {Davidson, Richard J and Sherer, Klaus R and Goldsmith, H Hill},
	Publisher = {Oxford University Press},
	Title = {Handbook of affective sciences},
	Year = {2009}}

@book{ben2001subtlety,
	Author = {Ben-Ze'ev, Aaron},
	Publisher = {MIT press},
	Title = {The subtlety of emotions},
	Year = {2001}}

@incollection{shelly2004emotions,
	Author = {Shelly, Robert K},
	Booktitle = {Theory and research on human emotions},
	Publisher = {Emerald Group Publishing Limited},
	Title = {Emotions, sentiments, and performance expectations},
	Year = {2004}}

@article{zhang2022mintrec,
	Author = {Zhang, Hanlei and Xu, Hua and Wang, Xin and Zhou, Qianrui and Zhao, Shaojie and Teng, Jiayan},
	Journal = {arXiv preprint arXiv:2209.04355},
	Title = {MIntRec: A New Dataset for Multimodal Intent Recognition},
	Year = {2022}}

@inproceedings{yuan2021transformer,
	Author = {Yuan, Ziqi and Li, Wei and Xu, Hua and Yu, Wenmeng},
	Booktitle = {Proceedings of the 29th ACM International Conference on Multimedia},
	Pages = {4400--4407},
	Title = {Transformer-based feature reconstruction network for robust multimodal sentiment analysis},
	Year = {2021}}

@inproceedings{DBLP:conf/mm/JainSG023,
	Author = {Raghav Jain and Apoorva Singh and Vivek Kumar Gangwar and Sriparna Saha},
	Booktitle = {Proceedings of the 31st {ACM} International Conference on Multimedia, {MM} 2023, Ottawa, ON, Canada, 29 October 2023- 3 November 2023},
	Pages = {8571--8579},
	Title = {AbCoRD: Exploiting multimodal generative approach for Aspect-based Complaint and Rationale Detection},
	Year = {2023}}

@inproceedings{DBLP:conf/acl/ZhouGLYZY23,
	Author = {Ru Zhou and Wenya Guo and Xumeng Liu and Shenglong Yu and Ying Zhang and Xiaojie Yuan},
	Booktitle = {Findings of the Association for Computational Linguistics: {ACL} 2023, Toronto, Canada, July 9-14, 2023},
	Pages = {8184--8196},
	Title = {AoM: Detecting Aspect-oriented Information for Multimodal Aspect-Based Sentiment Analysis},
	Year = {2023}}

@article{DBLP:journals/tcsv/ChenZLZ22,
	Author = {Rongfei Chen and Wenju Zhou and Yang Li and Huiyu Zhou},
	Journal = {{IEEE} Trans. Circuits Syst. Video Technol.},
	Number = {12},
	Pages = {8703--8716},
	Title = {Video-Based Cross-Modal Auxiliary Network for Multimodal Sentiment Analysis},
	Volume = {32},
	Year = {2022}}

@article{DBLP:journals/ipm/YangNY22,
	Author = {Li Yang and Jin{-}Cheon Na and Jianfei Yu},
	Journal = {Inf. Process. Manag.},
	Number = {5},
	Pages = {103038},
	Title = {Cross-Modal Multitask Transformer for End-to-End Multimodal Aspect-Based Sentiment Analysis},
	Volume = {59},
	Year = {2022}}

@inproceedings{DBLP:conf/ijcnlp/YuWYZ22,
	Author = {Zhewen Yu and Jin Wang and Liang{-}Chih Yu and Xuejie Zhang},
	Booktitle = {Proceedings of the 2nd Conference of the Asia-Pacific Chapter of the Association for Computational Linguistics and the 12th International Joint Conference on Natural Language Processing, {AACL/IJCNLP} 2022 - Volume 1: Long Papers, Online Only, November 20-23, 2022},
	Pages = {414--423},
	Title = {Dual-Encoder Transformers with Cross-modal Alignment for Multimodal Aspect-based Sentiment Analysis},
	Year = {2022}}

@inproceedings{DBLP:conf/nlpcc/LiuWZ23,
	Author = {Kuanghong Liu and Jin Wang and Xuejie Zhang},
	Booktitle = {Natural Language Processing and Chinese Computing - 12th National {CCF} Conference, {NLPCC} 2023, Foshan, China, October 12-15, 2023, Proceedings, Part {II}},
	Pages = {481--493},
	Title = {Entity-Related Unsupervised Pretraining with Visual Prompts for Multimodal Aspect-Based Sentiment Analysis},
	Year = {2023}}

@inproceedings{DBLP:conf/emnlp/YangZ022,
	Author = {Hao Yang and Yanyan Zhao and Bing Qin},
	Booktitle = {Proceedings of the 2022 Conference on Empirical Methods in Natural Language Processing, {EMNLP} 2022, Abu Dhabi, United Arab Emirates, December 7-11, 2022},
	Pages = {3324--3335},
	Title = {Face-Sensitive Image-to-Emotional-Text Cross-modal Translation for Multimodal Aspect-based Sentiment Analysis},
	Year = {2022}}

@inproceedings{DBLP:conf/acl/YangFWSWZHP23,
	Author = {Xiaocui Yang and Shi Feng and Daling Wang and Qi Sun and Wenfang Wu and Yifei Zhang and Pengfei Hong and Soujanya Poria},
	Booktitle = {Findings of the Association for Computational Linguistics: {ACL} 2023, Toronto, Canada, July 9-14, 2023},
	Pages = {11575--11589},
	Title = {Few-shot Joint Multimodal Aspect-Sentiment Analysis Based on Generative Multimodal Prompt},
	Year = {2023}}

@inproceedings{DBLP:conf/acl/ZhangZZZL0C22,
	Author = {Kai Zhang and Kun Zhang and Mengdi Zhang and Hongke Zhao and Qi Liu and Wei Wu and Enhong Chen},
	Booktitle = {Findings of the Association for Computational Linguistics: {ACL} 2022, Dublin, Ireland, May 22-27, 2022},
	Pages = {3599--3610},
	Title = {Incorporating Dynamic Semantics into Pre-Trained Language Model for Aspect-based Sentiment Analysis},
	Year = {2022}}

@inproceedings{DBLP:conf/emnlp/JuZXLLZZ21,
	Author = {Xincheng Ju and Dong Zhang and Rong Xiao and Junhui Li and Shoushan Li and Min Zhang and Guodong Zhou},
	Booktitle = {Proceedings of the 2021 Conference on Empirical Methods in Natural Language Processing, {EMNLP} 2021, Virtual Event / Punta Cana, Dominican Republic, 7-11 November, 2021},
	Pages = {4395--4405},
	Title = {Joint Multi-modal Aspect-Sentiment Analysis with Auxiliary Cross-modal Relation Detection},
	Year = {2021}}

@article{DBLP:journals/corr/abs-2310-14605,
	Author = {Fei Zhao and Chunhui Li and Zhen Wu and Yawen Ouyang and Jianbing Zhang and Xinyu Dai},
	Journal = {CoRR},
	Title = {{M2DF:} Multi-grained Multi-curriculum Denoising Framework for Multimodal Aspect-based Sentiment Analysis},
	Volume = {abs/2310.14605},
	Year = {2023}}

@article{DBLP:journals/corr/abs-2206-13969,
	Author = {Hao Yang and Yanyan Zhao and Jianwei Liu and Yang Wu and Bing Qin},
	Journal = {CoRR},
	Title = {{MACSA:} {A} Multimodal Aspect-Category Sentiment Analysis Dataset with Multimodal Fine-grained Aligned Annotations},
	Volume = {abs/2206.13969},
	Year = {2022}}

@inproceedings{DBLP:conf/iccc2/XuLW22,
	Author = {Carol Xu and Xuan Luo and Dan Wang},
	Booktitle = {Cognitive Computing - {ICCC} 2022 - 6th International Conference, Held as Part of the Services Conference Federation, {SCF} 2022, Honolulu, HI, USA, December 10-14, 2022, Proceedings},
	Pages = {83--90},
	Title = {{MCPR:} {A} Chinese Product Review Dataset for Multimodal Aspect-Based Sentiment Analysis},
	Year = {2022}}

@article{DBLP:journals/www/ZhangWLLGY21,
	Author = {Zhe Zhang and Zhu Wang and Xiaona Li and Nannan Liu and Bin Guo and Zhiwen Yu},
	Journal = {World Wide Web},
	Number = {6},
	Pages = {1957--1974},
	Title = {ModalNet: an aspect-level sentiment classification model by exploring multimodal data with fusion discriminant attentional network},
	Volume = {24},
	Year = {2021}}

@inproceedings{DBLP:conf/cikm/JinTLQYCZ23,
	Author = {Hanqi Jin and Jiwei Tan and Lixin Liu and Lisong Qiu and Shaowei Yao and Xi Chen and Xiaoyi Zeng},
	Booktitle = {Proceedings of the 32nd {ACM} International Conference on Information and Knowledge Management, {CIKM} 2023, Birmingham, United Kingdom, October 21-25, 2023},
	Pages = {3988--3992},
	Title = {{MSRA:} {A} Multi-Aspect Semantic Relevance Approach for E-Commerce via Multimodal Pre-Training},
	Year = {2023}}

@inproceedings{DBLP:conf/acl/LingYX22,
	Author = {Yan Ling and Jianfei Yu and Rui Xia},
	Booktitle = {Proceedings of the 60th Annual Meeting of the Association for Computational Linguistics (Volume 1: Long Papers), {ACL} 2022, Dublin, Ireland, May 22-27, 2022},
	Pages = {2149--2159},
	Title = {Vision-Language Pre-Training for Multimodal Aspect-Based Sentiment Analysis},
	Year = {2022}}

@article{DBLP:journals/kbs/RahmaniHZKKH23,
	Author = {Sana Rahmani and Saeid Hosseini and Raziyeh Zall and Mohammad Reza Kangavari and Sara Kamran and Wen Hua},
	Journal = {Knowl. Based Syst.},
	Pages = {110219},
	Title = {Transfer-based adaptive tree for multimodal sentiment analysis based on user latent aspects},
	Volume = {261},
	Year = {2023}}

@inproceedings{DBLP:conf/semco/AnschutzEG23,
	Author = {Miriam Ansch{\"{u}}tz and Tobias Eder and Georg Groh},
	Booktitle = {17th {IEEE} International Conference on Semantic Computing, {ICSC} 2023, Laguna Hills, CA, USA, February 1-3, 2023},
	Pages = {1--8},
	Title = {Retrieving Users' Opinions on Social Media with Multimodal Aspect-Based Sentiment Analysis},
	Year = {2023}}

@inproceedings{DBLP:conf/icmcs/XuSX23,
	Author = {Zenan Xu and Qinliang Su and Junxi Xiao},
	Booktitle = {{IEEE} International Conference on Multimedia and Expo, {ICME} 2023, Brisbane, Australia, July 10-14, 2023},
	Pages = {1379--1384},
	Title = {Multimodal Aspect-Based Sentiment Classification with Knowledge-Injected Transformer},
	Year = {2023}}

@inproceedings{DBLP:conf/aaai/XuMC19,
	Author = {Nan Xu and Wenji Mao and Guandan Chen},
	Booktitle = {The Thirty-Third {AAAI} Conference on Artificial Intelligence, {AAAI} 2019, The Thirty-First Innovative Applications of Artificial Intelligence Conference, {IAAI} 2019, The Ninth {AAAI} Symposium on Educational Advances in Artificial Intelligence, {EAAI} 2019, Honolulu, Hawaii, USA, January 27 - February 1, 2019},
	Pages = {371--378},
	Title = {Multi-Interactive Memory Network for Aspect Based Multimodal Sentiment Analysis},
	Year = {2019}}

@inproceedings{DBLP:conf/aaai/ZhangCSW22,
	Author = {Yi Zhang and Mingyuan Chen and Jundong Shen and Chongjun Wang},
	Booktitle = {Thirty-Sixth {AAAI} Conference on Artificial Intelligence, {AAAI} 2022, Thirty-Fourth Conference on Innovative Applications of Artificial Intelligence, {IAAI} 2022, The Twelveth Symposium on Educational Advances in Artificial Intelligence, {EAAI} 2022 Virtual Event, February 22 - March 1, 2022},
	Pages = {9100--9108},
	Title = {Tailor Versatile Multi-Modal Learning for Multi-Label Emotion Recognition},
	Year = {2022}}

@inproceedings{DBLP:conf/mm/ZhaoZL21,
	Author = {Jiawei Zhao and Yifan Zhao and Jia Li},
	Booktitle = {{MM} '21: {ACM} Multimedia Conference, Virtual Event, China, October 20 - 24, 2021},
	Pages = {469--477},
	Title = {{M3TR:} Multi-modal Multi-label Recognition with Transformer},
	Year = {2021}}

@inproceedings{DBLP:conf/aaai/ZhangJZLLZZ21,
	Author = {Dong Zhang and Xincheng Ju and Wei Zhang and Junhui Li and Shoushan Li and Qiaoming Zhu and Guodong Zhou},
	Booktitle = {Thirty-Fifth {AAAI} Conference on Artificial Intelligence, {AAAI} 2021, Thirty-Third Conference on Innovative Applications of Artificial Intelligence, {IAAI} 2021, The Eleventh Symposium on Educational Advances in Artificial Intelligence, {EAAI} 2021, Virtual Event, February 2-9, 2021},
	Pages = {14338--14346},
	Title = {Multi-modal Multi-label Emotion Recognition with Heterogeneous Hierarchical Message Passing},
	Year = {2021}}

@inproceedings{DBLP:conf/emnlp/ZhangJLLZZ20,
	Author = {Dong Zhang and Xincheng Ju and Junhui Li and Shoushan Li and Qiaoming Zhu and Guodong Zhou},
	Booktitle = {Proceedings of the 2020 Conference on Empirical Methods in Natural Language Processing, {EMNLP} 2020, Online, November 16-20, 2020},
	Pages = {3584--3593},
	Title = {Multi-modal Multi-label Emotion Detection with Modality and Label Dependence},
	Year = {2020}}

@inproceedings{DBLP:conf/emnlp/PenningtonSM14,
	Author = {Jeffrey Pennington and Richard Socher and Christopher D. Manning},
	Booktitle = {Proceedings of the 2014 Conference on Empirical Methods in Natural Language Processing, {EMNLP} 2014, October 25-29, 2014, Doha, Qatar, {A} meeting of SIGDAT, a Special Interest Group of the {ACL}},
	Pages = {1532--1543},
	Title = {Glove: Global Vectors for Word Representation},
	Year = {2014}}

@inproceedings{DBLP:conf/mm/JuZLZ20,
	Author = {Xincheng Ju and Dong Zhang and Junhui Li and Guodong Zhou},
	Booktitle = {{MM} '20: The 28th {ACM} International Conference on Multimedia, Virtual Event / Seattle, WA, USA, October 12-16, 2020},
	Pages = {512--520},
	Title = {Transformer-based Label Set Generation for Multi-modal Multi-label Emotion Detection},
	Year = {2020}}

@inproceedings{DBLP:conf/www/GeJCWYG23,
	Author = {Shiping Ge and Zhiwei Jiang and Zifeng Cheng and Cong Wang and Yafeng Yin and Qing Gu},
	Booktitle = {Proceedings of the {ACM} Web Conference 2023, {WWW} 2023, Austin, TX, USA, 30 April 2023 - 4 May 2023},
	Pages = {1510--1518},
	Title = {Learning Robust Multi-Modal Representation for Multi-Label Emotion Recognition via Adversarial Masking and Perturbation},
	Year = {2023}}

@article{9926166,
	Author = {Li, Wei and Li, Yang and Pandelea, Vlad and Ge, Mengshi and Zhu, Luyao and Cambria, Erik},
	Doi = {10.1109/TAFFC.2022.3216551},
	Journal = {IEEE Transactions on Affective Computing},
	Number = {3},
	Pages = {1754-1765},
	Title = {{ECPEC}: Emotion-Cause Pair Extraction in Conversations},
	Volume = {14},
	Year = {2023}}

@article{DBLP:journals/corr/abs-2110-08020,
	Author = {Fanfan Wang and Zixiang Ding and Rui Xia and Zhaoyu Li and Jianfei Yu},
	Journal = {CoRR},
	Title = {Multimodal Emotion-Cause Pair Extraction in Conversations},
	Volume = {abs/2110.08020},
	Year = {2021}}

@inproceedings{DBLP:conf/emnlp/HuLZLWL22,
	Author = {Guimin Hu and Ting{-}En Lin and Yi Zhao and Guangming Lu and Yuchuan Wu and Yongbin Li},
	Booktitle = {Proceedings of the 2022 Conference on Empirical Methods in Natural Language Processing, {EMNLP} 2022, Abu Dhabi, United Arab Emirates, December 7-11, 2022},
	Pages = {7837--7851},
	Title = {UniMSE: Towards Unified Multimodal Sentiment Analysis and Emotion Recognition},
	Year = {2022}}

@inproceedings{DBLP:conf/acl/TangLJCZK20,
	Author = {Jiajia Tang and Kang Li and Xuanyu Jin and Andrzej Cichocki and Qibin Zhao and Wanzeng Kong},
	Booktitle = {Proceedings of the 59th Annual Meeting of the Association for Computational Linguistics and the 11th International Joint Conference on Natural Language Processing, {ACL/IJCNLP} 2021, (Volume 1: Long Papers), Virtual Event, August 1-6, 2021},
	Pages = {5301--5311},
	Title = {{CTFN:} Hierarchical Learning for Multimodal Sentiment Analysis Using Coupled-Translation Fusion Network},
	Year = {2021}}

@proceedings{DBLP:conf/acl/2021-1,
	Editor = {Chengqing Zong and Fei Xia and Wenjie Li and Roberto Navigli},
	Publisher = {Association for Computational Linguistics},
	Title = {Proceedings of the 59th Annual Meeting of the Association for Computational Linguistics and the 11th International Joint Conference on Natural Language Processing, {ACL/IJCNLP} 2021, (Volume 1: Long Papers), Virtual Event, August 1-6, 2021},
	Year = {2021}}

@article{KIM202337,
	Doi = {https://doi.org/10.1016/j.inffus.2022.11.022},
	Issn = {1566-2535},
	Journal = {Information Fusion},
	Pages = {37-45},
	Title = {AOBERT: All-modalities-in-One BERT for multimodal sentiment analysis},
	Volume = {92},
	Year = {2023}}

@inproceedings{yu2020ch,
	Author = {Yu, Wenmeng and Xu, Hua and Meng, Fanyang and Zhu, Yilin and Ma, Yixiao and Wu, Jiele and Zou, Jiyun and Yang, Kaicheng},
	Booktitle = {Proceedings of the 58th annual meeting of the association for computational linguistics},
	Pages = {3718--3727},
	Title = {Ch-sims: A chinese multimodal sentiment analysis dataset with fine-grained annotation of modality},
	Year = {2020}}

@inproceedings{zeng2022mitigating,
	Author = {Zeng, Jiandian and Zhou, Jiantao and Liu, Tianyi},
	Booktitle = {Proceedings of the 2022 Conference on Empirical Methods in Natural Language Processing},
	Pages = {2924--2934},
	Title = {Mitigating Inconsistencies in Multimodal Sentiment Analysis under Uncertain Missing Modalities},
	Year = {2022}}

@article{luo2021scalevlad,
	Author = {Luo, Huaishao and Ji, Lei and Huang, Yanyong and Wang, Bin and Ji, Shenggong and Li, Tianrui},
	Journal = {arXiv preprint arXiv:2112.01368},
	Title = {Scalevlad: Improving multimodal sentiment analysis via multi-scale fusion of locally descriptors},
	Year = {2021}}

@inproceedings{chen2020swafn,
	Author = {Chen, Minping and Li, Xia},
	Booktitle = {Proceedings of the 28th international conference on computational linguistics},
	Pages = {1067--1077},
	Title = {Swafn: Sentimental words aware fusion network for multimodal sentiment analysis},
	Year = {2020}}

@inproceedings{qian2023sentiment,
	Author = {Qian, Fan and Han, Jiqing and He, Yongjun and Zheng, Tieran and Zheng, Guibin},
	Booktitle = {Findings of the Association for Computational Linguistics: ACL 2023},
	Pages = {12966--12978},
	Title = {Sentiment Knowledge Enhanced Self-supervised Learning for Multimodal Sentiment Analysis},
	Year = {2023}}

@inproceedings{DBLP:conf/acl/ZhaoZ0LJW022,
	Author = {Jinming Zhao and Tenggan Zhang and Jingwen Hu and Yuchen Liu and Qin Jin and Xinchao Wang and Haizhou Li},
	Booktitle = {{ACL} 2022},
	Pages = {5699--5710},
	Title = {{M3ED:} Multi-modal Multi-scene Multi-label Emotional Dialogue Database},
	Year = {2022}}

@inproceedings{zadeh2020cmu,
	Author = {Zadeh, Amir and Cao, Yan Sheng and Hessner, Simon and Liang, Paul Pu and Poria, Soujanya and Morency, Louis-Philippe},
	Booktitle = {Proceedings of the Conference on Empirical Methods in Natural Language Processing. Conference on Empirical Methods in Natural Language Processing},
	Pages = {1801},
	Title = {CMU-MOSEAS: A multimodal language dataset for Spanish, Portuguese, German and French},
	Volume = {2020},
	Year = {2020}}

@inproceedings{xu2022mcpr,
	Author = {Xu, Carol and Luo, Xuan and Wang, Dan},
	Booktitle = {International Conference on Cognitive Computing},
	Organization = {Springer},
	Pages = {83--90},
	Title = {MCPR: A Chinese Product Review Dataset for Multimodal Aspect-Based Sentiment Analysis},
	Year = {2022}}

@article{DBLP:journals/taslp/ChenHGS23,
	Author = {Chen Chen and Hansheng Hong and Jie Guo and Bin Song},
	Journal = {{IEEE} {ACM} Trans. Audio Speech Lang. Process.},
	Pages = {1476--1488},
	Title = {Inter-Intra Modal Representation Augmentation With Trimodal Collaborative Disentanglement Network for Multimodal Sentiment Analysis},
	Volume = {31},
	Year = {2023}}

@inproceedings{DBLP:conf/mm/YangXG20,
	Author = {Kaicheng Yang and Hua Xu and Kai Gao},
	Booktitle = {{MM} '20: The 28th {ACM} International Conference on Multimedia, Virtual Event / Seattle, WA, USA, October 12-16, 2020},
	Pages = {521--528},
	Title = {{CM-BERT:} Cross-Modal {BERT} for Text-Audio Sentiment Analysis},
	Year = {2020}}

@inproceedings{DBLP:conf/coling/LinLLDYZX22,
	Author = {Zijie Lin and Bin Liang and Yunfei Long and Yixue Dang and Min Yang and Min Zhang and Ruifeng Xu},
	Booktitle = {Proceedings of the 29th International Conference on Computational Linguistics, {COLING} 2022, Gyeongju, Republic of Korea, October 12-17, 2022},
	Pages = {7124--7135},
	Title = {Modeling Intra- and Inter-Modal Relations: Hierarchical Graph Contrastive Learning for Multimodal Sentiment Analysis},
	Year = {2022}}

@inproceedings{DBLP:conf/naacl/LiXZZ22,
	Author = {Zhen Li and Bing Xu and Conghui Zhu and Tiejun Zhao},
	Booktitle = {Findings of the Association for Computational Linguistics: {NAACL} 2022, Seattle, WA, United States, July 10-15, 2022},
	Pages = {2282--2294},
	Title = {{CLMLF:} {A} Contrastive Learning and Multi-Layer Fusion Method for Multimodal Sentiment Detection},
	Year = {2022}}

@inproceedings{DBLP:conf/acl/MaiHX19,
	Author = {Sijie Mai and Haifeng Hu and Songlong Xing},
	Booktitle = {Proceedings of the 57th Conference of the Association for Computational Linguistics, {ACL} 2019, Florence, Italy, July 28- August 2, 2019, Volume 1: Long Papers},
	Pages = {481--492},
	Title = {Divide, Conquer and Combine: Hierarchical Feature Fusion Network with Local and Global Perspectives for Multimodal Affective Computing},
	Year = {2019}}

@inproceedings{DBLP:conf/acl/ZhengYXW23,
	Author = {Wenjie Zheng and Jianfei Yu and Rui Xia and Shijin Wang},
	Booktitle = {Proceedings of the 61st Annual Meeting of the Association for Computational Linguistics (Volume 1: Long Papers), {ACL} 2023, Toronto, Canada, July 9-14, 2023},
	Pages = {15445--15459},
	Title = {A Facial Expression-Aware Multimodal Multi-task Learning Framework for Emotion Recognition in Multi-party Conversations},
	Year = {2023}}

@article{DBLP:journals/inffus/ZhangWLRZSTQ23,
	Author = {Yazhou Zhang and Jinglin Wang and Yaochen Liu and Lu Rong and Qian Zheng and Dawei Song and Prayag Tiwari and Jing Qin},
	Journal = {Inf. Fusion},
	Pages = {282--301},
	Title = {A Multitask learning model for multimodal sarcasm, sentiment and emotion recognition in conversations},
	Volume = {93},
	Year = {2023}}

@article{DBLP:journals/corr/abs-2310-20494,
	  author       = {Hui Ma and
                  Jian Wang and
                  Hongfei Lin and
                  Bo Zhang and
                  Yi{-}Jia Zhang and
                  Bo Xu},
  title        = {A Transformer-Based Model With Self-Distillation for Multimodal Emotion
                  Recognition in Conversations},
  journal      = {{IEEE} Trans. Multim.},
  volume       = {26},
  pages        = {776--788},
  year         = {2024},
  url          = {https://doi.org/10.1109/TMM.2023.3271019},
  doi          = {10.1109/TMM.2023.3271019},
  bibsource    = {dblp computer science bibliography, https://dblp.org}}

@inproceedings{DBLP:conf/acl/ShiH23,
	Author = {Tao Shi and Shao{-}Lun Huang},
	Booktitle = {Proceedings of the 61st Annual Meeting of the Association for Computational Linguistics (Volume 1: Long Papers), {ACL} 2023, Toronto, Canada, July 9-14, 2023},
	Pages = {14752--14766},
	Title = {MultiEMO: An Attention-Based Correlation-Aware Multimodal Fusion Framework for Emotion Recognition in Conversations},
	Year = {2023}}

@inproceedings{DBLP:conf/acl/LiZLZYH23,
	Author = {Ziming Li and Yan Zhou and Yaxin Liu and Fuqing Zhu and Chuanpeng Yang and Songlin Hu},
	Booktitle = {Findings of the Association for Computational Linguistics: {ACL} 2023, Toronto, Canada, July 9-14, 2023},
	Pages = {12191--12204},
	Title = {{QAP:} {A} Quantum-Inspired Adaptive-Priority-Learning Model for Multimodal Emotion Recognition},
	Year = {2023}}

@inproceedings{DBLP:conf/icml/RadfordKHRGASAM21,
	Author = {Alec Radford and Jong Wook Kim and Chris Hallacy and Aditya Ramesh and Gabriel Goh and Sandhini Agarwal and Girish Sastry and Amanda Askell and Pamela Mishkin and Jack Clark and Gretchen Krueger and Ilya Sutskever},
	Booktitle = {Proceedings of the 38th International Conference on Machine Learning, {ICML} 2021, 18-24 July 2021, Virtual Event},
	Pages = {8748--8763},
	Title = {Learning Transferable Visual Models From Natural Language Supervision},
	Year = {2021}}

@inproceedings{DBLP:conf/nips/BaoW0LMASPW22,
	Author = {Hangbo Bao and Wenhui Wang and Li Dong and Qiang Liu and Owais Khan Mohammed and Kriti Aggarwal and Subhojit Som and Songhao Piao and Furu Wei},
	Booktitle = {NeurIPS},
	Title = {VLMo: Unified Vision-Language Pre-Training with Mixture-of-Modality-Experts},
	Year = {2022}}

@article{DBLP:journals/tmlr/YuWVYSW22,
	Author = {Jiahui Yu and Zirui Wang and Vijay Vasudevan and Legg Yeung and Mojtaba Seyedhosseini and Yonghui Wu},
	Journal = {Trans. Mach. Learn. Res.},
	Title = {CoCa: Contrastive Captioners are Image-Text Foundation Models},
	Volume = {2022},
	Year = {2022}}

@article{DBLP:journals/corr/abs-2109-05522,
	Author = {Mehdi Arjmand and Mohammad Javad Dousti and Hadi Moradi},
	Journal = {CoRR},
	Title = {{TEASEL:} {A} Transformer-Based Speech-Prefixed Language Model},
	Volume = {abs/2109.05522},
	Year = {2021}}

@inproceedings{DBLP:conf/nips/AkbariYQCCCG21,
	Author = {Hassan Akbari and Liangzhe Yuan and Rui Qian and Wei{-}Hong Chuang and Shih{-}Fu Chang and Yin Cui and Boqing Gong},
	Booktitle = {Advances in Neural Information Processing Systems 34: Annual Conference on Neural Information Processing Systems 2021, NeurIPS 2021, December 6-14, 2021, virtual},
	Pages = {24206--24221},
	Title = {{VATT:} Transformers for Multimodal Self-Supervised Learning from Raw Video, Audio and Text},
	Year = {2021}}

@inproceedings{DBLP:conf/ijcai/Yu019,
	Author = {Jianfei Yu and Jing Jiang},
	Booktitle = {Proceedings of the Twenty-Eighth International Joint Conference on Artificial Intelligence, {IJCAI} 2019, Macao, China, August 10-16, 2019},
	Pages = {5408--5414},
	Title = {Adapting {BERT} for Target-Oriented Multimodal Sentiment Classification},
	Year = {2019}}

@inproceedings{DBLP:conf/aaai/0001FLH18,
	Author = {Qi Zhang and Jinlan Fu and Xiaoyu Liu and Xuanjing Huang},
	Booktitle = {Proceedings of the Thirty-Second {AAAI} Conference on Artificial Intelligence, (AAAI-18), the 30th innovative Applications of Artificial Intelligence (IAAI-18), and the 8th {AAAI} Symposium on Educational Advances in Artificial Intelligence (EAAI-18), New Orleans, Louisiana, USA, February 2-7, 2018},
	Pages = {5674--5681},
	Title = {Adaptive Co-attention Network for Named Entity Recognition in Tweets},
	Year = {2018}}

@inproceedings{DBLP:conf/acl/JiZCLN18,
	Author = {Di Lu and Leonardo Neves and Vitor Carvalho and Ning Zhang and Heng Ji},
	Booktitle = {Proceedings of the 56th Annual Meeting of the Association for Computational Linguistics, {ACL} 2018, Melbourne, Australia, July 15-20, 2018, Volume 1: Long Papers},
	Pages = {1990--1999},
	Title = {Visual Attention Model for Name Tagging in Multimodal Social Media},
	Year = {2018}}

@article{DBLP:journals/tkde/ZhangY22,
	Author = {Yu Zhang and Qiang Yang},
	Journal = {{IEEE} Trans. Knowl. Data Eng.},
	Number = {12},
	Pages = {5586--5609},
	Title = {A Survey on Multi-Task Learning},
	Volume = {34},
	Year = {2022}}

@article{ge2020mutual,
	Author = {Ge, Yixiao and Chen, Dapeng and Li, Hongsheng},
	Journal = {arXiv preprint arXiv:2001.01526},
	Title = {Mutual mean-teaching: Pseudo label refinery for unsupervised domain adaptation on person re-identification},
	Year = {2020}}

@article{zheng2022multimodal,
	Author = {Zheng, Jiahao and Zhang, Sen and Wang, Xiaoping and Zeng, Zhigang},
	Journal = {arXiv preprint arXiv:2201.03969},
	Title = {Multimodal representations learning based on mutual information maximization and minimization and identity embedding for multimodal sentiment analysis},
	Year = {2022}}

@article{DBLP:journals/pami/ZengPRH09,
	Author = {Zhihong Zeng and Maja Pantic and Glenn I. Roisman and Thomas S. Huang},
	Journal = {{IEEE} Trans. Pattern Anal. Mach. Intell.},
	Number = {1},
	Pages = {39--58},
	Title = {A Survey of Affect Recognition Methods: Audio, Visual, and Spontaneous Expressions},
	Volume = {31},
	Year = {2009}}

@inproceedings{DBLP:conf/acii/RamirezBM11,
	Author = {Geovany A. Ram{\'{\i}}rez and Tadas Baltrusaitis and Louis{-}Philippe Morency},
	Booktitle = {Affective Computing and Intelligent Interaction - Fourth International Conference, {ACII} 2011, Memphis, TN, USA, October 9-12, 2011, Proceedings, Part {II}},
	Pages = {396--406},
	Title = {Modeling Latent Discriminative Dynamic of Multi-dimensional Affective Signals},
	Year = {2011}}

@article{koelstra2011deap,
	Author = {Koelstra, Sander and Muhl, Christian and Soleymani, Mohammad and Lee, Jong-Seok and Yazdani, Ashkan and Ebrahimi, Touradj and Pun, Thierry and Nijholt, Anton and Patras, Ioannis},
	Journal = {IEEE transactions on affective computing},
	Number = {1},
	Pages = {18--31},
	Publisher = {IEEE},
	Title = {Deap: A database for emotion analysis; using physiological signals},
	Volume = {3},
	Year = {2011}}

@inproceedings{xu2019multi,
	Author = {Xu, Nan and Mao, Wenji and Chen, Guandan},
	Booktitle = {Proceedings of the AAAI Conference on Artificial Intelligence},
	Number = {01},
	Pages = {371--378},
	Title = {Multi-interactive memory network for aspect based multimodal sentiment analysis},
	Volume = {33},
	Year = {2019}}

@article{DBLP:journals/inffus/EzzameliM23,
	Author = {Kaouther Ezzameli and Hela Mahersia},
	Journal = {Inf. Fusion},
	Pages = {101847},
	Title = {Emotion recognition from unimodal to multimodal analysis: {A} review},
	Volume = {99},
	Year = {2023}}

@article{DBLP:journals/ijon/PanHJD23,
	Author = {Bei Pan and Kaoru Hirota and Zhiyang Jia and Yaping Dai},
	Journal = {Neurocomputing},
	Pages = {126866},
	Title = {A review of multimodal emotion recognition from datasets, preprocessing, features, and fusion methods},
	Volume = {561},
	Year = {2023}}

@article{DBLP:journals/tcsv/RenHLLLL23,
	Author = {Minjie Ren and Xiangdong Huang and Jing Liu and Ming Liu and Xuanya Li and An{-}An Liu},
	Journal = {{IEEE} Trans. Circuits Syst. Video Technol.},
	Number = {11},
	Pages = {6965--6980},
	Title = {{MALN:} Multimodal Adversarial Learning Network for Conversational Emotion Recognition},
	Volume = {33},
	Year = {2023}}

@inproceedings{DBLP:conf/cvpr/LiW023,
	Author = {Yong Li and Yuanzhi Wang and Zhen Cui},
	Booktitle = {{IEEE/CVF} Conference on Computer Vision and Pattern Recognition, {CVPR} 2023, Vancouver, BC, Canada, June 17-24, 2023},
	Pages = {6631--6640},
	Title = {Decoupled Multimodal Distilling for Emotion Recognition},
	Year = {2023}}

@inproceedings{DBLP:conf/ccis/SunHTZS23,
	Author = {Xinwei Sun and Huijie He and Haoyang Tang and Kai Zeng and Tao Shen},
	Booktitle = {9th {IEEE} International Conference on Cloud Computing and Intelligent Systems, {CCIS} 2023, Dali, China, August 12-13, 2023},
	Pages = {250--259},
	Title = {Multimodal rough set transformer for sentiment analysis and emotion recognition},
	Year = {2023}}

@article{DBLP:journals/corr/abs-2310-04456,
	Author = {Shihao Zou and Xianying Huang and Xudong Shen},
	Journal = {CoRR},
	Title = {Multimodal Prompt Transformer with Hybrid Contrastive Learning for Emotion Recognition in Conversation},
	Volume = {abs/2310.04456},
	Year = {2023}}

@article{DBLP:journals/corr/abs-2309-02106,
	Author = {Peiying Wang and Sunlu Zeng and Junqing Chen and Lu Fan and Meng Chen and Youzheng Wu and Xiaodong He},
	Journal = {CoRR},
	Title = {Leveraging Label Information for Multimodal Emotion Recognition},
	Volume = {abs/2309.02106},
	Year = {2023}}

@article{DBLP:journals/corr/abs-2207-12895,
	Author = {Yoonhyung Lee and Seunghyun Yoon and Kyomin Jung},
	Journal = {CoRR},
	Title = {Multimodal Speech Emotion Recognition using Cross Attention with Aligned Audio and Text},
	Volume = {abs/2207.12895},
	Year = {2022}}

@article{DBLP:journals/corr/abs-2302-13661,
	Author = {Dekai Sun and Yancheng He and Jiqing Han},
	Journal = {CoRR},
	Title = {Using Auxiliary Tasks In Multimodal Fusion Of Wav2vec 2.0 And {BERT} For Multimodal Emotion Recognition},
	Volume = {abs/2302.13661},
	Year = {2023}}

@article{DBLP:journals/entropy/LiuSFWZQ22,
	Author = {Feng Liu and Siyuan Shen and Ziwang Fu and Hanyang Wang and Aimin Zhou and Jia{-}Yin Qi},
	Journal = {Entropy},
	Number = {7},
	Pages = {1010},
	Title = {{LGCCT:} {A} Light Gated and Crossed Complementation Transformer for Multimodal Speech Emotion Recognition},
	Volume = {24},
	Year = {2022}}

@inproceedings{DBLP:conf/acl/QiuSS23,
	Author = {Shuwen Qiu and Nitesh Sekhar and Prateek Singhal},
	Booktitle = {Findings of the Association for Computational Linguistics: {ACL} 2023, Toronto, Canada, July 9-14, 2023},
	Pages = {2074--2082},
	Title = {Topic and Style-aware Transformer for Multimodal Emotion Recognition},
	Year = {2023}}

@article{DBLP:journals/ieeemm/FuOWGSLD22,
	Author = {Yahui Fu and Shogo Okada and Longbiao Wang and Lili Guo and Yaodong Song and Jiaxing Liu and Jianwu Dang},
	Journal = {{IEEE} Multim.},
	Number = {3},
	Pages = {91--100},
	Title = {Context- and Knowledge-Aware Graph Convolutional Network for Multimodal Emotion Recognition},
	Volume = {29},
	Year = {2022}}

@article{DBLP:journals/corr/abs-2211-15425,
	Author = {Zhongyu Fang and Aoyun He and Qihui Yu and Baopeng Gao and Weiping Ding and Tong Zhang and Lei Ma},
	Journal = {CoRR},
	Title = {{FAF:} {A} novel multimodal emotion recognition approach integrating face, body and text},
	Volume = {abs/2211.15425},
	Year = {2022}}

@inproceedings{DBLP:conf/icassp/ZuoLZGL23,
	Author = {Haolin Zuo and Rui Liu and Jinming Zhao and Guanglai Gao and Haizhou Li},
	Booktitle = {{IEEE} International Conference on Acoustics, Speech and Signal Processing {ICASSP} 2023, Rhodes Island, Greece, June 4-10, 2023},
	Pages = {1--5},
	Title = {Exploiting Modality-Invariant Feature for Robust Multimodal Emotion Recognition with Missing Modalities},
	Year = {2023}}

@article{DBLP:journals/bspc/SharafiYRN22,
	Author = {Masoumeh Sharafi and Mohammadreza Yazdchi and Reza Rasti and Fahimeh Nasimi},
	Journal = {Biomed. Signal Process. Control.},
	Pages = {103970},
	Title = {A novel spatio-temporal convolutional neural framework for multimodal emotion recognition},
	Volume = {78},
	Year = {2022}}

@article{DBLP:journals/inffus/ZhangLLHDZ22,
	Author = {Feng Zhang and Xi{-}Cheng Li and Chee Peng Lim and Qiang Hua and Chun{-}Ru Dong and Jun{-}Hai Zhai},
	Journal = {Inf. Fusion},
	Pages = {296--304},
	Title = {Deep Emotional Arousal Network for Multimodal Sentiment Analysis and Emotion Recognition},
	Volume = {88},
	Year = {2022}}

@article{DBLP:journals/kbs/MiddyaNR22,
	Author = {Asif Iqbal Middya and Baibhav Nag and Sarbani Roy},
	Journal = {Knowl. Based Syst.},
	Pages = {108580},
	Title = {Deep learning based multimodal emotion recognition using model-level fusion of audio-visual modalities},
	Volume = {244},
	Year = {2022}}

@article{DBLP:journals/tmm/WangG08,
	Author = {Yongjin Wang and Ling Guan},
	Journal = {{IEEE} Trans. Multim.},
	Number = {4},
	Pages = {659--668},
	Title = {Recognizing Human Emotional State From Audiovisual Signals},
	Volume = {10},
	Year = {2008}}

@article{DBLP:journals/taffco/ZhalehpourOAE17,
	Author = {Sara Zhalehpour and Onur Onder and Zahid Akhtar and Cigdem Eroglu Erdem},
	Journal = {{IEEE} Trans. Affect. Comput.},
	Number = {3},
	Pages = {300--313},
	Title = {{BAUM-1:} {A} Spontaneous Audio-Visual Face Database of Affective and Mental States},
	Volume = {8},
	Year = {2017}}

@article{DBLP:journals/taslp/ZhaoWSXZ23,
	Author = {Zhengdao Zhao and Yuhua Wang and Guang Shen and Yuezhu Xu and Jiayuan Zhang},
	Journal = {{IEEE} {ACM} Trans. Audio Speech Lang. Process.},
	Pages = {3771--3782},
	Title = {TDFNet: Transformer-Based Deep-Scale Fusion Network for Multimodal Emotion Recognition},
	Volume = {31},
	Year = {2023}}

@article{DBLP:journals/taffco/SoleymaniLPP12,
	Author = {Mohammad Soleymani and Jeroen Lichtenauer and Thierry Pun and Maja Pantic},
	Journal = {{IEEE} Trans. Affect. Comput.},
	Number = {1},
	Pages = {42--55},
	Title = {A Multimodal Database for Affect Recognition and Implicit Tagging},
	Volume = {3},
	Year = {2012}}

@article{DBLP:journals/corr/abs-2310-05804,
	Author = {Haoyu Zhang and Yu Wang and Guanghao Yin and Kejun Liu and Yuanyuan Liu and Tianshu Yu},
	Journal = {CoRR},
	Title = {Learning Language-guided Adaptive Hyper-modality Representation for Multimodal Sentiment Analysis},
	Volume = {abs/2310.05804},
	Year = {2023}}

@article{DBLP:journals/tai/WangCZCYZ23,
	Author = {Ning Wang and Hui Cao and Jun Zhao and Ruilin Chen and Dapeng Yan and Jie Zhang},
	Journal = {{IEEE} Trans. Artif. Intell.},
	Number = {5},
	Pages = {1305--1316},
	Title = {{M2R2:} Missing-Modality Robust Emotion Recognition Framework With Iterative Data Augmentation},
	Volume = {4},
	Year = {2023}}

@inproceedings{DBLP:conf/acl/ZhaoLJ20,
	Author = {Jinming Zhao and Ruichen Li and Qin Jin},
	Booktitle = {Proceedings of the 59th Annual Meeting of the Association for Computational Linguistics and the 11th International Joint Conference on Natural Language Processing, {ACL/IJCNLP} 2021, (Volume 1: Long Papers), Virtual Event, August 1-6, 2021},
	Pages = {2608--2618},
	Title = {Missing Modality Imagination Network for Emotion Recognition with Uncertain Missing Modalities},
	Year = {2021}}

@inproceedings{DBLP:conf/icmcs/MaHZ21,
	Author = {Fei Ma and Shao{-}Lun Huang and Lin Zhang},
	Booktitle = {2021 {IEEE} International Conference on Multimedia and Expo, {ICME} 2021, Shenzhen, China, July 5-9, 2021},
	Pages = {1--6},
	Title = {An Efficient Approach for Audio-Visual Emotion Recognition With Missing Labels And Missing Modalities},
	Year = {2021}}

@article{DBLP:journals/ipm/XiaoWYXZH23,
	Author = {Luwei Xiao and Xingjiao Wu and Shuwen Yang and Junjie Xu and Jie Zhou and Liang He},
	Journal = {Inf. Process. Manag.},
	Number = {6},
	Pages = {103508},
	Title = {Cross-modal fine-grained alignment and fusion network for multimodal aspect-based sentiment analysis},
	Volume = {60},
	Year = {2023}}

@inproceedings{DBLP:conf/mrac/LianLLZTZZ23,
	Author = {Hailun Lian and Cheng Lu and Sunan Li and Yan Zhao and Chuangao Tang and Yuan Zong and Wenming Zheng},
	Booktitle = {Proceedings of the 1st International Workshop on Multimodal and Responsible Affective Computing, {MRAC} 2023, Ottawa, ON, Canada, 29 October 2023},
	Pages = {51--58},
	Title = {Label Distribution Adaptation for Multimodal Emotion Recognition with Multi-label Learning},
	Year = {2023}}

@article{wang2023image,
	Author = {Wang, Qianlong and Xu, Hongling and Wen, Zhiyuan and Liang, Bin and Yang, Min and Qin, Bing and Xu, Ruifeng},
	Journal = {IEEE Transactions on Affective Computing},
	Number = {01},
	Pages = {1--15},
	Publisher = {IEEE Computer Society},
	Title = {Image-to-Text Conversion and Aspect-Oriented Filtration for Multimodal Aspect-Based Sentiment Analysis},
	Year = {2023}}

@article{DBLP:journals/inffus/LiuZCSM24,
	Author = {Zhizhong Liu and Bin Zhou and Dianhui Chu and Yuhang Sun and Lingqiang Meng},
	Journal = {Inf. Fusion},
	Pages = {101973},
	Title = {Modality translation-based multimodal sentiment analysis under uncertain missing modalities},
	Volume = {101},
	Year = {2024}}

@article{DBLP:journals/inffus/ZengYMH24,
	Author = {Ying Zeng and Wenjun Yan and Sijie Mai and Haifeng Hu},
	Journal = {Inf. Fusion},
	Pages = {102031},
	Title = {Disentanglement Translation Network for multimodal sentiment analysis},
	Volume = {102},
	Year = {2024}}

@article{DBLP:journals/cogcom/WangTYLWLW23,
	Author = {Fan Wang and Shengwei Tian and Long Yu and Jing Liu and Junwen Wang and Kun Li and Yongtao Wang},
	Journal = {Cogn. Comput.},
	Number = {1},
	Pages = {289--303},
	Title = {{TEDT:} Transformer-Based Encoding-Decoding Translation Network for Multimodal Sentiment Analysis},
	Volume = {15},
	Year = {2023}}

@article{DBLP:journals/eswa/PengWZCTYH23,
	Author = {Junjie Peng and Ting Wu and Wenqiang Zhang and Feng Cheng and Shuhua Tan and Fen Yi and Yansong Huang},
	Journal = {Expert Syst. Appl.},
	Pages = {119721},
	Title = {A fine-grained modal label-based multi-stage network for multimodal sentiment analysis},
	Volume = {221},
	Year = {2023}}

@article{DBLP:journals/inffus/LiGPDYZLCWX23,
	Author = {Zuhe Li and Qingbing Guo and Yushan Pan and Weiping Ding and Jun Yu and Yazhou Zhang and Weihua Liu and Haoran Chen and Hao Wang and Ying Xie},
	Journal = {Inf. Fusion},
	Pages = {101891},
	Title = {Multi-level correlation mining framework with self-supervised label generation for multimodal sentiment analysis},
	Volume = {99},
	Year = {2023}}

@article{DBLP:journals/inffus/GandhiAPCH23,
	Author = {Ankita Gandhi and Kinjal Adhvaryu and Soujanya Poria and Erik Cambria and Amir Hussain},
	Journal = {Inf. Fusion},
	Pages = {424--444},
	Title = {Multimodal sentiment analysis: {A} systematic review of history, datasets, multimodal fusion methods, applications, challenges and future directions},
	Volume = {91},
	Year = {2023}}

@article{DBLP:journals/inffus/ZhuCZSLLC23,
	Author = {Chuanbo Zhu and Min Chen and Sheng Zhang and Chao Sun and Han Liang and Yifan Liu and Jincai Chen},
	Journal = {Inf. Fusion},
	Pages = {101958},
	Title = {{SKEAFN:} Sentiment Knowledge Enhanced Attention Fusion Network for multimodal sentiment analysis},
	Volume = {100},
	Year = {2023}}

@article{DBLP:journals/kbs/HuangZWW0H23,
	Author = {Changqin Huang and Junling Zhang and Xuemei Wu and Yi Wang and Ming Li and Xiaodi Huang},
	Journal = {Knowl. Based Syst.},
	Pages = {110502},
	Title = {TeFNA: Text-centered fusion network with crossmodal attention for multimodal sentiment analysis},
	Volume = {269},
	Year = {2023}}

@article{DBLP:journals/pr/WangGTLHL23,
	Author = {Di Wang and Xutong Guo and Yumin Tian and Jinhui Liu and Lihuo He and Xuemei Luo},
	Journal = {Pattern Recognit.},
	Pages = {109259},
	Title = {{TETFN:} {A} text enhanced transformer fusion network for multimodal sentiment analysis},
	Volume = {136},
	Year = {2023}}

@article{DBLP:journals/taffco/StappenBSS23,
	Author = {Lukas Stappen and Alice Baird and Lea Schumann and Bj{\"{o}}rn W. Schuller},
	Journal = {{IEEE} Trans. Affect. Comput.},
	Number = {2},
	Pages = {1334--1350},
	Title = {The Multimodal Sentiment Analysis in Car Reviews (MuSe-CaR) Dataset: Collection, Insights and Improvements},
	Volume = {14},
	Year = {2023}}

@article{DBLP:journals/taslp/HuangQTX23,
	Author = {Maochun Huang and Chunmei Qing and Junpeng Tan and Xiangmin Xu},
	Journal = {{IEEE} {ACM} Trans. Audio Speech Lang. Process.},
	Pages = {3468--3477},
	Title = {Context-Based Adaptive Multimodal Fusion Network for Continuous Frame-Level Sentiment Prediction},
	Volume = {31},
	Year = {2023}}

@article{DBLP:journals/tcsv/TangLJPZDK23,
	Author = {Jiajia Tang and Dongjun Liu and Xuanyu Jin and Yong Peng and Qibin Zhao and Yu Ding and Wanzeng Kong},
	Journal = {{IEEE} Trans. Circuits Syst. Video Technol.},
	Number = {4},
	Pages = {1966--1978},
	Title = {{BAFN:} Bi-Direction Attention Based Fusion Network for Multimodal Sentiment Analysis},
	Volume = {33},
	Year = {2023}}

@inproceedings{DBLP:conf/aaai/MaoZXYL23,
	Author = {Huisheng Mao and Baozheng Zhang and Hua Xu and Ziqi Yuan and Yihe Liu},
	Booktitle = {Thirty-Seventh {AAAI} Conference on Artificial Intelligence, {AAAI} 2023, Thirty-Fifth Conference on Innovative Applications of Artificial Intelligence, {IAAI} 2023, Thirteenth Symposium on Educational Advances in Artificial Intelligence, {EAAI} 2023, Washington, DC, USA, February 7-14, 2023},
	Pages = {16458--16460},
	Title = {Robust-MSA: Understanding the Impact of Modality Noise on Multimodal Sentiment Analysis},
	Year = {2023}}

@article{DBLP:journals/expert/WollmerWKSSSM13,
	Author = {Martin W{\"{o}}llmer and Felix Weninger and Tobias Knaup and Bj{\"{o}}rn W. Schuller and Congkai Sun and Kenji Sagae and Louis{-}Philippe Morency},
	Journal = {{IEEE} Intell. Syst.},
	Number = {3},
	Pages = {46--53},
	Title = {YouTube Movie Reviews: Sentiment Analysis in an Audio-Visual Context},
	Volume = {28},
	Year = {2013}}

@article{DBLP:journals/pami/LiSZT22,
	Author = {Zechao Li and Yanpeng Sun and Liyan Zhang and Jinhui Tang},
	Journal = {{IEEE} Trans. Pattern Anal. Mach. Intell.},
	Number = {12},
	Pages = {9904--9917},
	Title = {CTNet: Context-Based Tandem Network for Semantic Segmentation},
	Volume = {44},
	Year = {2022}}

@inproceedings{DBLP:conf/emnlp/ZhaoLWOZD23,
	Author = {Fei Zhao and Chunhui Li and Zhen Wu and Yawen Ouyang and Jianbing Zhang and Xinyu Dai},
	Booktitle = {Proceedings of the 2023 Conference on Empirical Methods in Natural Language Processing, {EMNLP} 2023, Singapore, December 6-10, 2023},
	Pages = {9057--9070},
	Title = {{M2DF:} Multi-grained Multi-curriculum Denoising Framework for Multimodal Aspect-based Sentiment Analysis},
	Year = {2023}}

@inproceedings{zheng2023facial,
	Author = {Zheng, Wenjie and Yu, Jianfei and Xia, Rui and Wang, Shijin},
	Booktitle = {Proceedings of the 61st Annual Meeting of the Association for Computational Linguistics (Volume 1: Long Papers)},
	Pages = {15445--15459},
	Title = {A Facial Expression-Aware Multimodal Multi-task Learning Framework for Emotion Recognition in Multi-party Conversations},
	Year = {2023}}

@article{li2022emocaps,
	Author = {Li, Zaijing and Tang, Fengxiao and Zhao, Ming and Zhu, Yusen},
	Journal = {arXiv preprint arXiv:2203.13504},
	Title = {Emocaps: Emotion capsule based model for conversational emotion recognition},
	Year = {2022}}

@inproceedings{li2018mec,
	Author = {Li, Ya and Tao, Jianhua and Schuller, Bj{\"o}rn and Shan, Shiguang and Jiang, Dongmei and Jia, Jia},
	Booktitle = {2018 First Asian Conference on Affective Computing and Intelligent Interaction (ACII Asia)},
	Pages = {1--5},
	Title = {Mec 2017: Multimodal emotion recognition challenge},
	Year = {2018}}

@article{7374697,
	Author = {Busso, Carlos and Parthasarathy, Srinivas and Burmania, Alec and AbdelWahab, Mohammed and Sadoughi, Najmeh and Provost, Emily Mower},
	Journal = {IEEE Transactions on Affective Computing},
	Number = {1},
	Pages = {67-80},
	Title = {MSP-IMPROV: An Acted Corpus of Dyadic Interactions to Study Emotion Perception},
	Volume = {8},
	Year = {2017}}

@inproceedings{DBLP:conf/nips/MikolovSCCD13,
	Author = {Tomas Mikolov and Ilya Sutskever and Kai Chen and Gregory S. Corrado and Jeffrey Dean},
	Booktitle = {Advances in Neural Information Processing Systems 26: 27th Annual Conference on Neural Information Processing Systems 2013. Proceedings of a meeting held December 5-8, 2013, Lake Tahoe, Nevada, United States},
	Pages = {3111--3119},
	Title = {Distributed Representations of Words and Phrases and their Compositionality},
	Year = {2013}}

@inproceedings{ji2020diversified,
	Author = {Ji, Yunjie and Liu, Hao and He, Bolei and Xiao, Xinyan and Wu, Hua and Yu, Yanhua},
	Booktitle = {Proceedings of the 2020 conference on empirical methods in natural language processing (EMNLP)},
	Pages = {7012--7023},
	Title = {Diversified multiple instance learning for document-level multi-aspect sentiment classification},
	Year = {2020}}

@inproceedings{DBLP:conf/icfsp/DonnellyP22,
	Author = {Patrick J. Donnelly and Alex Prestwich},
	Booktitle = {7th International Conference on Frontiers of Signal Processing, {ICFSP} 2022, Paris, France, September 7-9, 2022},
	Pages = {64--69},
	Title = {Identifying Sentiment from Crowd Audio},
	Year = {2022}}

@inproceedings{DBLP:conf/bigdataconf/ShangPSCLB17,
	Author = {Chao Shang and Aaron Palmer and Jiangwen Sun and Ko{-}Shin Chen and Jin Lu and Jinbo Bi},
	Booktitle = {2017 {IEEE} International Conference on Big Data {(IEEE} BigData 2017), Boston, MA, USA, December 11-14, 2017},
	Pages = {766--775},
	Title = {{VIGAN:} Missing view imputation with generative adversarial networks},
	Year = {2017}}

@article{DBLP:journals/ijon/ZhouCVR21,
	Author = {Tongxue Zhou and St{\'{e}}phane Canu and Pierre Vera and Su Ruan},
	Journal = {Neurocomputing},
	Pages = {102--112},
	Title = {Feature-enhanced generation and multi-modality fusion based deep neural network for brain tumor segmentation with missing {MR} modalities},
	Volume = {466},
	Year = {2021}}

@inproceedings{DBLP:conf/icmi/ParthasarathyS20,
	Author = {Srinivas Parthasarathy and Shiva Sundaram},
	Booktitle = {Companion Publication of the 2020 International Conference on Multimodal Interaction, {ICMI} Companion 2020, Virtual Event, The Netherlands, October, 2020},
	Pages = {400--404},
	Title = {Training Strategies to Handle Missing Modalities for Audio-Visual Expression Recognition},
	Year = {2020}}

@inproceedings{DBLP:conf/mmm/LuoXL23,
	Author = {Wei Luo and Mengying Xu and Hanjiang Lai},
	Booktitle = {MultiMedia Modeling - 29th International Conference, {MMM} 2023, Bergen, Norway, January 9-12, 2023, Proceedings, Part {II}},
	Pages = {411--422},
	Title = {Multimodal Reconstruct and Align Net for Missing Modality Problem in Sentiment Analysis},
	Year = {2023}}

@inproceedings{DBLP:conf/sigir/ZengL022,
	Author = {Jiandian Zeng and Tianyi Liu and Jiantao Zhou},
	Booktitle = {{SIGIR} '22: The 45th International {ACM} {SIGIR} Conference on Research and Development in Information Retrieval, Madrid, Spain, July 11 - 15, 2022},
	Pages = {1545--1554},
	Title = {Tag-assisted Multimodal Sentiment Analysis under Uncertain Missing Modalities},
	Year = {2022}}

@article{zhang2024multi,
	Author = {Zhang, Xiaoheng and Cui, Weigang and Hu, Bin and Li, Yang},
	Journal = {IEEE Transactions on Affective Computing},
	Title = {A Multi-Level Alignment and Cross-Modal Unified Semantic Graph Refinement Network for Conversational Emotion Recognition},
	Year = {2024}}

@inproceedings{yu2022dual,
	Author = {Yu, Zhewen and Wang, Jin and Yu, Liang-Chih and Zhang, Xuejie},
	Booktitle = {Proceedings of the 2nd Conference of the Asia-Pacific Chapter of the Association for Computational Linguistics and the 12th International Joint Conference on Natural Language Processing (Volume 1: Long Papers)},
	Pages = {414--423},
	Title = {Dual-encoder transformers with cross-modal alignment for multimodal aspect-based sentiment analysis},
	Year = {2022}}

@article{lai2023shared,
	Author = {Lai, Songning and Hu, Xifeng and Li, Yulong and Ren, Zhaoxia and Liu, Zhi and Miao, Danmin},
	Journal = {arXiv preprint arXiv:2305.08473},
	Title = {Shared and private information learning in multimodal sentiment analysis with deep modal alignment and self-supervised multi-task learning},
	Year = {2023}}

@inproceedings{zhang-li-2023-cross,
	Address = {Toronto, Canada},
	Author = {Zhang, Xiaoheng and Li, Yang},
	Booktitle = {Proceedings of the 61st Annual Meeting of the Association for Computational Linguistics (Volume 1: Long Papers)},
	Doi = {10.18653/v1/2023.acl-long.732},
	Month = jul,
	Pages = {13099--13110},
	Publisher = {Association for Computational Linguistics},
	Title = {A Cross-Modality Context Fusion and Semantic Refinement Network for Emotion Recognition in Conversation},
	Url = {https://aclanthology.org/2023.acl-long.732},
	Year = {2023}}

@article{DBLP:journals/taffco/LiWLZ24,
	Author = {Jiang Li and Xiaoping Wang and Guoqing Lv and Zhigang Zeng},
	Journal = {{IEEE} Trans. Affect. Comput.},
	Number = {1},
	Pages = {130--143},
	Title = {{GA2MIF:} Graph and Attention Based Two-Stage Multi-Source Information Fusion for Conversational Emotion Detection},
	Volume = {15},
	Year = {2024}}

@article{majumder2018multimodal,
	Author = {Majumder, Navonil and Hazarika, Devamanyu and Gelbukh, Alexander and Cambria, Erik and Poria, Soujanya},
	Journal = {Knowledge-based systems},
	Pages = {124--133},
	Publisher = {Elsevier},
	Title = {Multimodal sentiment analysis using hierarchical fusion with context modeling},
	Volume = {161},
	Year = {2018}}

@article{DBLP:journals/tmm/ZhangCCCT24,
	Author = {Duzhen Zhang and Feilong Chen and Jianlong Chang and Xiuyi Chen and Qi Tian},
	Journal = {{IEEE} Trans. Multim.},
	Pages = {3987--3997},
	Title = {Structure Aware Multi-Graph Network for Multi-Modal Emotion Recognition in Conversations},
	Volume = {26},
	Year = {2024}}

@inproceedings{DBLP:conf/cvpr/Chen0ZS23,
	Author = {Feiyu Chen and Jie Shao and Shuyuan Zhu and Heng Tao Shen},
	Booktitle = {{IEEE/CVF} Conference on Computer Vision and Pattern Recognition, {CVPR} 2023, Vancouver, BC, Canada, June 17-24, 2023},
	Pages = {10761--10770},
	Title = {Multivariate, Multi-Frequency and Multimodal: Rethinking Graph Neural Networks for Emotion Recognition in Conversation},
	Year = {2023}}

@article{DBLP:journals/tois/ZhangJWZZLHJSQ24,
	Author = {Yazhou Zhang and Ao Jia and Bo Wang and Peng Zhang and Dongming Zhao and Pu Li and Yuexian Hou and Xiaojia Jin and Dawei Song and Jing Qin},
	Journal = {{ACM} Trans. Inf. Syst.},
	Number = {1},
	Pages = {13:1--13:32},
	Title = {{M3GAT:} {A} Multi-modal, Multi-task Interactive Graph Attention Network for Conversational Sentiment Analysis and Emotion Recognition},
	Volume = {42},
	Year = {2024}}

@inproceedings{DBLP:conf/naacl/JiaHZU0L22,
	Author = {Ao Jia and Yu He and Yazhou Zhang and Sagar Uprety and Dawei Song and Christina Lioma},
	Booktitle = {Proceedings of the 2022 Conference of the North American Chapter of the Association for Computational Linguistics: Human Language Technologies, {NAACL} 2022, Seattle, WA, United States, July 10-15, 2022},
	Pages = {1512--1522},
	Title = {Beyond Emotion: {A} Multi-Modal Dataset for Human Desire Understanding},
	Year = {2022}}

@inproceedings{firdaus2020meisd,
	Author = {Firdaus, Mauajama and Chauhan, Hardik and Ekbal, Asif and Bhattacharyya, Pushpak},
	Booktitle = {Proceedings of the 28th international conference on computational linguistics},
	Pages = {4441--4453},
	Title = {MEISD: A multimodal multi-label emotion, intensity and sentiment dialogue dataset for emotion recognition and sentiment analysis in conversations},
	Year = {2020}}

@article{DBLP:journals/corr/abs-2401-15164,
	Author = {Naresh Kumar Devulapally and Sidharth Anand and Sreyasee Das Bhattacharjee and Junsong Yuan and Yu{-}Ping Chang},
	Journal = {CoRR},
	Title = {AMuSE: Adaptive Multimodal Analysis for Speaker Emotion Recognition in Group Conversations},
	Volume = {abs/2401.15164},
	Year = {2024}}

@inproceedings{DBLP:conf/acii/VazquezRodriguezLCC23,
	Author = {Juan Vazquez{-}Rodriguez and Gr{\'{e}}goire Lefebvre and Julien Cumin and James L. Crowley},
	Booktitle = {11th International Conference on Affective Computing and Intelligent Interaction, {ACII} 2023, Cambridge, MA, USA, September 10-13, 2023},
	Pages = {1--8},
	Publisher = {{IEEE}},
	Title = {Accommodating Missing Modalities in Time-Continuous Multimodal Emotion Recognition},
	Year = {2023}}

@inproceedings{DBLP:conf/mm/StappenBCSSMCZS21,
	Author = {Lukas Stappen and Alice Baird and Lukas Christ and Lea Schumann and Benjamin Sertolli and Eva{-}Maria Me{\ss}ner and Erik Cambria and Guoying Zhao and Bj{\"{o}}rn W. Schuller},
	Booktitle = {MuSe '21: Proceedings of the 2nd on Multimodal Sentiment Analysis Challenge, Virtual Event, China, 24 October 2021},
	Pages = {5--14},
	Title = {The MuSe 2021 Multimodal Sentiment Analysis Challenge: Sentiment, Emotion, Physiological-Emotion, and Stress},
	Year = {2021}}

@inproceedings{DBLP:conf/acii/Vazquez-Rodriguez22,
	Author = {Juan Vazquez{-}Rodriguez and Gr{\'{e}}goire Lefebvre and Julien Cumin and James L. Crowley},
	Booktitle = {10th International Conference on Affective Computing and Intelligent Interaction, {ACII} 2022, Nara, Japan, October 18-21, 2022},
	Pages = {1--8},
	Title = {Emotion Recognition with Pre-Trained Transformers Using Multimodal Signals},
	Year = {2022}}

@article{DBLP:journals/taffco/CorreaASP21,
	Author = {Juan Abdon Miranda Correa and Mojtaba Khomami Abadi and Nicu Sebe and Ioannis Patras},
	Journal = {{IEEE} Trans. Affect. Comput.},
	Number = {2},
	Pages = {479--493},
	Title = {{AMIGOS:} {A} Dataset for Affect, Personality and Mood Research on Individuals and Groups},
	Volume = {12},
	Year = {2021}}

@inproceedings{DBLP:conf/www/WangW020,
	Author = {Zilong Wang and Zhaohong Wan and Xiaojun Wan},
	Booktitle = {{WWW} '20: The Web Conference 2020, Taipei, Taiwan, April 20-24, 2020},
	Pages = {2514--2520},
	Title = {TransModality: An End2End Fusion Method with Transformer for Multimodal Sentiment Analysis},
	Year = {2020}}

@inproceedings{du2018semi,
	Author = {Du, Changde and Du, Changying and Wang, Hao and Li, Jinpeng and Zheng, Wei-Long and Lu, Bao-Liang and He, Huiguang},
	Booktitle = {Proceedings of the 26th ACM international conference on Multimedia},
	Pages = {108--116},
	Title = {Semi-supervised deep generative modelling of incomplete multi-modality emotional data},
	Year = {2018}}

@inproceedings{10446459,
	Author = {Zhang, Chengwen and Zhang, Yuhao and Cheng, Bo},
	Booktitle = {ICASSP 2024 - 2024 IEEE International Conference on Acoustics, Speech and Signal Processing (ICASSP)},
	Pages = {10246-10250},
	Title = {RL-EMO: A Reinforcement Learning Framework for Multimodal Emotion Recognition},
	Year = {2024}}

@inproceedings{10447720,
	Author = {Yao, Biyun and Shi, Wuzhen},
	Booktitle = {ICASSP 2024 - 2024 IEEE International Conference on Acoustics, Speech and Signal Processing (ICASSP)},
	Doi = {10.1109/ICASSP48485.2024.10447720},
	Pages = {8441-8445},
	Title = {Speaker-Centric Multimodal Fusion Networks for Emotion Recognition in Conversations},
	Year = {2024}}

@article{li2024tscl,
	Author = {Li, Yuqiang and Weng, Wenxuan and Liu, Chun},
	Journal = {Neural Computing and Applications},
	Pages = {1--15},
	Publisher = {Springer},
	Title = {TSCL-FHFN: two-stage contrastive learning and feature hierarchical fusion network for multimodal sentiment analysis},
	Year = {2024}}

@inproceedings{sun2024novel,
	Author = {Sun, Xin and Ren, Xiangyu and Xie, Xiaohao},
	Booktitle = {ICASSP 2024-2024 IEEE International Conference on Acoustics, Speech and Signal Processing (ICASSP)},
	Organization = {IEEE},
	Pages = {8336--8340},
	Title = {A Novel Multimodal Sentiment Analysis Model Based on Gated Fusion and Multi-Task Learning},
	Year = {2024}}

@inproceedings{ruan2024fusing,
	Author = {Ruan, Yu-Ping and Han, Shoukang and Li, Taihao and Wu, Yanfeng},
	Booktitle = {ICASSP 2024-2024 IEEE International Conference on Acoustics, Speech and Signal Processing (ICASSP)},
	Organization = {IEEE},
	Pages = {7925--7929},
	Title = {Fusing Modality-Specific Representations and Decisions for Multimodal Emotion Recognition},
	Year = {2024}}

@inproceedings{chen2024mmrbn,
	Author = {Chen, Xi},
	Booktitle = {ICASSP 2024-2024 IEEE International Conference on Acoustics, Speech and Signal Processing (ICASSP)},
	Organization = {IEEE},
	Pages = {8200--8204},
	Title = {MMRBN: Rule-Based Network for Multimodal Emotion Recognition},
	Year = {2024}}

@inproceedings{poria2017context,
	Author = {Poria, Soujanya and Cambria, Erik and Hazarika, Devamanyu and Majumder, Navonil and Zadeh, Amir and Morency, Louis-Philippe},
	Booktitle = {Proceedings of the 55th annual meeting of the association for computational linguistics (volume 1: Long papers)},
	Pages = {873--883},
	Title = {Context-dependent sentiment analysis in user-generated videos},
	Year = {2017}}

@inproceedings{sun2023layer,
	Author = {Sun, Jun and Han, Shoukang and Ruan, Yu-Ping and Zhang, Xiaoning and Zheng, Shu-Kai and Liu, Yulong and Huang, Yuxin and Li, Taihao},
	Booktitle = {Proceedings of the 61st Annual Meeting of the Association for Computational Linguistics (Volume 1: Long Papers)},
	Pages = {658--670},
	Title = {Layer-wise fusion with modality independence modeling for multi-modal emotion recognition},
	Year = {2023}}

@inproceedings{liu2022make,
	Author = {Liu, Yihe and Yuan, Ziqi and Mao, Huisheng and Liang, Zhiyun and Yang, Wanqiuyue and Qiu, Yuanzhe and Cheng, Tie and Li, Xiaoteng and Xu, Hua and Gao, Kai},
	Booktitle = {Proceedings of the 2022 International Conference on Multimodal Interaction},
	Pages = {247--258},
	Title = {Make acoustic and visual cues matter: CH-SIMS v2. 0 dataset and AV-Mixup consistent module},
	Year = {2022}}

@inproceedings{2023GPT4VisionSC,
	Title = {GPT-4V(ision) System Card},
	Url = {https://api.semanticscholar.org/CorpusID:263218031},
	Year = {2023}}

@article{team2023gemini,
	Author = {Team, Gemini and Anil, Rohan and Borgeaud, Sebastian and Wu, Yonghui and Alayrac, Jean-Baptiste and Yu, Jiahui and Soricut, Radu and Schalkwyk, Johan and Dai, Andrew M and Hauth, Anja and others},
	Journal = {arXiv preprint arXiv:2312.11805},
	Title = {Gemini: a family of highly capable multimodal models},
	Year = {2023}}

@article{li2023blip,
	Author = {Li, Junnan and Li, Dongxu and Savarese, Silvio and Hoi, Steven},
	Journal = {arXiv preprint arXiv:2301.12597},
	Title = {Blip-2: Bootstrapping language-image pre-training with frozen image encoders and large language models},
	Year = {2023}}

@misc{instructblip,
	Archiveprefix = {arXiv},
	Author = {Wenliang Dai and Junnan Li and Dongxu Li and Anthony Meng Huat Tiong and Junqi Zhao and Weisheng Wang and Boyang Li and Pascale Fung and Steven Hoi},
	Eprint = {2305.06500},
	Primaryclass = {cs.CV},
	Title = {InstructBLIP: Towards General-purpose Vision-Language Models with Instruction Tuning},
	Year = {2023}}

@article{zhu2023minigpt,
	Author = {Zhu, Deyao and Chen, Jun and Shen, Xiaoqian and Li, Xiang and Elhoseiny, Mohamed},
	Journal = {arXiv preprint arXiv:2304.10592},
	Title = {Minigpt-4: Enhancing vision-language understanding with advanced large language models},
	Year = {2023}}

@article{liu2024visual,
	Author = {Liu, Haotian and Li, Chunyuan and Wu, Qingyang and Lee, Yong Jae},
	Journal = {Advances in neural information processing systems},
	Title = {Visual instruction tuning},
	Volume = {36},
	Year = {2024}}

@article{alayrac2022flamingo,
	Author = {Alayrac, Jean-Baptiste and Donahue, Jeff and Luc, Pauline and Miech, Antoine and Barr, Iain and Hasson, Yana and Lenc, Karel and Mensch, Arthur and Millican, Katherine and Reynolds, Malcolm and others},
	Journal = {Advances in Neural Information Processing Systems},
	Pages = {23716--23736},
	Title = {Flamingo: a visual language model for few-shot learning},
	Volume = {35},
	Year = {2022}}

@article{brown2020language,
	Author = {Brown, Tom B},
	Journal = {arXiv preprint ArXiv:2005.14165},
	Title = {Language models are few-shot learners},
	Year = {2020}}

@article{wei2021finetuned,
	Author = {Wei, Jason and Bosma, Maarten and Zhao, Vincent Y and Guu, Kelvin and Yu, Adams Wei and Lester, Brian and Du, Nan and Dai, Andrew M and Le, Quoc V},
	Journal = {arXiv preprint arXiv:2109.01652},
	Title = {Finetuned language models are zero-shot learners},
	Year = {2021}}

@article{dong2022survey,
	Author = {Dong, Qingxiu and Li, Lei and Dai, Damai and Zheng, Ce and Wu, Zhiyong and Chang, Baobao and Sun, Xu and Xu, Jingjing and Sui, Zhifang},
	Journal = {arXiv preprint arXiv:2301.00234},
	Title = {A survey on in-context learning},
	Year = {2022}}

@article{yu2022hierarchical,
	Author = {Yu, Jianfei and Chen, Kai and Xia, Rui},
	Journal = {IEEE Transactions on Affective Computing},
	Publisher = {IEEE},
	Title = {Hierarchical interactive multimodal transformer for aspect-based multimodal sentiment analysis},
	Year = {2022}}

@article{zhou2021masad,
	Author = {Zhou, Jie and Zhao, Jiabao and Huang, Jimmy Xiangji and Hu, Qinmin Vivian and He, Liang},
	Journal = {Neurocomputing},
	Pages = {47--58},
	Publisher = {Elsevier},
	Title = {MASAD: A large-scale dataset for multimodal aspect-based sentiment analysis},
	Volume = {455},
	Year = {2021}}

@inproceedings{borth2013large,
	Author = {Borth, Damian and Ji, Rongrong and Chen, Tao and Breuel, Thomas and Chang, Shih-Fu},
	Booktitle = {Proceedings of the 21st ACM international conference on Multimedia},
	Pages = {223--232},
	Title = {Large-scale visual sentiment ontology and detectors using adjective noun pairs},
	Year = {2013}}

@article{wang2024adaptive,
	Author = {Wang, Chunmei and Luo, Yuan and Meng, Chunli and Yuan, Feiniu},
	Journal = {ACM Transactions on Asian and Low-Resource Language Information Processing},
	Publisher = {ACM New York, NY},
	Title = {An adaptive Dual Graph Convolution Fusion Network for Aspect-Based Sentiment Analysis},
	Year = {2024}}

@article{xiao2023cross,
	Author = {Xiao, Luwei and Wu, Xingjiao and Yang, Shuwen and Xu, Junjie and Zhou, Jie and He, Liang},
	Journal = {Information Processing \& Management},
	Number = {6},
	Pages = {103508},
	Title = {Cross-modal fine-grained alignment and fusion network for multimodal aspect-based sentiment analysis},
	Volume = {60},
	Year = {2023}}

@inproceedings{zhao2023fusion,
	Author = {Zhao, Jun and Yang, Fuping},
	Booktitle = {2023 IEEE 6th Information Technology, Networking, Electronic and Automation Control Conference (ITNEC)},
	Pages = {336--340},
	Title = {Fusion with GCN and SE-ResNeXt network for aspect based multimodal sentiment analysis},
	Volume = {6},
	Year = {2023}}

@article{mu2023mocolnet,
	Author = {Mu, Jie and Nie, Feiping and Wang, Wei and Xu, Jian and Zhang, Jing and Liu, Han},
	Journal = {IEEE Transactions on Knowledge and Data Engineering},
	Publisher = {IEEE},
	Title = {MOCOLNet: A Momentum Contrastive Learning Network for Multimodal Aspect-Level Sentiment Analysis},
	Year = {2023}}

@article{zhang2021modalnet,
	Author = {Zhang, Zhe and Wang, Zhu and Li, Xiaona and Liu, Nannan and Guo, Bin and Yu, Zhiwen},
	Journal = {World Wide Web},
	Pages = {1957--1974},
	Publisher = {Springer},
	Title = {ModalNet: an aspect-level sentiment classification model by exploring multimodal data with fusion discriminant attentional network},
	Volume = {24},
	Year = {2021}}

@article{yang2024multi,
	Author = {Yang, Juan and Xiao, Yali and Du, Xu},
	Journal = {Knowledge-Based Systems},
	Pages = {111724},
	Publisher = {Elsevier},
	Title = {Multi-grained fusion network with self-distillation for aspect-based multimodal sentiment analysis},
	Volume = {293},
	Year = {2024}}

@inproceedings{DBLP:conf/interspeech/GongCG21,
	Author = {Yuan Gong and Yu{-}An Chung and James R. Glass},
	Booktitle = {22nd Annual Conference of the International Speech Communication Association, Interspeech 2021, Brno, Czechia, August 30 - September 3, 2021},
	Editor = {Hynek Hermansky and Honza Cernock{\'{y}} and Luk{\'{a}}s Burget and Lori Lamel and Odette Scharenborg and Petr Motl{\'{\i}}cek},
	Pages = {571--575},
	Publisher = {{ISCA}},
	Title = {{AST:} Audio Spectrogram Transformer},
	Year = {2021}}

@inproceedings{DBLP:conf/iclr/DosovitskiyB0WZ21,
	Author = {Alexey Dosovitskiy and Lucas Beyer and Alexander Kolesnikov and Dirk Weissenborn and Xiaohua Zhai and Thomas Unterthiner and Mostafa Dehghani and Matthias Minderer and Georg Heigold and Sylvain Gelly and Jakob Uszkoreit and Neil Houlsby},
	Booktitle = {9th International Conference on Learning Representations, {ICLR} 2021, Virtual Event, Austria, May 3-7, 2021},
	Publisher = {OpenReview.net},
	Title = {An Image is Worth 16x16 Words: Transformers for Image Recognition at Scale},
	Year = {2021}}

@inproceedings{DBLP:conf/emnlp/FanYDGBYXM19,
	Author = {Chuang Fan and Hongyu Yan and Jiachen Du and Lin Gui and Lidong Bing and Min Yang and Ruifeng Xu and Ruibin Mao},
	Booktitle = {Proceedings of the 2019 Conference on Empirical Methods in Natural Language Processing and the 9th International Joint Conference on Natural Language Processing, {EMNLP-IJCNLP} 2019, Hong Kong, China, November 3-7, 2019},
	Pages = {5613--5623},
	Title = {A Knowledge Regularized Hierarchical Approach for Emotion Cause Analysis},
	Year = {2019}}

@inproceedings{DBLP:conf/aaai/SpeerCH17,
	Author = {Robyn Speer and Joshua Chin and Catherine Havasi},
	Booktitle = {Proceedings of the Thirty-First {AAAI} Conference on Artificial Intelligence, February 4-9, 2017, San Francisco, California, {USA}},
	Pages = {4444--4451},
	Title = {ConceptNet 5.5: An Open Multilingual Graph of General Knowledge},
	Year = {2017}}

@inproceedings{DBLP:conf/ijcai/ZhangKSR20,
	Author = {Hongming Zhang and Daniel Khashabi and Yangqiu Song and Dan Roth},
	Booktitle = {Proceedings of the Twenty-Ninth International Joint Conference on Artificial Intelligence, {IJCAI} 2020},
	Pages = {4004--4010},
	Title = {TransOMCS: From Linguistic Graphs to Commonsense Knowledge},
	Year = {2020}}

@inproceedings{DBLP:conf/aaai/SapBABLRRSC19,
	Author = {Maarten Sap and Ronan Le Bras and Emily Allaway and Chandra Bhagavatula and Nicholas Lourie and Hannah Rashkin and Brendan Roof and Noah A. Smith and Yejin Choi},
	Booktitle = {The Thirty-Third {AAAI} Conference on Artificial Intelligence, {AAAI} 2019, The Thirty-First Innovative Applications of Artificial Intelligence Conference, {IAAI} 2019, The Ninth {AAAI} Symposium on Educational Advances in Artificial Intelligence, {EAAI} 2019, Honolulu, Hawaii, USA, January 27 - February 1, 2019},
	Pages = {3027--3035},
	Title = {{ATOMIC:} An Atlas of Machine Commonsense for If-Then Reasoning},
	Year = {2019}}

@article{DBLP:journals/corr/abs-2204-02549,
	Author = {Dawei Li and Yanran Li and Jiayi Zhang and Ke Li and Chen Wei and Jianwei Cui and Bin Wang},
	Journal = {CoRR},
	Title = {{C3KG:} {A} Chinese Commonsense Conversation Knowledge Graph},
	Volume = {abs/2204.02549},
	Year = {2022}}

@inproceedings{DBLP:conf/acl/BosselutRSMCC19,
	Author = {Antoine Bosselut and Hannah Rashkin and Maarten Sap and Chaitanya Malaviya and Asli Celikyilmaz and Yejin Choi},
	Booktitle = {Proceedings of the 57th Conference of the Association for Computational Linguistics, {ACL} 2019, Florence, Italy, July 28- August 2, 2019, Volume 1: Long Papers},
	Pages = {4762--4779},
	Title = {{COMET:} Commonsense Transformers for Automatic Knowledge Graph Construction},
	Year = {2019}}

@inproceedings{DBLP:conf/acl/ChenQ19,
	Author = {Zhuang Chen and Tieyun Qian},
	Booktitle = {Proceedings of the 57th Conference of the Association for Computational Linguistics, {ACL} 2019, Florence, Italy, July 28- August 2, 2019, Volume 1: Long Papers},
	Editor = {Anna Korhonen and David R. Traum and Llu{\'{\i}}s M{\`{a}}rquez},
	Pages = {547--556},
	Publisher = {Association for Computational Linguistics},
	Title = {Transfer Capsule Network for Aspect Level Sentiment Classification},
	Year = {2019}}

@inproceedings{DBLP:conf/acl/YanDJQ020,
	Author = {Hang Yan and Junqi Dai and Tuo Ji and Xipeng Qiu and Zheng Zhang},
	Booktitle = {Proceedings of the 59th Annual Meeting of the Association for Computational Linguistics and the 11th International Joint Conference on Natural Language Processing, {ACL/IJCNLP} 2021, (Volume 1: Long Papers), Virtual Event, August 1-6, 2021},
	Editor = {Chengqing Zong and Fei Xia and Wenjie Li and Roberto Navigli},
	Pages = {2416--2429},
	Publisher = {Association for Computational Linguistics},
	Title = {A Unified Generative Framework for Aspect-based Sentiment Analysis},
	Year = {2021}}

@article{DBLP:journals/kbs/LiZGY23,
	Author = {Min Li and Hui Zhao and Tiquan Gu and Di Ying},
	Journal = {Knowl. Based Syst.},
	Pages = {110703},
	Title = {Experiencer-Driven and Knowledge-Aware Graph Model for emotion-cause pair extraction},
	Volume = {278},
	Year = {2023}}

@inproceedings{DBLP:conf/acl/ZhouMLXDZXL20,
	Author = {Jie Zhou and Chunping Ma and Dingkun Long and Guangwei Xu and Ning Ding and Haoyu Zhang and Pengjun Xie and Gongshen Liu},
	Booktitle = {Proceedings of the 58th Annual Meeting of the Association for Computational Linguistics, {ACL} 2020, Online, July 5-10, 2020},
	Editor = {Dan Jurafsky and Joyce Chai and Natalie Schluter and Joel R. Tetreault},
	Pages = {1106--1117},
	Publisher = {Association for Computational Linguistics},
	Title = {Hierarchy-Aware Global Model for Hierarchical Text Classification},
	Year = {2020}}

@inproceedings{DBLP:conf/naacl/YangYDHSH16,
	Author = {Zichao Yang and Diyi Yang and Chris Dyer and Xiaodong He and Alexander J. Smola and Eduard H. Hovy},
	Booktitle = {{NAACL} {HLT} 2016, The 2016 Conference of the North American Chapter of the Association for Computational Linguistics: Human Language Technologies, San Diego California, USA, June 12-17, 2016},
	Editor = {Kevin Knight and Ani Nenkova and Owen Rambow},
	Pages = {1480--1489},
	Publisher = {The Association for Computational Linguistics},
	Title = {Hierarchical Attention Networks for Document Classification},
	Year = {2016}}

@inproceedings{DBLP:conf/nips/LiSGJXH21,
	Author = {Junnan Li and Ramprasaath R. Selvaraju and Akhilesh Gotmare and Shafiq R. Joty and Caiming Xiong and Steven Chu{-}Hong Hoi},
	Booktitle = {Advances in Neural Information Processing Systems 34: Annual Conference on Neural Information Processing Systems 2021, NeurIPS 2021, December 6-14, 2021, virtual},
	Editor = {Marc'Aurelio Ranzato and Alina Beygelzimer and Yann N. Dauphin and Percy Liang and Jennifer Wortman Vaughan},
	Pages = {9694--9705},
	Title = {Align before Fuse: Vision and Language Representation Learning with Momentum Distillation},
	Year = {2021}}

@article{DBLP:journals/corr/abs-2208-10442,
	Author = {Wenhui Wang and Hangbo Bao and Li Dong and Johan Bjorck and Zhiliang Peng and Qiang Liu and Kriti Aggarwal and Owais Khan Mohammed and Saksham Singhal and Subhojit Som and Furu Wei},
	Journal = {CoRR},
	Title = {Image as a Foreign Language: BEiT Pretraining for All Vision and Vision-Language Tasks},
	Volume = {abs/2208.10442},
	Year = {2022}}

@inproceedings{DBLP:conf/icassp/AldenehP17,
	Author = {Zakaria Aldeneh and Emily Mower Provost},
	Booktitle = {2017 {IEEE} International Conference on Acoustics, Speech and Signal Processing, {ICASSP} 2017, New Orleans, LA, USA, March 5-9, 2017},
	Pages = {2741--2745},
	Publisher = {{IEEE}},
	Title = {Using regional saliency for speech emotion recognition},
	Year = {2017}}

@inproceedings{DBLP:conf/interspeech/LiSMGD18,
	Author = {Pengcheng Li and Yan Song and Ian McLoughlin and Wu Guo and Lirong Dai},
	Booktitle = {19th Annual Conference of the International Speech Communication Association, Interspeech 2018, Hyderabad, India, September 2-6, 2018},
	Editor = {B. Yegnanarayana},
	Pages = {3087--3091},
	Publisher = {{ISCA}},
	Title = {An Attention Pooling Based Representation Learning Method for Speech Emotion Recognition},
	Year = {2018}}

@inproceedings{DBLP:conf/icassp/TrigeorgisRBMNS16,
	Author = {George Trigeorgis and Fabien Ringeval and Raymond Brueckner and Erik Marchi and Mihalis A. Nicolaou and Bj{\"{o}}rn W. Schuller and Stefanos Zafeiriou},
	Booktitle = {2016 {IEEE} International Conference on Acoustics, Speech and Signal Processing, {ICASSP} 2016, Shanghai, China, March 20-25, 2016},
	Pages = {5200--5204},
	Publisher = {{IEEE}},
	Title = {Adieu features? End-to-end speech emotion recognition using a deep convolutional recurrent network},
	Year = {2016}}

@inproceedings{DBLP:conf/icassp/MirsamadiBZ17,
	Author = {Seyedmahdad Mirsamadi and Emad Barsoum and Cha Zhang},
	Booktitle = {2017 {IEEE} International Conference on Acoustics, Speech and Signal Processing, {ICASSP} 2017, New Orleans, LA, USA, March 5-9, 2017},
	Pages = {2227--2231},
	Publisher = {{IEEE}},
	Title = {Automatic speech emotion recognition using recurrent neural networks with local attention},
	Year = {2017}}

@inproceedings{DBLP:conf/interspeech/ZhaoZZWZL18,
	Author = {Ziping Zhao and Yu Zheng and Zixing Zhang and Haishuai Wang and Yiqin Zhao and Chao Li},
	Booktitle = {19th Annual Conference of the International Speech Communication Association, Interspeech 2018, Hyderabad, India, September 2-6, 2018},
	Editor = {B. Yegnanarayana},
	Pages = {272--276},
	Publisher = {{ISCA}},
	Title = {Exploring Spatio-Temporal Representations by Integrating Attention-based Bidirectional-LSTM-RNNs and FCNs for Speech Emotion Recognition},
	Year = {2018}}

@inproceedings{DBLP:conf/interspeech/LuoZH18,
	Author = {Danqing Luo and Yuexian Zou and Dongyan Huang},
	Booktitle = {19th Annual Conference of the International Speech Communication Association, Interspeech 2018, Hyderabad, India, September 2-6, 2018},
	Editor = {B. Yegnanarayana},
	Pages = {152--156},
	Publisher = {{ISCA}},
	Title = {Investigation on Joint Representation Learning for Robust Feature Extraction in Speech Emotion Recognition},
	Year = {2018}}

@article{DBLP:journals/corr/abs-2302-13971,
	Author = {Hugo Touvron and Thibaut Lavril and Gautier Izacard and Xavier Martinet and Marie{-}Anne Lachaux and Timoth{\'{e}}e Lacroix and Baptiste Rozi{\`{e}}re and Naman Goyal and Eric Hambro and Faisal Azhar and Aur{\'{e}}lien Rodriguez and Armand Joulin and Edouard Grave and Guillaume Lample},
	Journal = {CoRR},
	Title = {LLaMA: Open and Efficient Foundation Language Models},
	Volume = {abs/2302.13971},
	Year = {2023}}

@article{wangreview,
  title={A Review of Chinese Sentiment Analysis: Subjects, Methods, and Trends},
  author={WANG, Zhaoxia and ZHANG, Xinyue and CUI, Jingfeng and HO, Seng-Beng and CAMBRIA, Erik}
}
\bibliographystyle{manuscript}

\vfill

\end{document}